%% file: main.tex
\documentclass{article}

\PassOptionsToPackage{numbers}{natbib}
\usepackage[final]{tencent_hunyuan}

\usepackage[utf8]{inputenc}
\usepackage[T1]{fontenc}
\usepackage{url}
\usepackage{booktabs}
\usepackage{amsfonts}
\usepackage{nicefrac}
\usepackage{microtype}
\usepackage{xcolor} 

\usepackage{graphicx}

\definecolor{darkblue}{rgb}{0, 0.12, 0.55}
\definecolor{darkgreen}{rgb}{0, 0.55, 0.12}
\definecolor{darkred}{rgb}{0.6,0,0}
\definecolor{darkgreen}{rgb}{0,0.6,0}
\usepackage[colorlinks, linkcolor=darkred, urlcolor=darkred, citecolor=teal]{hyperref}

\newcommand\eg{\textit{e.g}}
\newcommand\ie{\textit{i.e}}

\usepackage{cleveref}
\newcommand{\name}{Hunyuan-Game}
\newcommand{\logo}{\parbox[c]{.075\linewidth}{\includegraphics[width=\linewidth]{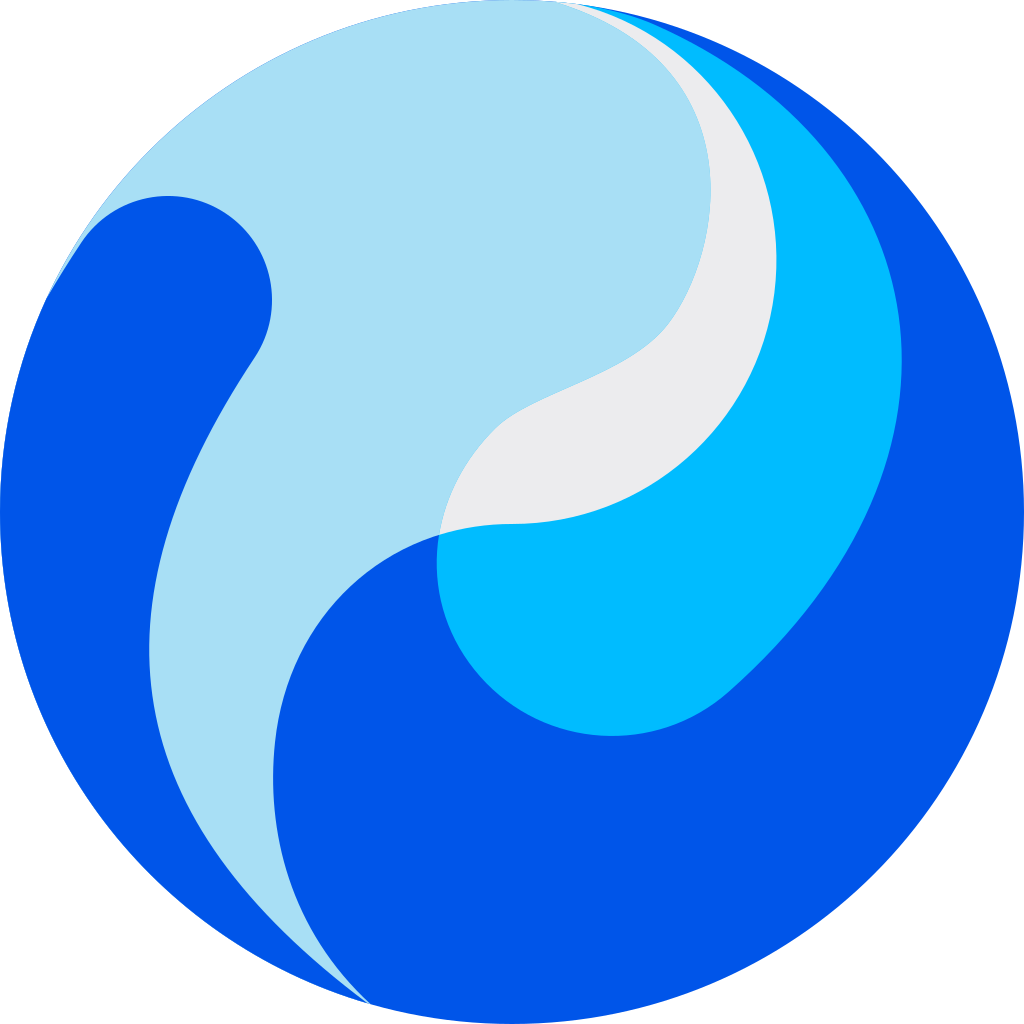}}}

\begin{document}
\title{\logo \ \ \name}

\author{
  Tencent Hunyuan\thanks{ corresponding author (Email: \url{qinglinlu@tencent.com})}
}

\maketitle
% 摘要+intro
\input{sections/abstract}
\input{sections/introduction}

% 生图
\input{sections/image/1_base_model}

\input{sections/image/2_text2texiao}

\input{sections/image/3_image2texiao}

\input{sections/image/4_transparent}

\input{sections/image/5_character}

% 生视频
\input{sections/video/1_i2v}

\input{sections/video/2_avatar}
\input{sections/video/3_lihui}

\input{sections/video/4_vsr}
\input{sections/video/5_interactive}
% 结论
\input{sections/conclusion}

{
\small
\bibliographystyle{plain}  
\bibliography{reference} 
}

\end{document}

%% file: sections/abstract.tex
\begin{abstract}

Intelligent game creation represents a transformative advancement in game development, utilizing generative artificial intelligence to dynamically generate and enhance game content. 
Despite notable progress in generative models, the comprehensive synthesis of high-quality game assets, including both images and videos, remains a challenging frontier. To create high-fidelity game content that simultaneously aligns with player preferences and significantly boosts designer efficiency, we present \textbf{\name}, an innovative 
% open-source 
project designed to revolutionize intelligent game production. 
{\name} encompasses two primary branches: image generation and video generation. The image generation component is built upon a vast dataset comprising billions of game images, leading to the development of a group of customized image generation models tailored for game scenarios:
(1) \textit{General Text-to-Image Generation}. 
(2) \textit{Game Visual Effects Generation}, involving text-to-effect and reference image-based game visual effect generation.
(3) \textit{Transparent Image Generation} for characters, scenes, and game visual effects. 
(4) \textit{Game Character Generation} based on sketches, black-and-white images, and white models.
The video generation component is built upon a comprehensive dataset of millions of game and anime videos, leading to the development of five core algorithmic models, each targeting critical pain points in game development and having robust adaptation to diverse game video scenarios:
(1) \textit{Image-to-Video Generation}.
(2) \textit{360$^{\circ}$ A/T Pose Character Video Generation}.
(3) \textit{Dynamic Illustration Generation}.
(4) \textit{Generative Video Super-Resolution}.
(5) \textit{Interactive Game Video Generation}. These image and video generation models not only exhibit high-level aesthetic expression but also deeply integrate domain-specific knowledge, establishing a systematic understanding of diverse game and anime art styles. Extensive experiments demonstrate our models' state-of-the-art performance, particularly in visual fidelity and motion naturalness, surpassing competitors like Midjourney, Kling and Wan in game scenarios. 
% By open-sourcing these models, 
We aim to encourage community-driven innovation, foster collaborative development, and pave the way for broader applications in the gaming industry. 
% Our codes are available at \url{https://github.com/Tencent/Hunyuan-Game}.

\begin{figure}[htbp]
    \centering
    \includegraphics[width=1.0\textwidth]{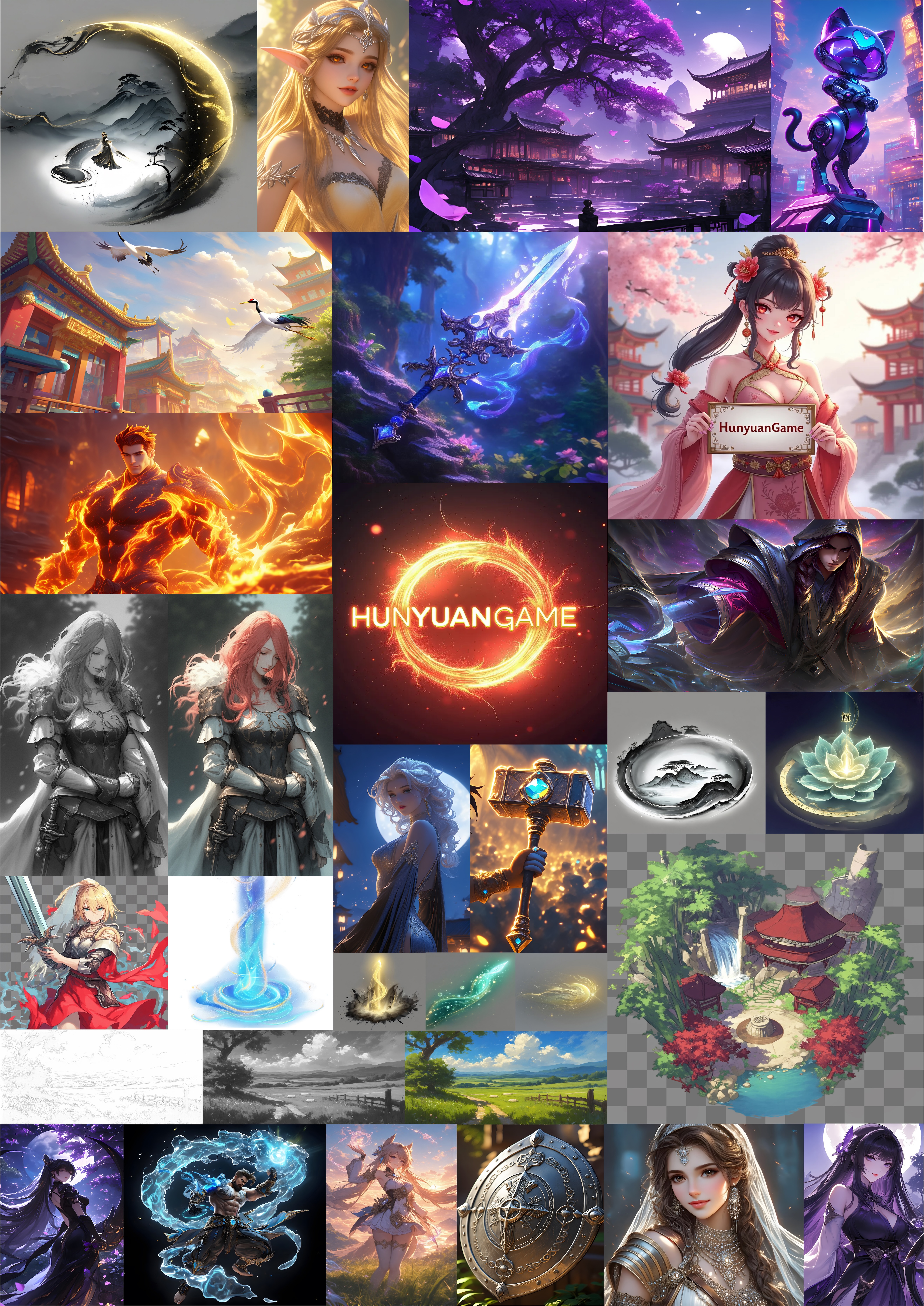}
    \caption{\textbf{Hunyuan-Game-Image}. The image generation capabilities of Hunyuan-Game include text-to-image generation, text-to-game effects generation, reference-based game visual effects generation, transparent and seamless image generation, and game character/scene generation. These capabilities offer a powerful toolset that significantly reduces the time and resources required for content creation, thereby greatly enhancing the efficiency of game asset production.}
    \label{fig:image_teaser}
\end{figure}

\begin{figure}[htbp]
    \centering
    \includegraphics[width=1.0\textwidth]{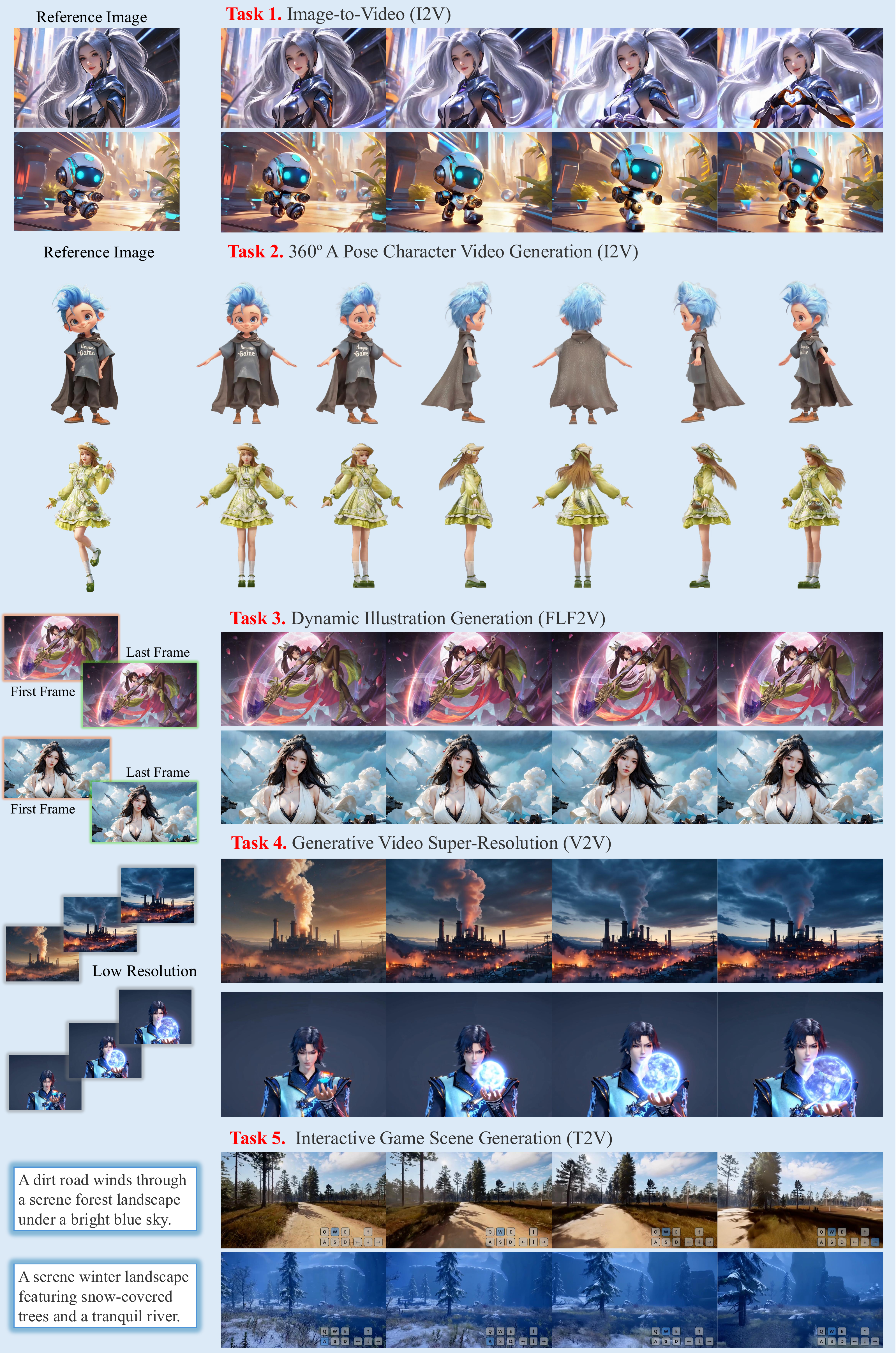}
    \caption{\textbf{Hunyuan-Game-Video}. The five key video generation capabilities of Hunyuan-Game are demonstrated as follows: an Image-to-Video generation and 360-degree A/T Pose Avatar Video Synthesis (I2V); Dynamic Illustration Generation based on first and last frame generation (FLF2V); video super-resolution from original video content (V2V); and Interactive Game Video Generation based on text or image input.}
    \label{fig:video_teaser}
\end{figure}

\end{abstract}

%% file: sections/introduction.tex
\section{Introduction}

Intelligent game creation represents a transformative advancement in the field of game development, leveraging generative artificial intelligence to automate and enhance various aspects of game creation. This technology enables the dynamic generation of game content, such as scenes, characters, and game visual effects, which can adapt to player preferences and behaviors. From the perspective of game designers, intelligent game generation offers a powerful toolset that significantly reduces the time and resources required for content creation. It allows designers to focus more on creative and strategic elements, as AI handles repetitive and labor-intensive tasks. This not only accelerates the development process but also fosters innovation by enabling the exploration of complex game mechanics and narratives that were previously impractical due to resource constraints.

The rapid advancement of generative models has revolutionized various domains, from entertainment to education. However, the synthesis of high-quality game assets, including both images and videos, remains a relatively untapped area, despite the growing demand for realistic and immersive gaming experiences. Game assets are not only a form of entertainment but also a crucial component in game development, marketing, and community engagement. The ability to generate professional-grade game assets can significantly enhance the storytelling and visual appeal of games, offering players a more engaging and dynamic experience. Additionally, it can improve designer efficiency, streamlining the creative process and allowing for faster iteration and innovation.

The creation of professional game assets involves several complex tasks, such as generating high-resolution content, transforming static images into dynamic sequences, and ensuring temporal consistency in animations. Traditional methods often require extensive manual effort and specialized skills, which can be both time-consuming and costly. The integration of advanced generation techniques into game development promises to address these challenges by automating content creation, thus enabling developers to focus on more creative aspects of game design. However, despite the progress in generative models, there remains a notable gap in the availability of comprehensive, open-source tools specifically tailored for game asset synthesis. In game asset generation, there are multiple challenges and difficulties that require advanced technology and innovative solutions to overcome:
\setlength{\leftmargini}{10pt}
\begin{itemize}
\item \textbf{Large-scale Data for Game-specific Scenarios:} The generation of high-quality game assets requires extensive datasets specific to game scenarios, which are often lacking.

\item \textbf{Alignment with Vertical Game Scenarios:} General models often lack the ability to align with the specific needs of the game industry, lacking a deep understanding and precise handling of the unique requirements involved in game development.

\item \textbf{Aesthetic Evaluation System:} There is a lack of a comprehensive, multi-dimensional aesthetic evaluation system tailored for game design, along with refined aesthetic evaluation operators. This results in insufficient integration of domain-specific knowledge from the gaming industry.

\item \textbf{Multi-dimensional Labeling System:} There is a need for a sophisticated captioning system that provides multi-dimensional labels (such as image content, art style, technical parameters) for game assets. This would enable precise and professional text-based control over the generated content.

\item \textbf{Visual Fidelity:} Game assets need high-quality visual effects to ensure an immersive experience for players. The generated content must have realistic details and complex textures to match the visual standards of modern games.

\item \textbf{Interactive Content Generation:} Game scenes often require dynamic changes and interactive elements, which demand that generation models can create and adjust content in real-time to adapt to player actions and game contexts.

\item \textbf{Open-domain Adaptability:} Game scenes need to adapt to different themes, styles, and settings, which requires models to have a high degree of adaptability to meet the needs of different games.
\end{itemize}
% This gap presents a significant opportunity to enhance game development processes, reduce production time, and elevate the overall quality of game content.

To address these unmet needs and further enhance the efficiency of intelligent game production, we introduce \textbf{\name}, the first innovative 
% open-source
project specifically designed for professional-grade game asset generation. {\name} is developed using an extensive dataset comprising billions of image data and millions of video data collected in game and anime scenarios, resulting in the creation of four image generation models and five video generation models. These models address significant challenges in game development and are highly adaptable to various game scenarios. The \textbf{Hunyuan-Game-Image} models (please refer to~\Cref{fig:image_teaser}) include:

\setlength{\leftmargini}{10pt}
\begin{itemize}
\item
\textbf{General Text-to-Image Generation:} The model is designed specifically for game scenarios, integrating deep knowledge of the game to enhance aesthetic expression and systematically understand diverse art styles. We introduce a prompt optimization model to improve image quality and provide precise semantic descriptions, lowering the barrier of professional use.

\item \textbf{Game Visual Effects Generation:} This is the first model for game visual effects generation, which generates game effects of different colors, shapes, styles, patterns, and additional elements. Designers can gain inspirational elements from the generated game visual effects.

\item \textbf{Transparent and Seamless Image Generation:}  Transparent image assets provide considerable flexibility, enabling the preservation of key elements, such as character, scene, and game visual effect, while facilitating swift update of background. Seamless images are widely applied in the design of game environments, contributing to resource optimization and other functions. Consequently, the integration of generative models to generate transparent images and seamless images can significantly improve efficiency.

\item \textbf{Game Character Generation:} In the realm of game character generation, a suite of model capabilities, encompassing the generation from \textit{line art} to \textit{grayscale drafts} and subsequently to \textit{character illustrations}, equips designers with a comprehensive workflow for generating character illustrations. Besides, we propose a consistent character generation method. Given a reference image, we can generate consistent character images from novel perspectives or poses based on the structural information derived from white models and depth maps. This approach ensures the maintenance of character integrity across varying viewpoints and poses.

% \item \textbf{Interactive Game Scene Generation:} This real-time interactive generation model facilitates dynamic content creation in game scenes, enabling users to produce video content through peripheral input. It is perfect for open-world games with complex interactions, offering a more engaging and responsive gaming experience.
\end{itemize}

The \textbf{Hunyuan-Game-Video} models (please refer to~\Cref{fig:video_teaser}) include:
\setlength{\leftmargini}{10pt}
\begin{itemize}
\item
\textbf{Image-to-Video Generation:} This model has demonstrated exceptional performance in the field of game video generation, surpassing the previous methods, such as Kling and Wan, in visual fidelity and temporal consistency. Moreover, the I2V model can serve as a foundation for many downstream tasks, thereby providing a robust fundamental capability.

\item \textbf{360° A/T Pose Character Video Synthesis:} As the first model of its kind in the industry, it allows for the creation of 360-degree rotation videos from any character illustration, ensuring A/T-pose standards. The generated video helps to eliminate visual blind spots from a single perspective, allowing the designer to evaluate the character design from different views.

\item \textbf{Dynamic Illustration Generation:} This model introduces the ability to generate dynamic illustrations, creating seamless looping animations from static character illustrations while maintaining high temporal consistency. It meets the demand for smooth and natural character movements in game animations. Dynamic Illustration has a wide range of applications, such as character entry animations and promotional videos.

\item \textbf{Generative Video Super-Resolution:} Our video super-resolution model is exceptional in the game and anime fields, transforming low-resolution assets into impressive 2K videos. This greatly improves visual quality without sacrificing detail, which is crucial for preserving the aesthetic quality of game visuals.

\item \textbf{Interactive Game Video Generation:} This real-time interactive generation model facilitates dynamic content creation in game scenes, enabling users to produce video content through peripheral input. It is perfect for open-world games with complex interactions, offering a more engaging and responsive gaming experience.
\end{itemize}

The capabilities of the above models are demonstrated through extensive experiments, where they outperform existing competitors, particularly in terms of visual fidelity and motion naturalness. In general, the key advantages of our project include:

\begin{itemize}
\item
\textbf{Expert-Level Models at Industry-Leading Standards:} Our models are specifically designed for game and anime scenarios, achieving state-of-the-art performance in the industry.
\item \textbf{Comprehensive Capability Coverage:} 
% We open-source four image generation abilities and five video generation capabilities, 
We introduce four image generation abilities and five video generation capabilities, 
ranging from text-to-image to image-to-video generation, evolving from text-to-game effects to dynamic game effects generation, and advancing from game character generation to dynamic character illustration generation. This comprehensive capability ensures the automated generation of high-quality and highly consistent image and video assets.

\item\textbf{Industry-First Capabilities:} We introduce the first-ever text-to-game visual effects generation, reference-based game visual effects generation, A/T pose Avatar video generation, and dynamic illustration generation model, setting new standards in the industry.

\item\textbf{High Stability, Generalization, and Consistency:} Our models exhibit exceptional stability, generalization, and consistency across various game scenarios.
\end{itemize}

% By open-sourcing the entire series of models, 
We aim to encourage community-driven innovation and foster collaborative development. We believe that \name\ will pave the way for broader applications in the game industry, ultimately transforming how game content is created and experienced.

%% file: sections/image/1_base_model.tex
\section{Hunyuan-Game-Image Generation}
\subsection{Text-to-Image Generation}
\subsubsection{Introduction}

With the development of large text-to-image models such as Flux~\cite{flux2024}, SD3~\cite{esser2024scaling}, and Midjourney~\cite{midjourney}, there exists escalating connections between AIGC models and the gaming industry. 
During the concept design phase, designers can use image generation models to inspire creativity, which requires that the models possess outstanding semantic understanding capabilities. 
Among all publicly available models, Midjourney's image generation models~\cite{midjourney} have gained extensive attention from designers of various industries, especially the gaming industry, due to their excellent aesthetic expressiveness. 
However, significant shortcomings still exist on Midjourney's models, especially when adapting them to specific scenarios within the gaming industry. This is mainly because these models possess drawbacks such as lack of deep understanding capabilities, and fail to precisely grasp the unique demands from users.
In recent days, domain-specific models and LoRA~\cite{hu2022lora, kumari2023multi} have become indispensable for many designers, as they can accurately match specific artistic styles and design concepts, enabling controllable generation of results. This indicates that it is necessary to train a group of image generation models that can meet the requirements of designers in the gaming industry.

To effectively tackle the challenges mentioned above, we propose a customized image generation model tailored for gaming scenarios. This model not only delivers high-level aesthetic expression but also deeply integrates domain knowledge from the gaming field, establishing a systematic cognitive framework for diverse game and animation art styles. 
The research team constructed a large-scale, high-quality game image dataset through rigorous collection and filtering, and built an efficient data processing and storage system based on data management operators. 

Additionally, a multidimensional aesthetic evaluation system was introduced, significantly enhancing the artistic quality of the model’s output images by developing refined aesthetic evaluation operators. To enable fine-grained textual control, we developed a multidimensional caption annotation system for gaming scenarios, capable of precisely describing image content, artistic style, technical parameters, and other relevant aspects. 

Furthermore, to lower the professional usage threshold and improve image generation quality, the team trained a prompt optimization model specialized for gaming scenarios, which automatically converts user-input natural language prompts into high-quality prompt sequences that comply with professional standards and technical requirements. The integrated application of these innovative technologies provides an intelligent solution for the game design domain that balances professionalism with ease of use.

\subsubsection{Data filtering}
\begin{figure}[htbp]
    \centering
    \includegraphics[width=0.8\textwidth]{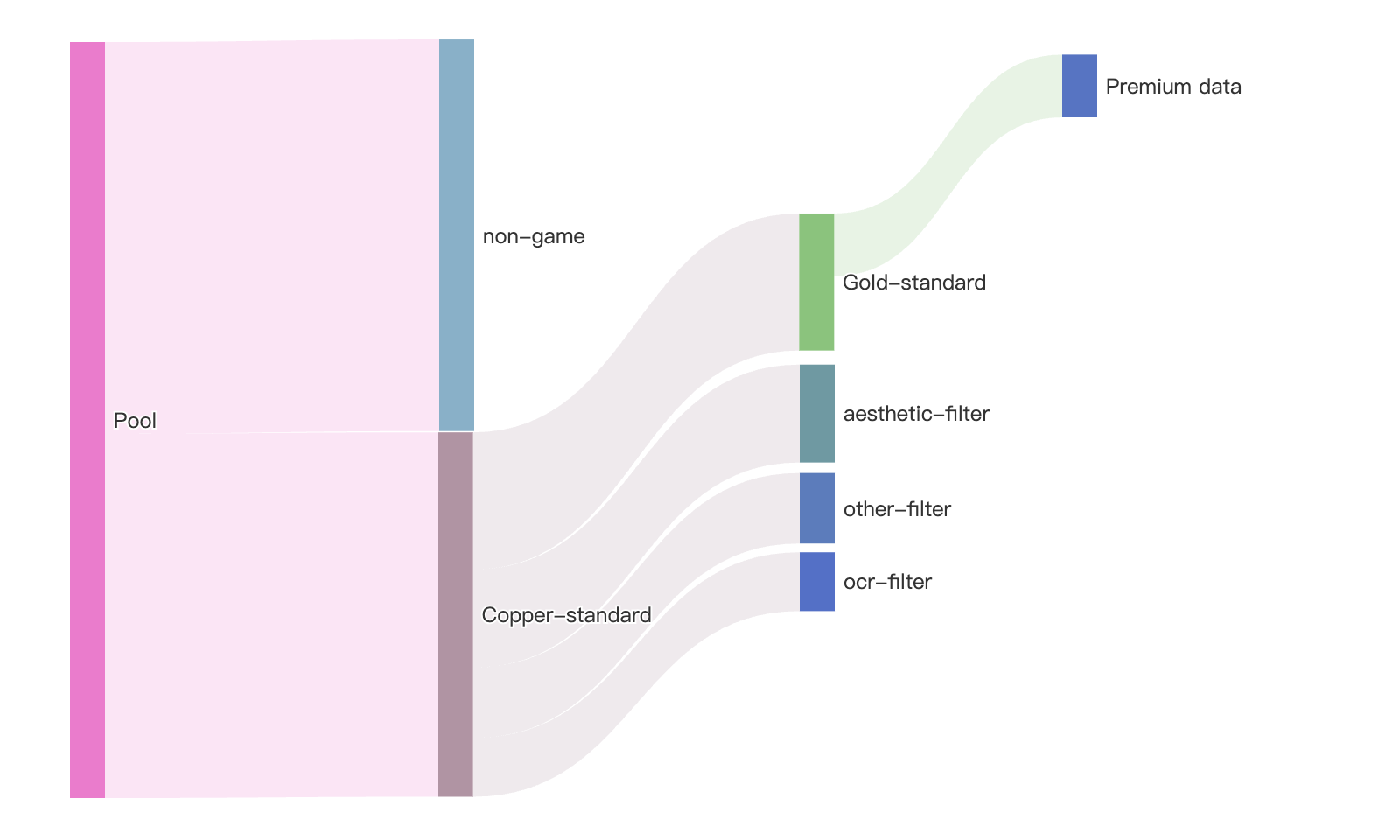}
    \caption{\textbf{The data filtering pipeline}.}
    \label{fig:2.1.2 data filtering}
\end{figure}
\textbf{Construction and Tiered Filtering of Game Datasets}

As shown in Figure ~\ref{fig:2.1.2 data filtering}, when constructing the game dataset for this study, we mainly utilize image data related to games, animation, and artistic image works; we also build a three-tier filtering system that rates these images into \textit{Bronze}, \textit{Gold}, and \textit{Premium} tiers. 
Based on the Premium tier, we developed a proprietary fine-grained data management system. Initially, a candidate game dataset comprising 193 million images was assembled, and a game image classification operator was trained to filter out data unrelated to games and animation. After this initial filtering by the classification operator, a Bronze-tier dataset of 93 million images conforming to game styles was obtained. Subsequently, the Bronze-tier dataset underwent basic image quality screening, with criteria including resolution (both dimensions $\geq$ 1024 pixels), clarity, Laion aesthetic scores, watermark presence, and optical character recognition (OCR) checks. This process yielded a Gold-tier dataset of 35 million images. To reduce machine annotation errors, the research team manually annotated the entire Gold-tier dataset, removing images with defects, AI-generated content (AIGC), and insufficient aesthetic quality. The final Premium-tier dataset comprises 16 million high-quality images.

\textbf{Construction of a Proprietary Aesthetic Scoring System}

\begin{figure}[htbp]
    \centering
    \includegraphics[width=1.0\textwidth]{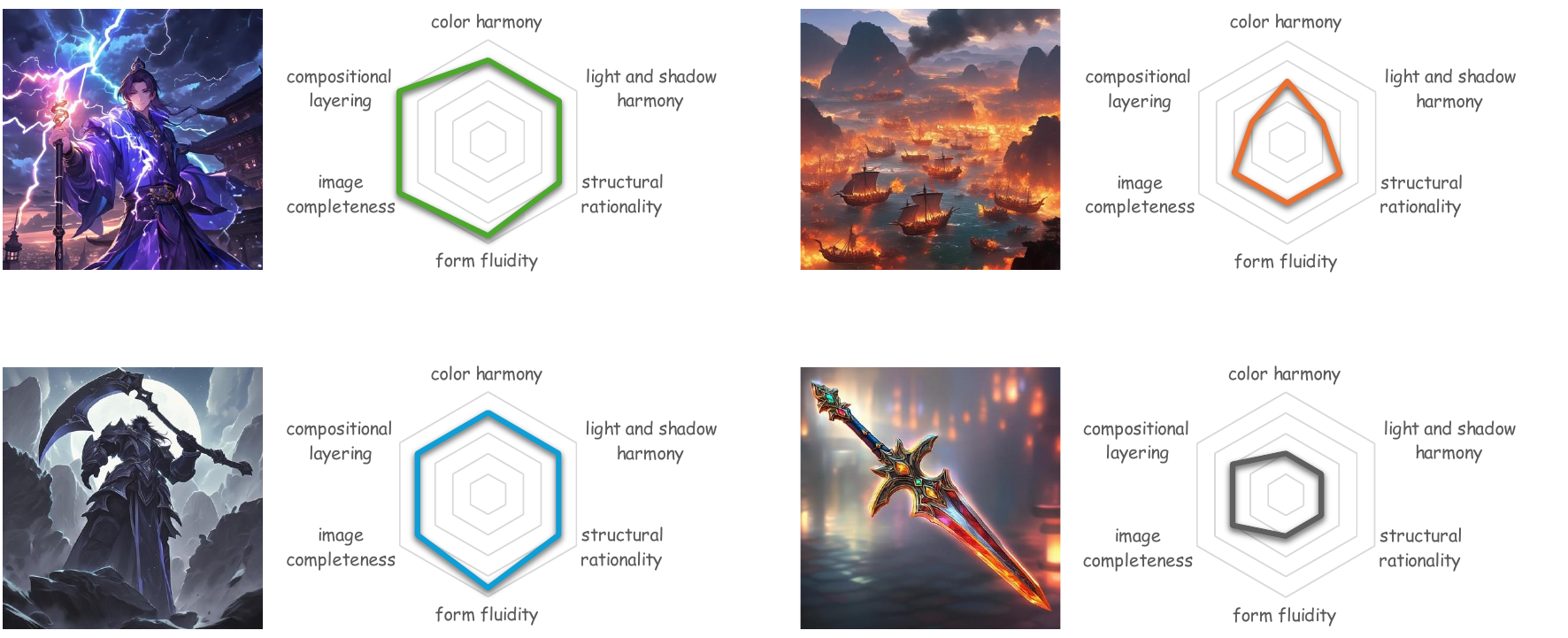}
    \caption{Examples of multi-dimensional aesthetic scores.}
    \label{fig:Aesthetic-Score}
\end{figure}

While the LAION aesthetic scoring operator~\cite{schuhmann2022laion} provides auxiliary support in filtering low-quality data, its discriminative capability significantly diminishes once image quality reaches a high level, making it insufficient for the fine-grained data selection required in downstream training tasks. To address this, our study developed a fine-grained, as shown in Figure~\ref{fig:Aesthetic-Score}, proprietary aesthetic scoring system encompassing six core dimensions: \textbf{color harmony}, \textbf{light and shadow harmony}, \textbf{structural rationality}, \textbf{form fluidity}, \textbf{image completeness}, and \textbf{compositional layering}.

Given the subjective nature of aesthetic evaluation, we collaborated with professional game designers to define detailed scoring dimensions and corresponding standards, employing a 1-to-5 rating scale for each dimension. During the annotation phase, we partnered with art institutions to form an annotation team whose members possess systematic aesthetic literacy and professional training. The annotation process was divided into three stages:

$\bullet$ \textbf{Standard Development, Training, and Pilot Annotation}. Professional designers conducted multiple rounds of training and pilot annotations for annotators, refining the scoring dimensions and standards based on feedback.

$\bullet$ \textbf{Formal Annotation}. To ensure quality, annotators rated only a single dimension per task, with each task cross-annotated by five individuals. An annotation was considered valid only if at least four out of five annotators (80\% agreement) assigned the same score. Annotators with persistent inconsistent annotations were removed. The mode of the five scores was taken as the final annotation score for each task.

$\bullet$ \textbf{Acceptance and Calibration}. Five percent of the annotated data was randomly sampled for quality inspection. Each sampled task was jointly scored by three standard-setting designers to serve as reference scores. The batch passed acceptance if the proportion of annotations exactly matching the reference scores was at least 70\%, and the proportion with a score difference of no more than one point was at least 95\%.
Ultimately, a team of 50 annotators was established. Following the above annotation strategy, 100,000 labeled samples were collected for each scoring dimension to train a multimodal model-based image feature extraction and regression scoring prediction model.

\subsubsection{Data Annotation}
\textbf{Multi-Length Caption Generation Strategy}

\begin{figure}[htbp]
    \centering
    \includegraphics[width=1.0\textwidth]{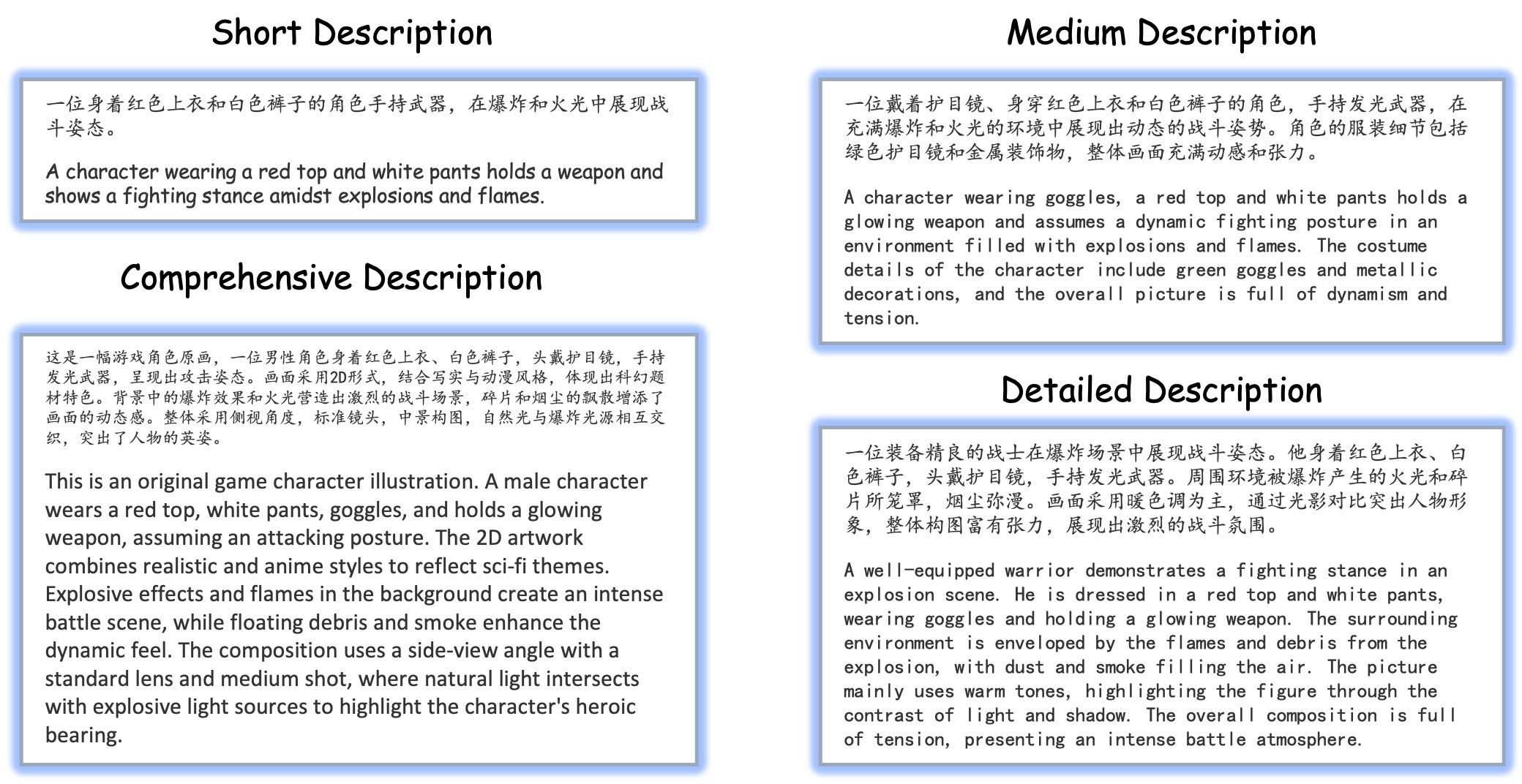}
    \caption{Examples of Multi-Length Captions.}
    \label{fig:caption-example_t2i}
\end{figure}
\textbf{Construction and Tiered Filtering of Game Datasets}

In this study, a proprietary captioning model was employed to generate textual annotations for image data. To ensure stable output across prompts of varying lengths, as shown in Figure~\ref{fig:caption-example_t2i}, four captions of different lengths were generated for each image, including:

$\bullet$ Short Description: Approximately 30 characters, briefly summarizing the main content;

$\bullet$ Medium Description: Approximately 60 characters, adding some image details;

$\bullet$ Detailed Description: Approximately 100 characters, comprehensively elaborating on image details;

$\bullet$ Comprehensive Description: Based on the detailed description, enriched with professional terminology covering image style, theme, composition, camera angle, depth of field, lighting, and specific IP-related information, enabling users to precisely control image generation through fine-grained professional terms.

During model training, captions were randomly sampled according to a 1:1:1:7 ratio.

\textbf{Aesthetic Information Fusion Mechanism}

Unlike VMix~\cite{wu2024vmix}, which takes aesthetic information as an additional control condition, we map fine-grained aesthetic scores to natural language descriptions and incorporate them into comprehensive descriptions according to a specific proportion, achieving a deep integration of aesthetic information and captions. This mechanism allows users to directly control aesthetic features during the image generation process through aesthetic descriptive language, significantly enhancing the controllability and professionalism of image generation.

\subsubsection{Training}
The game model in this study is developed based on the proprietary DiT architecture. It is derived from a self-developed general-purpose model and fine-tuned using game-specific data. The fine-tuning process consists of three stages, each employing different data and strategies to progressively enhance model performance, as detailed below: 

$\bullet$ \textbf{Full-Scale Data Fine-Tuning}.
In the initial fine-tuning stage, the model is trained on a comprehensive game dataset~(16 million). This stage aims to enable the model to preliminarily learn game style features and establish foundational recognition capabilities for game-related concepts. By learning from large-scale game data, the model captures visual patterns and semantic information of game scenes, characters, props, and other elements, laying the groundwork for subsequent optimization.

$\bullet$ \textbf{High-Quality Data Selection Fine-Tuning (Quality Tuning, QT Stage)}. In the second stage, i.e., QT stage, images within the game dataset are filtered using a proprietary aesthetic scoring operator. This operator evaluates multiple dimensions to effectively identify and select high-quality and aesthetically pleasing images. Fine-tuning the model with these curated images enhances its understanding and recognition of aesthetic features, significantly improving the quality of generated images to better align with game design aesthetics. 

\begin{figure}[htbp]
    \centering
    \includegraphics[width=1.0\textwidth]{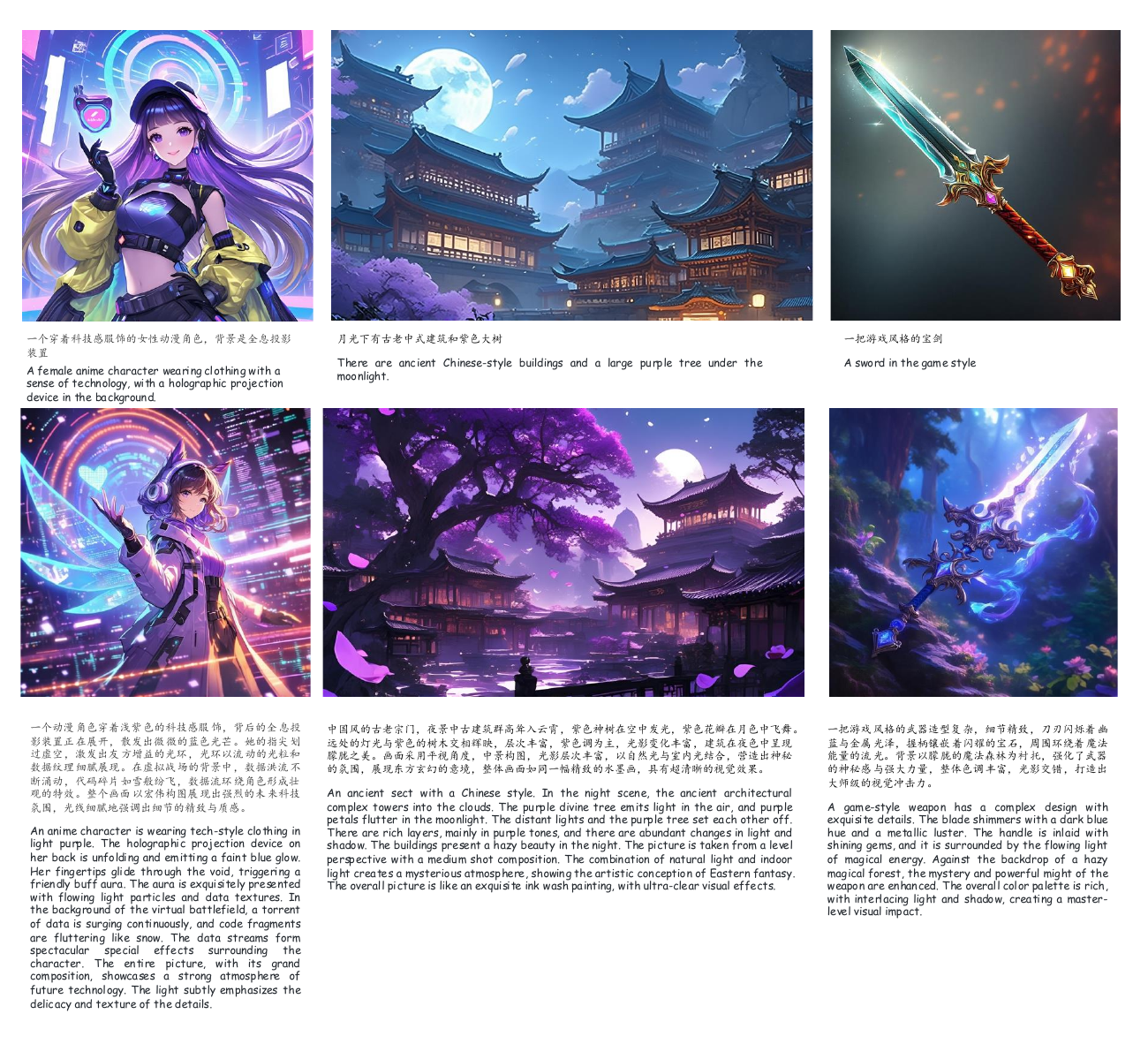}
    \vspace{-2em}
    \caption{Prompt rewriting can significantly add content information to the picture, thus enhancing the quality of the image. \textbf{Top row}: w/o prompt rewriting. \textbf{Bottom row}: w/ prompt rewriting.}
    \label{fig:t2i-recap-v2}
\end{figure}

\begin{figure}[htbp]
    \centering
    \includegraphics[width=1.0\textwidth]{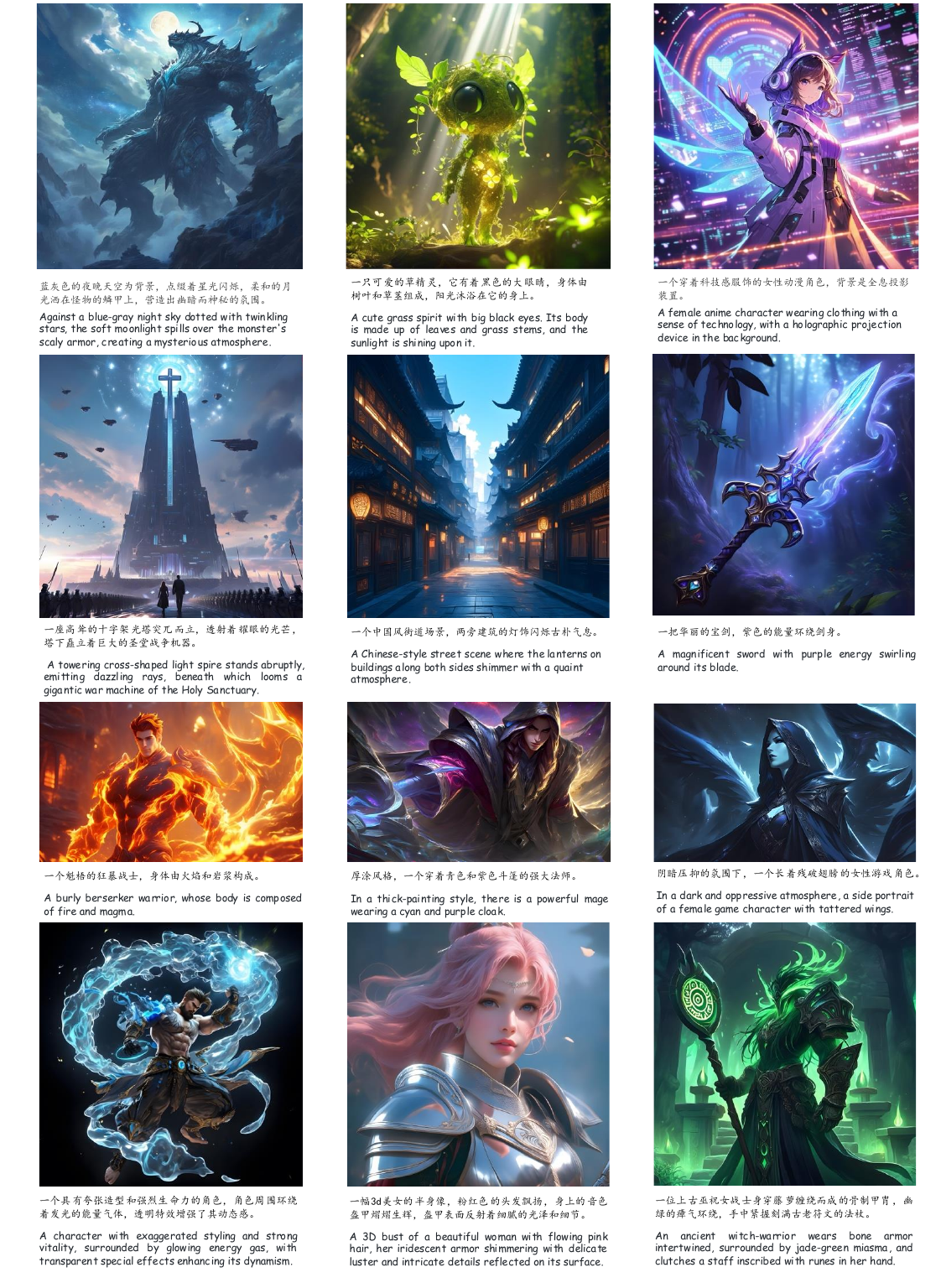}
    \caption{Visualizations of text-to-image generation results.}
    \label{fig:t2i_e_results}
\end{figure}

$\bullet$ \textbf{Post-Training Reinforcement Learning (Direct Preference Optimization, DPO~\cite{rafailov2023direct}~\cite{wallace2024diffusion} Strategy)}.
The third stage involves post-training reinforcement learning using the DPO strategy. This approach guides model parameter optimization based on preferences over generated outputs, further improving the stability of image generation. During this stage, the model maintains high aesthetic quality while reducing output variability, ensuring consistent and stable generation of high-quality images across diverse input conditions.

\subsubsection{Inference}

\textbf{Construction and Tiered Filtering of Game Datasets.}
In image generation tasks, high-quality input prompts play a critical role in determining the quality of the generated images. Detailed prompts that align with the aesthetic scoring system can guide the model to produce higher-quality and more targeted image content. However, this places significant demands on users’ professional knowledge and expressive abilities. To lower the user entry barrier and enhance interaction convenience, this study designed and implemented a specialized prompt optimization solution.

{\bf Prompt Rewriting}. This rewriting system is centered on natural language processing techniques. Through semantic understanding and generation algorithms, it transforms users’ original input prompts into more professional and directive natural language descriptions. Specifically, the system first performs semantic parsing of the user’s input text to identify key information. Then, based on a pre-constructed aesthetic knowledge graph and a style element database, combined with the semantic features of the user input, it automatically supplements matching artistic styles (e.g., cyberpunk, traditional Chinese ink painting), thematic categories (e.g., fantasy adventure, urban life), and visual elements (e.g., special props, iconic scenes). As shown in Figure~\ref{fig:t2i-recap-v2}, prompt rewriting can significantly add content information to the picture, thus enhancing the quality of the image.
In this way, the optimized prompts are not only semantically richer and more complete but also guide the model to generate images with more harmonious composition and stronger aesthetic expression. This approach significantly improves user creativity efficiency while effectively lowering the threshold for professional design use.

\subsubsection{Evaluation}
\textbf{Qualitative Results.}
As shown in Fig.~\ref{fig:t2i_e_results}, we present the results of our game-oriented text-to-image model, demonstrating that our method achieves high image fidelity and aesthetic quality across various categories, including portraits, characters, scenes, and weapons.

\textbf{Quantitative Results.} To systematically evaluate the professional performance of models in the gaming domain, this study constructs a validation set comprising 268 prompts. The validation set primarily contains 2D/3D anime-style descriptions and gaming-specific terminology, covering semantic representations of typical gaming visual elements. The experiment selects state-of-the-art commercial models (Jimeng 2.1, Flux-Pro 1.1, Midjourney 6.1) as comparison objects and employs a multi-dimensional cross-evaluation framework. For each image generated by the model, this study evaluates it across four dimensions: text-image consistency, game concept recognition, vividness of subject modeling, and pictorial aesthetics. The overall score is formed by calculating the average score of each dimension. The evaluation process involves a jury of three senior game art designers independently scoring each image using a 5-point scale, with the final score being the inter-rater mean to ensure reliability. The evaluation index system are presented in Tab. ~\ref{hunyuan-game t2i eval results}, in the scenario of generating images in the gaming field, our model has achieved the best results in both accuracy and aesthetics.

\begin{table}[h]
\centering
\caption{Results of Evaluation for Hunyuan-Game-Image Generation}
\label{hunyuan-game t2i eval results}
\scalebox{0.9}{
\begin{tabular}{lccccc}
\toprule
Model& T-I Alignment & Game Concept & Subject modeling & Aesthetics & Overall \\
\midrule
Jimeng 2.1& 3.68 & 4.47 & 3.55 & 3.65 & 3.83 \\
Midjourney 6.1 Pro& 3.55 & 4.2 & 3.50 & 3.54 & 3.71 \\
Flux-Pro 1.1& 3.57 & 4.25 & 3.35 & 3.47 & 3.66 \\
Ours& \textbf{4.09} & \textbf{4.65} & \textbf{3.87} & \textbf{3.89}& \textbf{4.12} \\
\bottomrule
\end{tabular}
}
\end{table}

%% file: sections/image/2_text2texiao.tex
\subsection{Text-to-\textit{Game Visual Effects}}

\subsubsection{Introduction}
% In the current flourishing development of the gaming industry, visual effects have become a core element in shaping the visual experience of games. Whether it is the skill effects during in-game battles or the highly dynamic atmospheric effects in character illustration, visual effects have deeply permeated the entire lifecycle of game development. They not only enhance the immersive experience of players through dynamic visual effects but also serve as a crucial medium for players to understand the artistic style of the game and promote the game's IP image. Today, the investment in visual effects for leading industry products continues to rise, with top-tier visual effects performance even becoming a key decision factor for users when choosing games, signifying that the quality of visual effects has become a core dimension for measuring the competitiveness of the game market.
In the rapidly growing game industry, visual effects have become a pivotal element in shaping the overall visual experience. Whether triggered during in-game combat skills or featured prominently in character promotional artwork, visual effects have deeply permeated the entire game development lifecycle. They not only enhance player immersion through dynamic visual feedback but also serve as critical carriers for conveying the game’s artistic style and propagating its intellectual property image. Currently, leading industry products are allocating an increasing proportion of resources to visual effects development, with top-tier visual effects quality emerging as a decisive factor influencing user game selection. This trend underscores that visual effects quality has ascended to a core dimension for evaluating market competitiveness in gaming.

However, in stark contrast to the urgent demand for high-quality visual effects in the market is the long-standing structural contradiction in the field of visual effects design. While current AIGC technology has established standardized solutions in subfields such as character concept art and scene generation, visual effects design remains predominantly manual. Mainstream AI image generation models in the market have low accuracy in recognizing visual effects-specific elements such as type, style, and texture, resulting in poor visual effects quality, which severely limits the potential for technological empowerment.

In this work, we propose a novel approach based on diffusion transformers (DiTs) as the foundational architecture. We first establish a comprehensive annotation system encompassing six dimensions of visual effects feature labels—including effect style, elemental composition, and motion trajectory—and construct a multi-dimensional captioning framework to enhance the model’s visual effects perception and generation robustness. Subsequently, we adopt a three-stage progressive training strategy to incrementally improve the visual fidelity of generated effects. Our method culminates in the development of the industry’s first dedicated AI-generated image model specialized for the visual effects domain.

\begin{figure}[!t]
    \centering
    \includegraphics[width=0.9\textwidth]{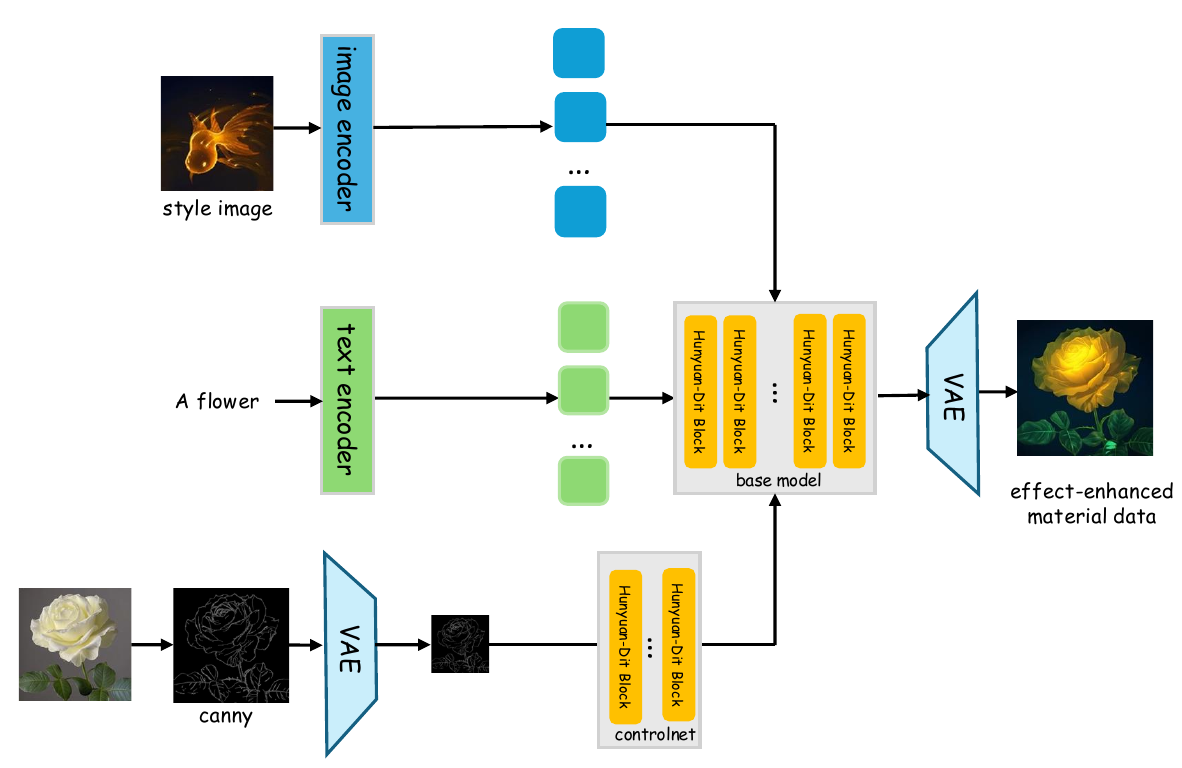}
    \caption{The pipeline to create effectualized materials.}
    \label{fig:Effect-enhanced_data_generate_method}
    \vspace{-0.5em}
\end{figure}
\subsubsection{Data}
Visual effects design, as a highly specialized field, faces dual challenges in data development: firstly, the gaming industry lacks a systematic standard for visual effects description; secondly, there is a scarcity of high-quality visual effects data in the market. To address these bottlenecks, this study adopts a three-stage advancement strategy to construct a visual effects data system: 
% initially, a manual approach is used to screen and manually process concept art-level visual effects data. 
Initially, high-quality visual effects data are manually curated and filtered. 
% Subsequently, the original visual effects model, fine-tuned using the data from the first step, is employed to infer visual effects prompts, achieving data self-iteration. 
Subsequently, a preliminary visual effects generation model is trained on the curated dataset obtained in the first stage, enabling further data expansion and augmentation.
Finally, the approach of material effectualization is used to expand the concept of materials with visual effects texture, forming a pyramid-shaped data ecosystem.

% $\bullet$ High-quality concept art-level visual effects data (thousands): 
$\bullet$ High-quality visual effects data (thousands scale): 
Collect visual effects design concept art from leading gaming companies and conduct structured annotation across six dimensions, including effect type, effect color, effect shape, effect style, motion trend, and effect elements, forming an original high-quality benchmark dataset.

% $\bullet$ Self-iterated data (tens of thousands): Use LLM to construct diverse prompts and employ the visual effects model fine-tuned with concept art to infer visual effects data. This data is manually screened and further annotated as self-iterated samples, which in turn feed back into the visual effects generation model.
$\bullet$ Progressive data augmentation (tens of thousands scale): A preliminary visual effects generation model is trained on the curated dataset obtained in the first stage. This model takes prompt inputs generated by a large language model (LLM) and produces additional synthesized visual effects samples. The generated samples serve to effectively augment the original dataset by introducing greater diversity and richness in data instances. Through this iterative data augmentation process, the dataset is progressively expanded, thereby enabling continuous refinement and improved performance of the visual effects generation model.

% $\bullet$ Effectualized data (millions): 
$\bullet$ Effect-enhanced material data (millions scale):
As shown in ~\Cref{fig:Effect-enhanced_data_generate_method}, we utilize a general text-to-image model to generate hundreds of thousands of material data (e.g., plants, animals, and objects). Given the original visual effects data as input, we employ a combined control generation method with ControlNet and IP-Adapter to expand effectualized materials.
\begin{figure}[!t]
    \centering
    \includegraphics[width=0.9\textwidth]{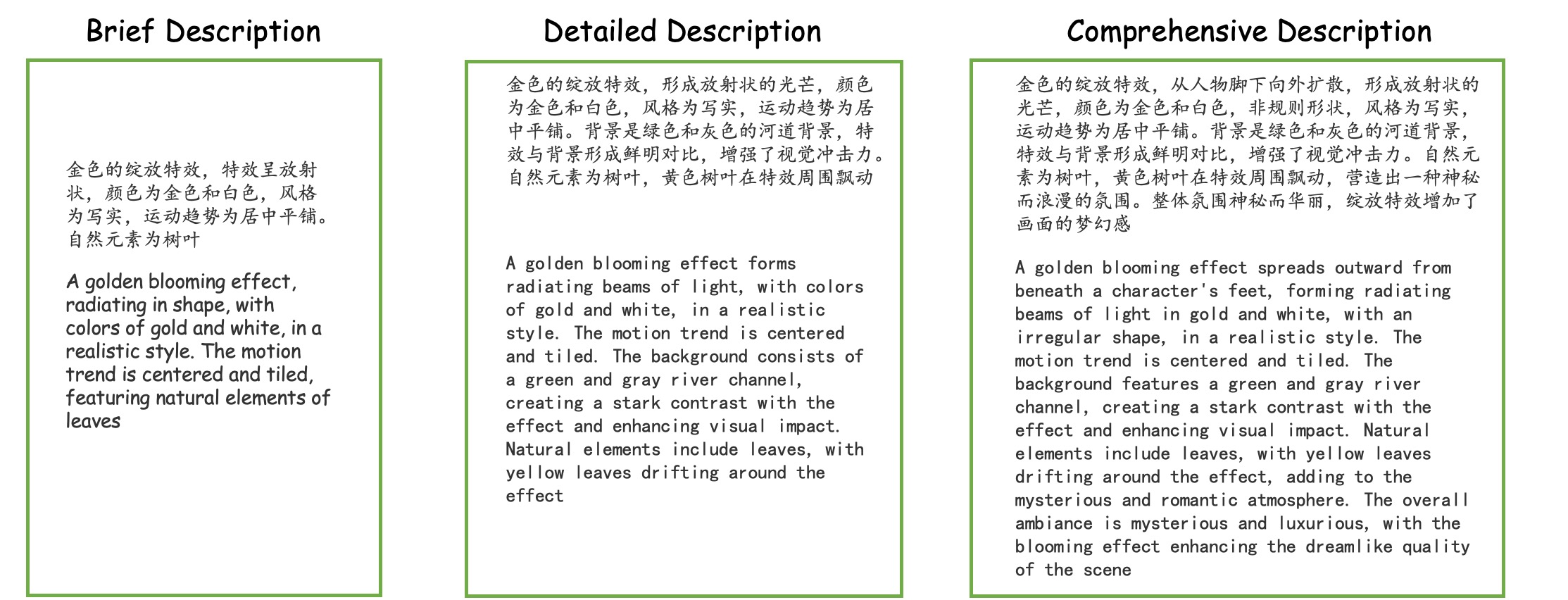}
    \caption{Examples of brief, detailed, and comprehensive descriptions.}
    \label{fig:caption-example}
\end{figure}
\begin{figure}[htbp]
    \centering
    \includegraphics[width=1\textwidth]{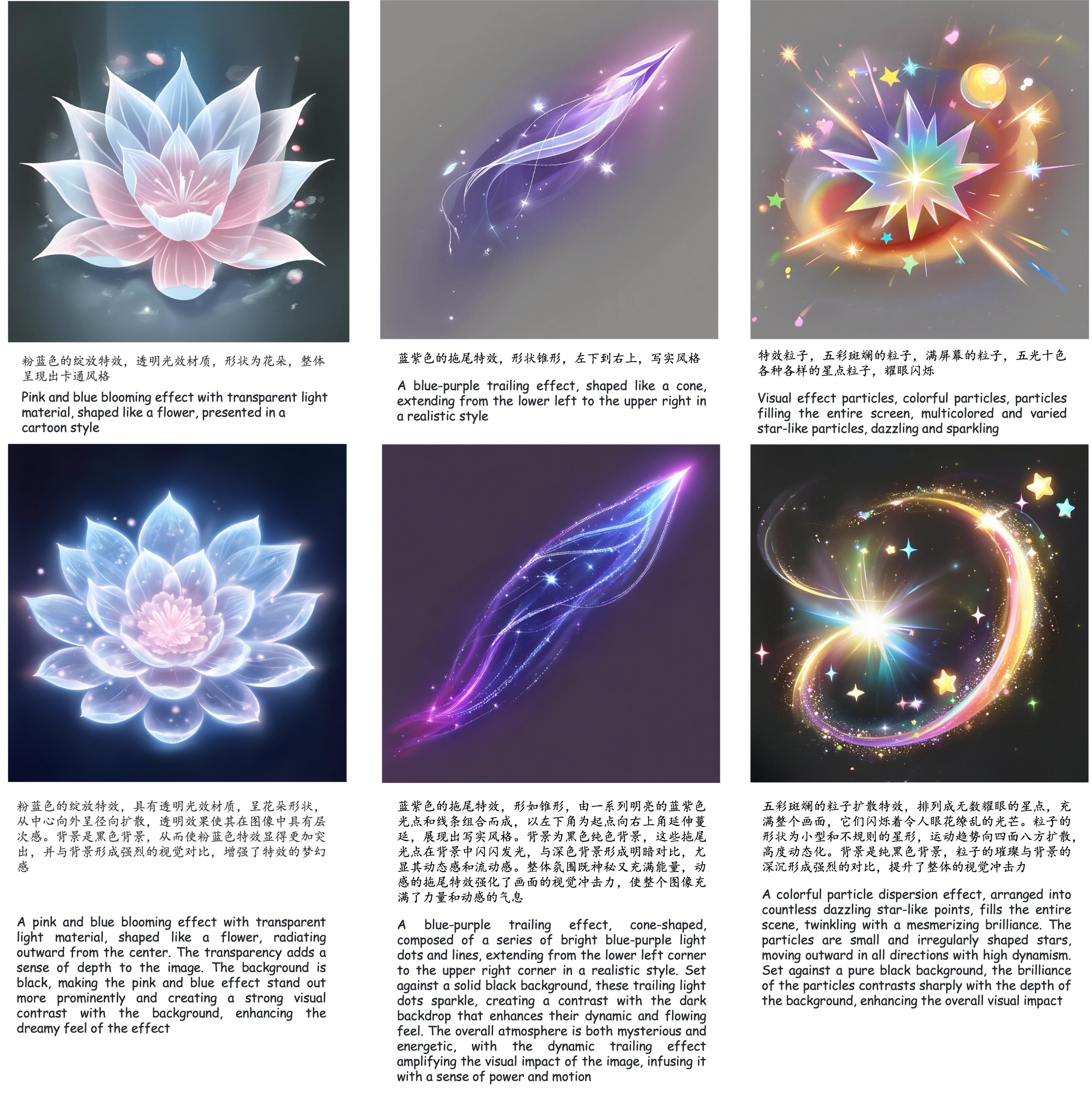}
    \caption{Rewritten descriptions can significantly enhance the details and texture of generated images. \textbf{Top row}: w/o prompt rewriting. \textbf{Bottom row}: w/ prompt rewriting.}
    \label{fig:t2v_e_recap}
\end{figure}
\subsubsection{Method}
In the design of the visual effects text-to-image generation scheme, optimizations are made around the caption system, training strategy, and inference mechanism.

{\bf Caption}

In the dimension of caption construction, a diverse system comprising brief descriptions, detailed descriptions, and comprehensive descriptions is established, as shown in ~\Cref{fig:caption-example}. During training, a random proportional sampling strategy is employed to enhance the generalization ability of the image generation model through diverse text inputs.

$\bullet$ Brief Description: Combining the six feature dimensions of visual effects into phrases.

$\bullet$ Detailed Description: Expressing the brief description in natural language while adding general information such as background color to comprehensively describe the visual effects image.

$\bullet$ Comprehensive Description: Building on the detailed description with more precise shape descriptions and overall visual effects atmosphere, aiming to accurately depict visual effects details.

{\bf Training}

% For the training method, a multi-stage hierarchical training mechanism is adopted, sequentially training on effect-enhanced material data, progressively augmented data, and high-quality visual effects data to gradually enhance the texture and realism of visual effects generation.
Our training approach employs a multi-stage hierarchical mechanism, sequentially utilizing effect-enhanced material data, progressively augmented data, and high-quality visual effects data to progressively enhance the texture detail and realism in visual effects generation.

% $\bullet$ Phase 1: Pre-training on effect-enhanced material data enables the model to perceive visual effects concepts while maintaining the base model's conceptual understanding.
$\bullet$ Phase 1: Pre-training on effect-enhanced material data allows the model to grasp fundamental visual effects concepts while retaining the base model’s prior knowledge.

% $\bullet$ Phase 2: Training on progressively augmented data with better visual effects that align more closely with the domain distribution of visual effects significantly enhances the texture of visual effects.
$\bullet$ Phase 2: Training on progressively augmented data, which better aligns with the target domain distribution, significantly improves texture representation and detail synthesis.

% $\bullet$ Phase 3: High-quality visual effects data is used as the original data for quality-tuning to improve the aesthetic quality of the final generated visual effects.
$\bullet$ Phase 3: Quality-tuning with high-quality visual effects data further enhances the aesthetic fidelity and overall visual quality of the generated results.

{\bf Inference}

% In the inference stage, a dedicated rewriting model is designed to transform the user's original prompt into an information-rich, comprehensive description, aligning user input with the training domain. By reinforcing the precise depiction of visual effects details, a regularization constraint on the model output is formed, significantly optimizing the quality of generated visual effects (as shown in ~\Cref{fig:t2v_e_recap}), achieving full-chain optimization from data annotation, model training to inference application.
During inference, we design a dedicated prompt rewriting model that transforms the user’s original input into an information-rich and comprehensive description, effectively aligning it with the training domain. By enforcing a regularization constraint that emphasizes precise depiction of visual effects details, the model output quality is significantly enhanced (see ~\Cref{fig:t2v_e_recap}). This approach enables full-pipeline optimization spanning data annotation, model training, and inference deployment.

\begin{figure}[!h]
    \centering
    \includegraphics[width=1.0\textwidth]{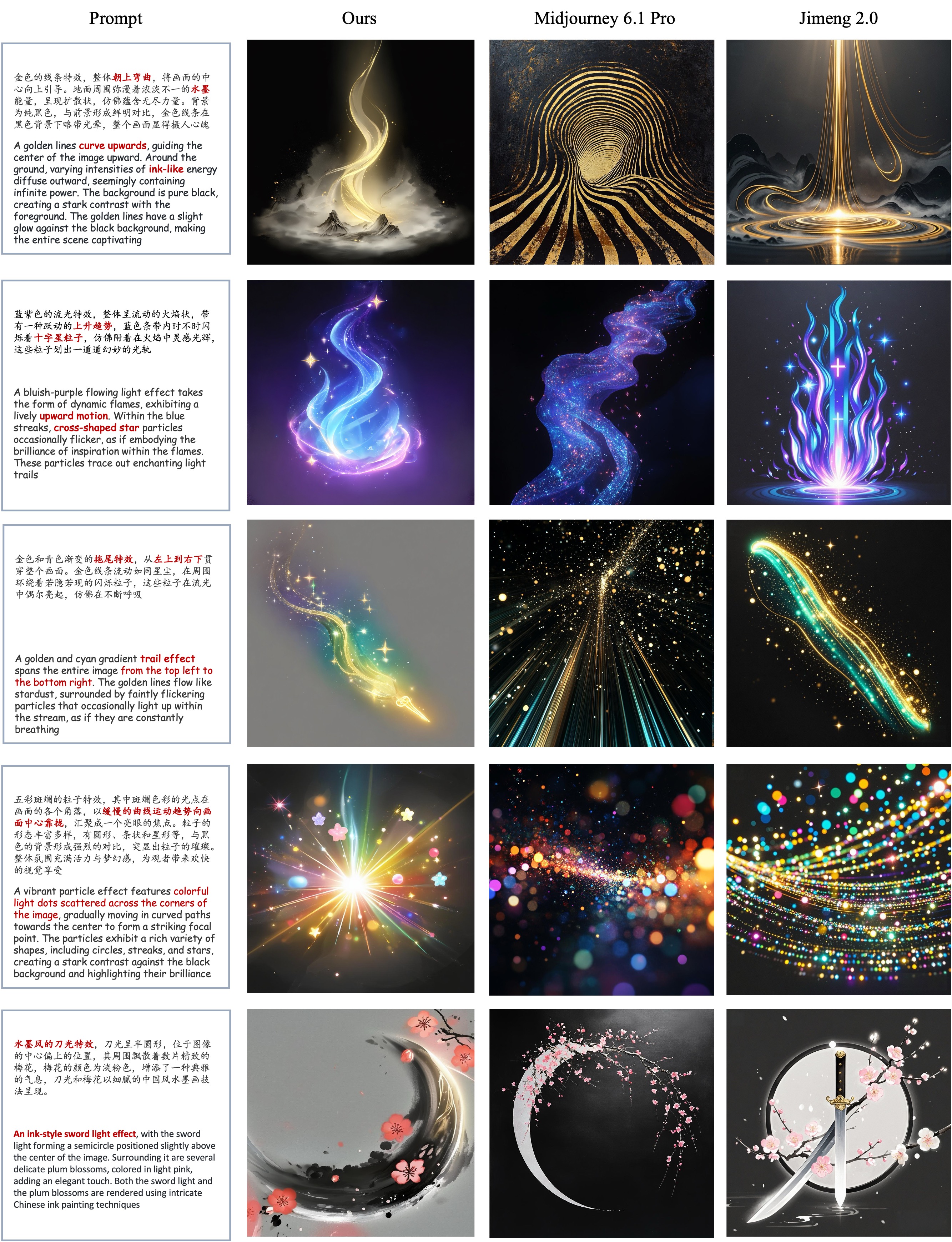}
    \caption{Qualitative comparisons with State-of-The-Art methods, \ie, Midjourney 6.1 Pro and Jimeng 2.0.}
    \label{fig:t2e_result_v2}
\end{figure}
\begin{figure}[!h]
    \centering
    \includegraphics[width=0.85\textwidth]{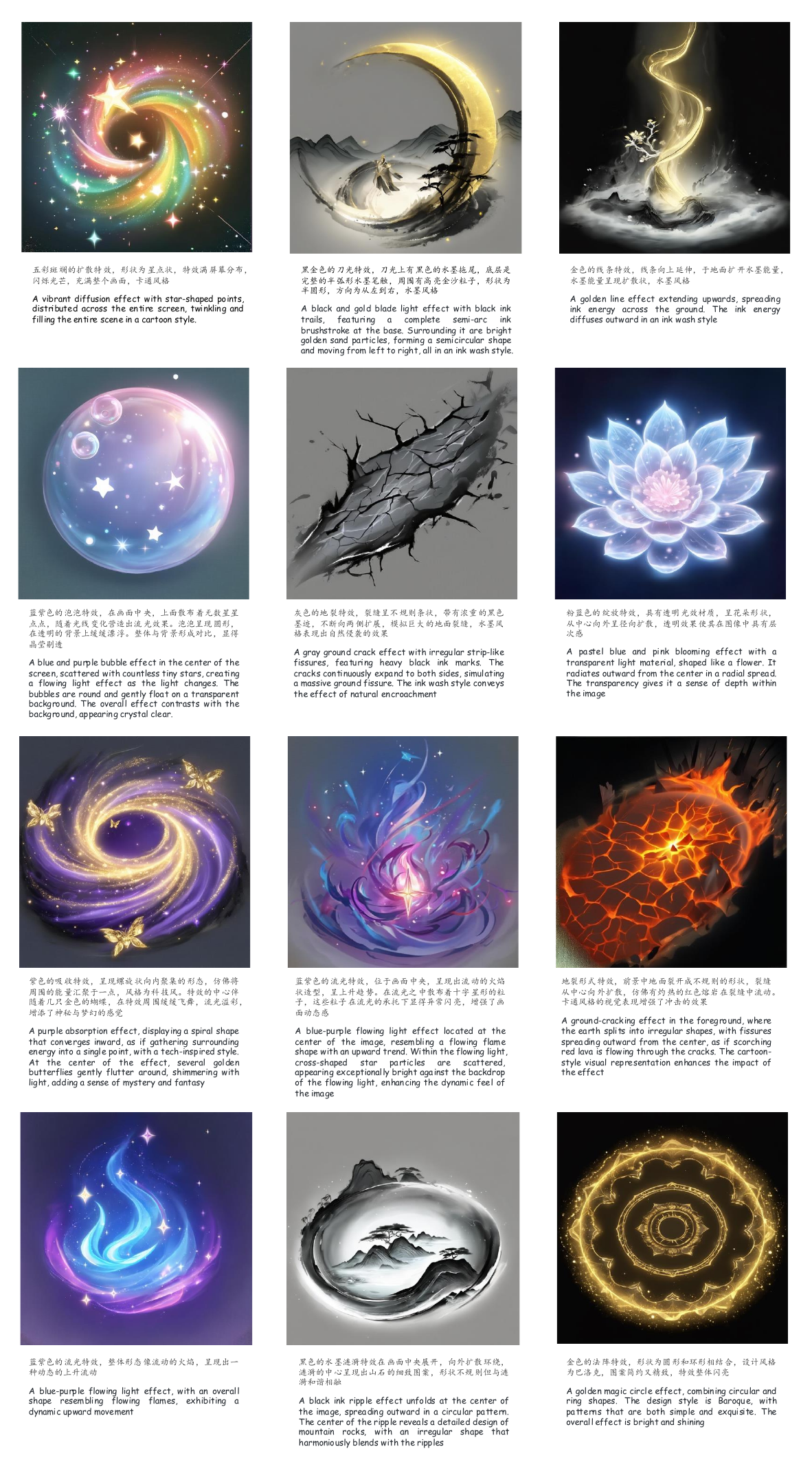}
    \caption{Visualizations of text-to-game visual effects generation results.}
    \label{fig:t2v_e_result}
\end{figure}

\subsubsection{Evaluation}
% This is the first systematic text-to-game visual effects generation that employs gamified language. The original base model is trained in Chinese, demonstrating strong recognition capabilities of Chinese elements, as shown in Fig~\ref{fig:t2v_e_result}. 
To the best of our knowledge, this work represents the first systematic approach to text-to-game visual effects generation that explicitly incorporates gamified language as a core component. The underlying base model is originally trained on Chinese-language data, exhibiting strong recognition and understanding capabilities of Chinese-specific elements and semantics. This linguistic specialization enables more accurate and contextually relevant generation of visual effects tailored to Chinese game design scenarios. By comparing our model with Midjourney 6.1 Pro and JiMeng 2.0, it can be observed that our model demonstrates significantly better semantic adherence compared to Midjourney 6.1 Pro. Additionally, when compared to JiMeng 2.0, our model exhibits superior texture quality in visual effects. Consequently, our model shows enhanced performance in visual effects scenarios, as demonstrated in Fig.~\ref{fig:t2e_result_v2}.

% In terms of special effects texture, it surpasses conventional general text-to-image base models, and can even be directly used in the designer's production process, providing inspiration for design and material production.
From the perspective of visual effects texture quality, our model significantly outperforms conventional text-to-image generation models. It is capable of generating highly detailed and visually coherent effects textures that meet the practical requirements of game production pipelines, as demonstrated in Fig.~\ref{fig:t2v_e_result}. Notably, the generated outputs are of sufficient quality to be directly integrated into designers’ workflows, serving not only as final assets but also as valuable sources of inspiration for creative design and material development. This practical applicability underscores the potential of our approach to bridge the gap between AI-driven generation and real-world game content creation.

%% file: sections/image/3_image2texiao.tex
\subsection{Image-to-\textit{Game Visual Effects}}
\subsubsection{Introduction}
Text-to-Game visual effects technology is a crucial tool in visual effects production. However, in advanced creative processes, game developers and designers demand higher precision in visual effects generation, necessitating precise control over effect details, dynamic changes, and visual styles. The current limitations of text-to-game visual effects technology are evident in two main areas. Firstly, models struggle to achieve precise control over generated content, leading to discrepancies between the final effects and design expectations. Secondly, designers' needs for control intensity over generated images vary dynamically across different creative stages, such as concept ideation and detail optimization, which current text-to-game visual effects cannot effectively address. Additionally, the realism and artistic expressiveness of effect materials (such as the glossy texture of metal or the detailed texture of flames) pose stringent challenges to text-to-game visual effects technology, with current solutions falling short of achieving high-quality material presentation.

This study proposes a systematic solution based on a text-to-game visual effects model as the foundational model. By designing capabilities such as (1) \textbf{black-and-white draft} control generation, (2) \textbf{color sketch} control generation, and (3) \textbf{black sketch} control generation, a hierarchical and multi-dimensional image generation control system is constructed, which can flexibly adapt to the control intensity needs of different creative stages. Simultaneously, innovative \textbf{material transfer capabilities} are introduced, utilizing a material transfer model to achieve precise capture and transfer of material textures and visual styles, thereby effectively enhancing the quality of generated effect materials. Our study provides new insights and methods for developing game visual effects technology, advancing game visual effects creation toward greater precision and efficiency.
\subsubsection{Data}
In the research of controllable game visual effect generation, data construction is a core component for achieving precise image generation control and high-quality material transfer. Given the significant differences in control prior strength among black-and-white draft control, color sketch control, and black sketch control, this study designs different data collection and processing strategies for different strengths of control signal.

\begin{figure}[!h]
    \centering
    \includegraphics[width=0.95\textwidth]{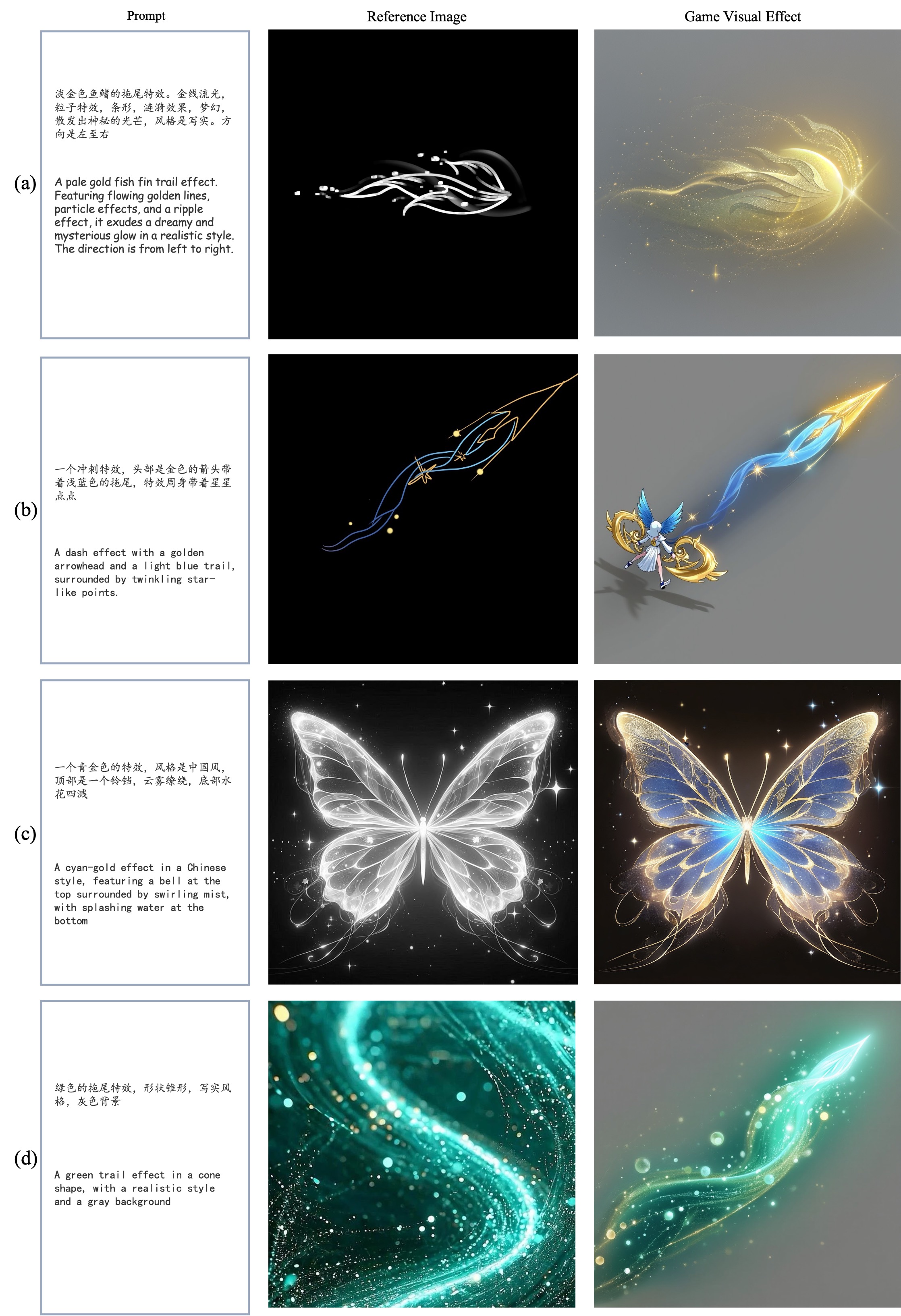}
    \caption{Game effects generation results based on (a) black sketch control, (b) color sketch control, (c) black-and-white draft, and (d) material transfer.}
        \label{fig:black_and_white_draft_control}
\end{figure}

\begin{figure}[!h]
    \centering
    \includegraphics[width=0.9\textwidth]{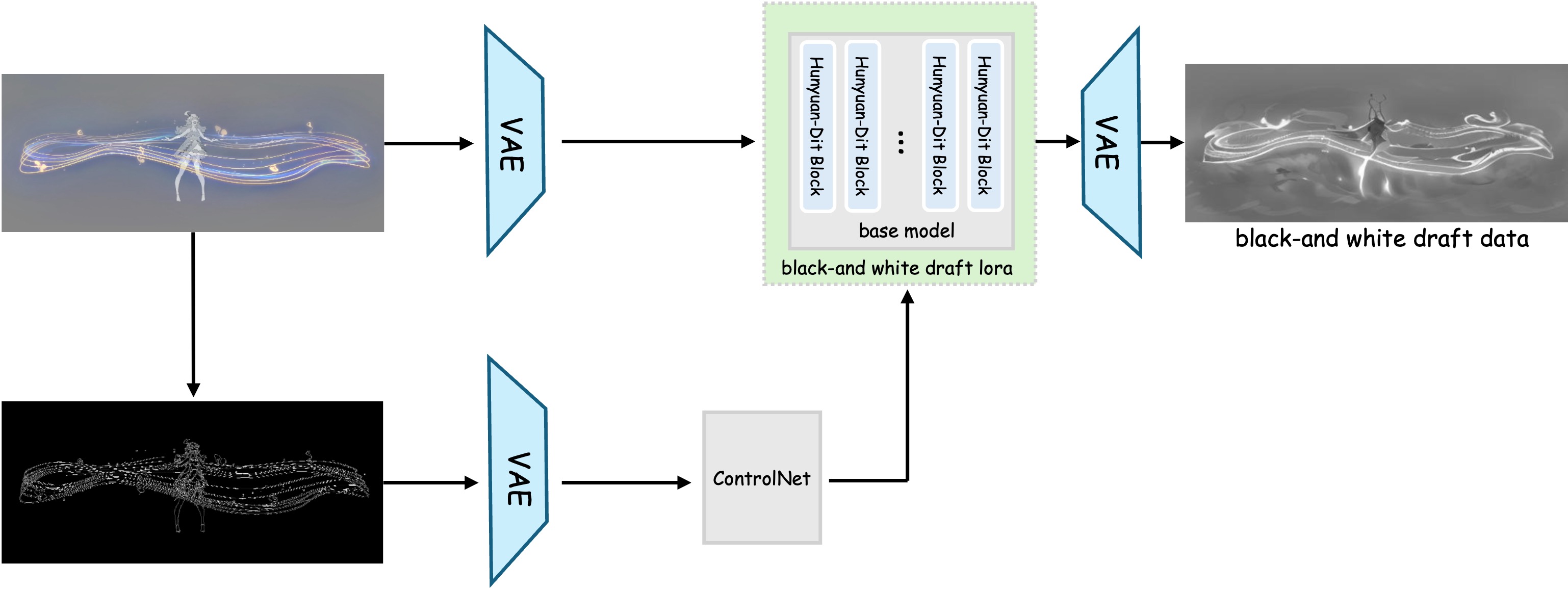}
    \caption{The pipeline of black-and-white draft generation.}
        \label{fig:black-and-white-draft-control-pipeline}
\end{figure}

% $\bullet$ Low-intensity control (black sketch control): As shown in~\Cref{fig:black_and_white_draft_control} (a), as the least constrained method, black sketch control employs a `general data pre-training + high-quality data fine-tuning' strategy. Combined with a scribble control preprocessing method, pre-training is conducted on a million-level general dataset to ensure effective model fitting to control inputs. Fine-tuning on a ten-thousand-level high-quality dataset ensures the control model can fit the texture of effects.
$\bullet$ Low-intensity control (black sketch control): As shown in~\Cref{fig:black_and_white_draft_control}(a), black sketch control, the least constrained method, employs a two-stage strategy: general data pre-training followed by high-quality data fine-tuning. Using a scribble control preprocessing technique, pre-training is performed on a dataset of approximately one million samples to ensure effective model adaptation to control inputs. Fine-tuning on a high-quality dataset of around ten thousand samples enables the model to accurately capture effect textures.

$\bullet$ Medium-intensity control (color sketch control): As shown in~\Cref{fig:black_and_white_draft_control} (b), color sketch control relies on a ten-thousand-level dataset of professional design data. This involves hiring experienced designers to manually draw color control images corresponding to visual effects images, constructing high-precision training data pairs.

% $\bullet$ High-intensity control (black-and-white draft control): As shown in Fig.~\Cref{fig:black_and_white_draft_control} (c), for black-and-white draft control, due to its strong prior constraints, only a few thousand data samples are needed to meet the training requirements. This portion of the data is primarily generated by a LoRA model. Initially, a black-and-white draft LoRA is trained using a small amount of data. Then, utilizing this LoRA model, control images are produced using visual effects data as the base image, employing an image-to-image approach to construct image-control condition data pairs, as is shown in~\Cref{fig:black-and-white-draft-control-pipeline}.
$\bullet$ High-intensity control (black-and-white draft control): As illustrated in Fig.~\Cref{fig:black_and_white_draft_control} (c), the black-and-white draft control benefits from strong prior constraints, enabling effective training with only a few thousand data samples. This subset of data is predominantly generated via a LoRA model. Specifically, a black-and-white draft LoRA is first trained on a limited dataset. Subsequently, leveraging this LoRA, control images are synthesized by applying an image-to-image translation approach, using visual effects data as base images to construct paired image-control condition datasets, as depicted in~\Cref{fig:black-and-white-draft-control-pipeline}.

% \begin{figure}[!h]
%     \centering
%     \includegraphics[width=0.9\textwidth]{figures/color sketch control.pdf}
%     \caption{Game effects generation based on color sketch control.}
%         \label{fig:color_sketch_control}
% \end{figure}

% \begin{figure}[!htbp]
%     \centering
%     \includegraphics[width=0.9\textwidth]{figures/black sketch control.pdf}
%     \caption{Game effects generation based on black sketch control.}
%         \label{fig:black_sketch_control}
% \end{figure}

% $\bullet$ Material transfer: As shown in ~\Cref{fig:black_and_white_draft_control} (d), the material transfer task also uses a `general data pre-training + high-quality data fine-tuning' strategy. Pre-training on a million-level general dataset is used to fit general material transfer capabilities. Based on a thousand-level PSD layered dataset, leveraging its layer separation characteristics, visual effect layers are combined with diverse backgrounds. Through data augmentation, the training data scale is expanded to hundreds of thousands, providing ample data support for precise material feature transfer.
$\bullet$ Material transfer: As shown in~\Cref{fig:black_and_white_draft_control}~(d), employs a two-stage strategy: general data pre-training followed by high-quality data fine-tuning. Pre-training on a million-sample-level general dataset establishes general material transfer capabilities. Leveraging a thousand-level Photoshop Document~(PSD) layered dataset and its layer separation properties, visual effect layers are combined with diverse backgrounds. Data augmentation expands the training set to hundreds of thousands of samples, providing sufficient data to support precise material feature transfer.

% \begin{figure}[!htbp]
%     \centering
%     \includegraphics[width=0.9\textwidth]{figures/Material transfer.pdf}
%     \caption{\textbf{Image-to-Game Visual Effects}. material transfer}
%         \label{fig:Material_transfer}
% \end{figure}

The above data strategies fully leverage the advantages of different control methods, achieving an optimal balance between data volume and control precision, laying a solid foundation for enhancing the performance of effect generation models.

\subsubsection{Method}
This section proposes different technical approaches for the two core tasks of controllable image generation and material transfer, based on data characteristics and task requirements.

% \textbf{Controllable Effects Generation.} In the process of training a control model, the richer the signals provided by the control conditions, meaning the more priors available, the faster the control model converges, and correspondingly, the less control data is required. Black-and-white draft control provides very detailed transition information, thus offering the most prior information and requiring the least amount of control pair data. Color sketch control, although a form of sketch control, includes color priors, so the amount of training pair data needed is moderate. Finally, black sketch control requires the most training pair data. Due to significant differences in data source characteristics and data volume among  black-and-white draft control, color sketch control, and black sketch control (ranging from thousands to millions of data points), independent training strategies are employed to construct each image generation control model separately to maximize model performance and avoid training bias caused by data mixing. Black sketch control, with weaker control priors, requires more training data and is specifically designed with a two-stage training process: a pre-training stage using million-level general data to fit basic control signals, followed by a post-training stage using ten-thousand-level fine-tuning data to enhance the texture of visual effects generation.
\textbf{Controllable Effects Generation.} During the training of a control model, richer control signals—i.e., more available priors—accelerate model convergence and reduce the required amount of control data. Black-and-white draft control provides highly detailed transition information, offering the most prior knowledge and thus requiring the least amount of paired training data. Color sketch control, while still a form of sketch control, incorporates color priors, resulting in a moderate need for training pairs. In contrast, black sketch control demands the largest volume of training data. Due to substantial differences in data source characteristics and scale—ranging from thousands to millions of samples—among black-and-white draft control, color sketch control, and black sketch control, independent training strategies are adopted to build each image generation control model separately. This approach maximizes model performance and prevents training bias caused by data mixing. Black sketch control, which has weaker control priors, employs a two-stage training process: a pre-training phase on a general dataset of approximately one million samples to learn basic control signals, followed by a fine-tuning phase on a dataset of around ten thousand samples to enhance the texture quality of visual effects generation.

% \textbf{Material Transfer.} this study employs an IP-Adapter-like architecture combined with a 'pre-training + fine-tuning' two-stage training strategy to build an efficient material transfer model. The pre-training stage uses large-scale data to enable the model to learn general material features, while the fine-tuning stage optimizes model parameters for game visual effects scenarios, improving the accuracy of material transfer. During inference, designers are innovatively supported to achieve precise generation of visual effects in local areas through hand-drawn attention masks, meeting diverse creative needs. This approach balances the model's generalization capability with the flexibility of visual effects creation, providing an effective technical solution for game visual effects material generation.
\textbf{Material Transfer.} This study employs an IP-Adapter-like architecture combined with a two-stage training strategy of pre-training and fine-tuning to develop an efficient material transfer model. The pre-training stage leverages large-scale data to enable the model to learn general material features, while the fine-tuning stage optimizes model parameters specifically for game visual effects scenarios, enhancing the accuracy of material transfer. During inference, designers are innovatively supported in achieving precise generation of visual effects in local areas through hand-drawn attention masks, thereby meeting diverse creative needs. This approach balances the model’s generalization capability with the flexibility required for visual effects creation, providing an effective technical solution for game visual effects material generation.

% In practical inference, by organically integrating image generation control capabilities with material transfer capabilities, diverse and high-precision visual effects control generation can be achieved, satisfying the dual demands for visual effect diversity and detail realism in game visual effects creation.
In practical inference, the seamless integration of image generation control and material transfer capabilities enables diverse and high-precision visual effects generation, meeting the dual demands of visual effect variety and detailed realism in game visual effects creation.
\begin{figure}[htbp]
    \centering
    \includegraphics[width=0.9\textwidth]{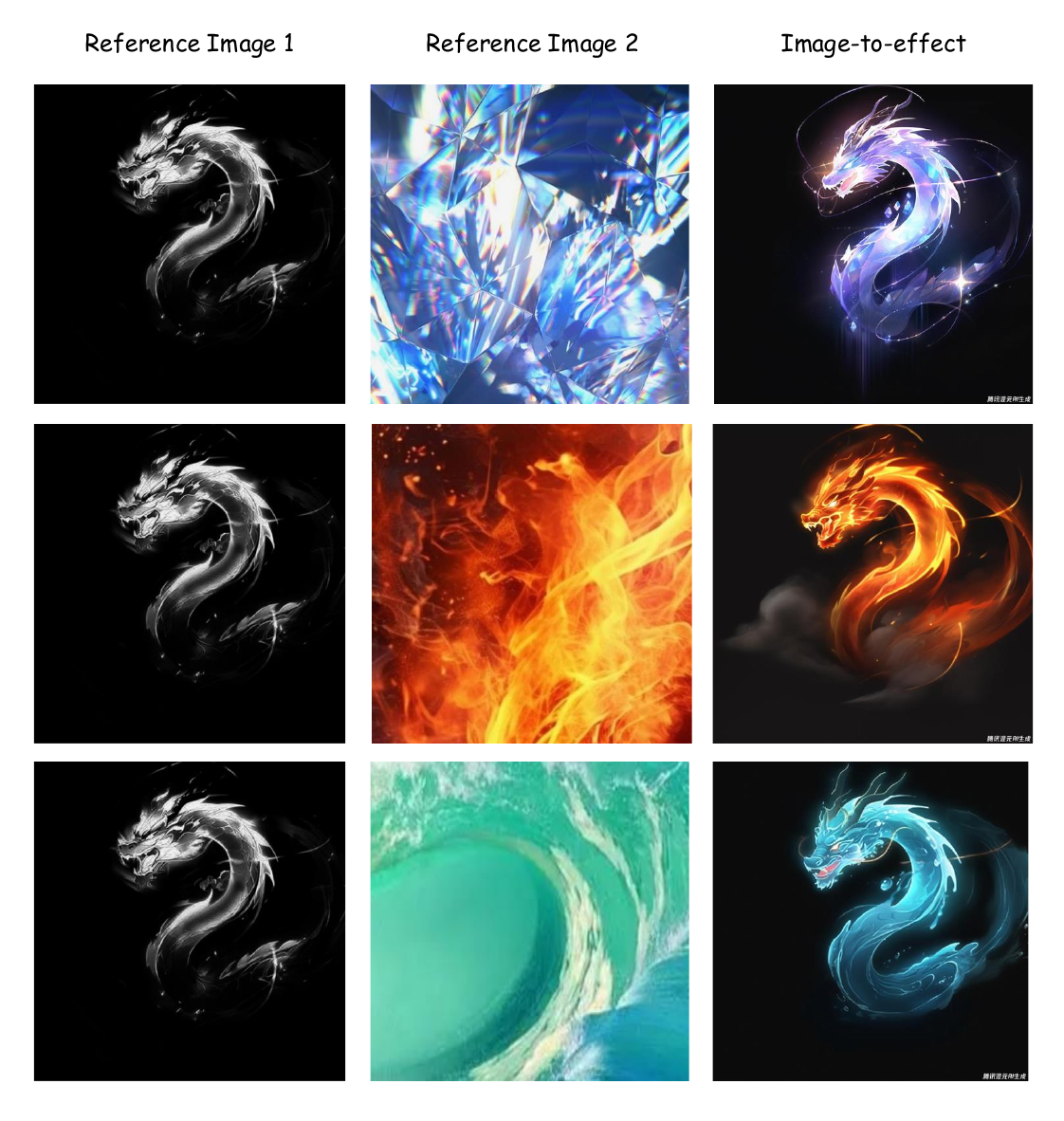}
    \caption{Visualizations of image-to-game visual effects generation results.}
        \label{fig:i2v_effect_control_material}
\end{figure}
\subsubsection{Evaluation}
This is the first series of reference-based game visual effects generation, providing a comprehensive solution for text-to-game visual effects by combining image generation control capabilities with material transfer capabilities. Based on feedback from designers who have used the model in practice, it has improved the efficiency of visual effects iteration by 60\%.

%% file: sections/image/4_transparent.tex
\subsection{Transparent and Seamless Image Generation}
% @jianxiang
\subsubsection{Introduction}
In the field of game design creation, designers frequently rely on base materials for their work and use layered drawing techniques to achieve various effects. Many special effects in games, such as shadows, glows, and explosions, depend on transparent image materials. Transparent backgrounds allow developers to stack different elements on multiple layers, such as interaction effects between objects and the environment, or the combination of characters and weapons. Additionally, seamless textures are widely used in constructing scenes and large-scale environments, ensuring natural transitions and repetitive use of textures while optimizing performance.

% We have developed the capability to generate transparent images and extended it to the application of seamless textures. Our system can generate transparent images for various categories, including characters, props, scenes, and special effects. Additionally, we can convert user-provided transparent textures into seamless transparent textures. These capabilities not only enhance the quality and efficiency of the creative process but also reduce the cumbersome manual design work.
Our system generates transparent images and supports seamless texture synthesis. Our approach can produce transparent images across a range of categories, including characters, props, scenes, and special effects. Furthermore, it enables the conversion of user-provided transparent textures into seamless transparent textures. These capabilities not only improve the quality and efficiency of the creative process but also significantly reduce the need for labor-intensive manual design work.

% To achieve exceptional transparent image capabilities in the gaming domain, we have tailored the training process on the DiT game foundation model and established a high-quality data collection pipeline specifically designed for game features. This section will provide a detailed overview of the algorithms and data construction involved in building this capability.
To achieve exceptional transparent image capabilities in the gaming domain, we tailored the training process on the DiT game foundation model and established a high-quality data collection pipeline specifically designed for game features. This section provides a detailed overview of the algorithms and data construction process underlying this capability.

\subsubsection{Data}
% In the gaming field, the collection of transparent image data faces two primary challenges. First, compared to RGB image data, transparent image data is much rarer, especially in specific verticals such as high-quality transparent images in game special effects, which are particularly difficult to collect. Second, the quality of transparent images can be inconsistent. Many of these images are obtained through cutting algorithms, which often result in noticeable edge blurring and artifacts, affecting their overall quality.
In the gaming domain, collecting transparent image data presents two major challenges. First, such data is significantly scarcer than standard RGB images, particularly in specialized areas like high-quality assets for game special effects, where transparent images are especially difficult to obtain. Second, the quality of transparent images is often inconsistent. Many are derived using cutout algorithms, which frequently introduce edge blurring and artifacts, thereby degrading overall quality.

To address these challenges, we designed a multi-level data construction pipeline to accommodate subsequent training strategies and the generation of transparent images in the special effects domain.

\begin{itemize}
% \item[$\bullet$] \textbf{Extensive data collection.} At this level, our data collection is not limited to a specific domain but focuses on the large-scale gathering of transparent images that meet certain criteria to enhance the model's ability to recognize such images. All collected images have a resolution of at least 1024x1024. To ensure data quality, we perform an initial screening using algorithms such as aesthetic scoring and logo detection. Additionally, we have developed a custom filtering logic based on the subject-to-background ratio and transparency effects, allowing us to further select images with high-quality edges and clear, well-defined subjects.
\item[$\bullet$] \textbf{Extensive data collection.}~At this stage, our data collection is not confined to a specific domain but instead emphasizes the large-scale acquisition of transparent images that meet defined quality criteria to enhance the model’s ability to recognize such content. All collected images have a minimum resolution of 1024×1024. To ensure data quality, we apply an initial screening using algorithms such as aesthetic scoring and logo detection. Furthermore, we employ a custom filtering strategy based on the subject-to-background ratio and transparency effects, enabling the selection of images with high-quality edges and clear, well-defined subjects.

% \item[$\bullet$] \textbf{High-quality game domain data.} Furthermore, we have collected high-quality transparent image data from outstanding games to enhance the model's performance in the gaming domain. We have gathered game data from categories such as character portraits, characters, and props, all of which exhibit high clarity, aesthetic appeal, and excellent composition. Additionally, we have accumulated approximately 40,000 3D models of game characters, which can be rendered into a large number of transparent images through various actions. To date, we have collected around 300,000 transparent image data samples.
\item[$\bullet$] \textbf{High-quality game domain data.} 
Furthermore, we collected high-quality transparent image data from leading games to further improve the model’s performance in the gaming domain. This dataset includes categories such as character portraits, full-body characters, and props, all characterized by high clarity, strong aesthetic appeal, and well-composed layouts. Furthermore, we have collected approximately 40,000 3D character models, which can be rendered into a large volume of transparent images across diverse actions and poses. To date, we have collected around 300,000 transparent image data samples.

\item[$\bullet$] \textbf{Rejection sampling of special effects transparent image data.} 
% Game special effects transparent image data is one of the rarest and most difficult-to-create data categories, making its expansion a significant challenge. To address this challenge, we employed a rejection sampling approach to construct the dataset. First, we extracted individual layers from high-quality special effects PSD files and used a certain level of automation to separate the effect layers, thus obtaining the initial data. Next, we fine-tuned a base transparent image generation model using these initial special effects data. The fine-tuned model was then used to generate candidate special effects transparent images, which were manually screened for quality. Finally, we mixed the selected images with the initial data and refined the model again, iterating the process multiple times to ultimately build a large-scale, high-quality special effects transparent image dataset.
Transparent images of game special effects represent one of the rarest and most challenging data categories to produce, making their large-scale acquisition particularly difficult. To address this challenge, we adopted a rejection sampling strategy for dataset construction. We began by extracting individual layers from high-quality special effects PSD files and employed semi-automated methods to isolate effect layers, thereby generating the initial data. This initial set was then used to fine-tune a base transparent image generation model. The fine-tuned model produced candidate special effects images, which were manually screened for quality. The selected images were combined with the original dataset to further refine the model. This process was iterated multiple times, ultimately resulting in a large-scale, high-quality special effects transparent image dataset.

\end{itemize}
\begin{figure}[htbp]
    \centering
    \hspace{-1.6cm}
    \scalebox{1.1}{
    \includegraphics[width=1.0\textwidth]{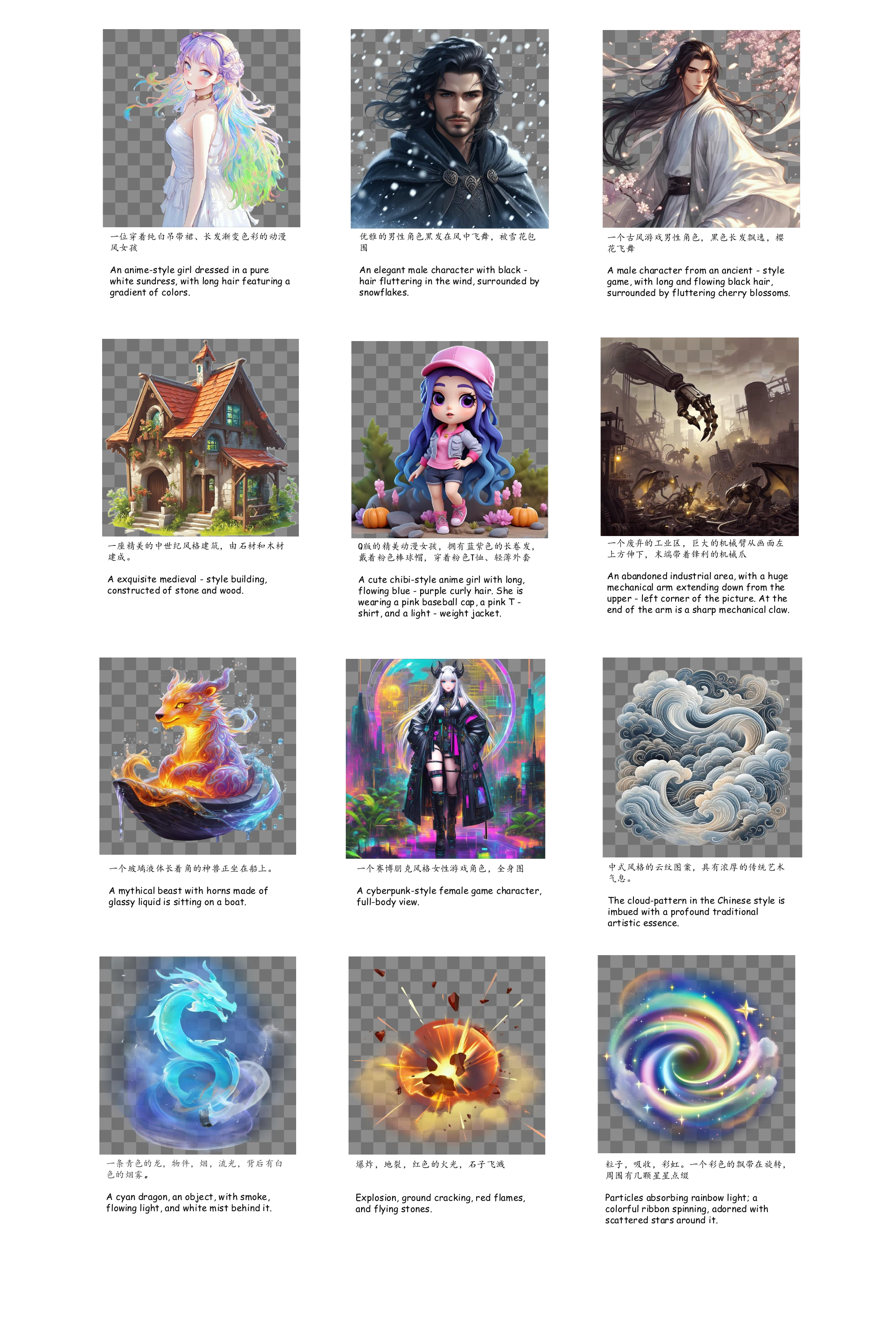}}
    \vspace{-1.6cm}
    \caption{Hunyuan transparent image generation}
    \label{fig:transparent_1}
\end{figure}

\begin{figure}[htbp]
    \centering
    % \hspace{-0.7cm}
    % \scalebox{1.1}{
    \includegraphics[width=1.0\textwidth]{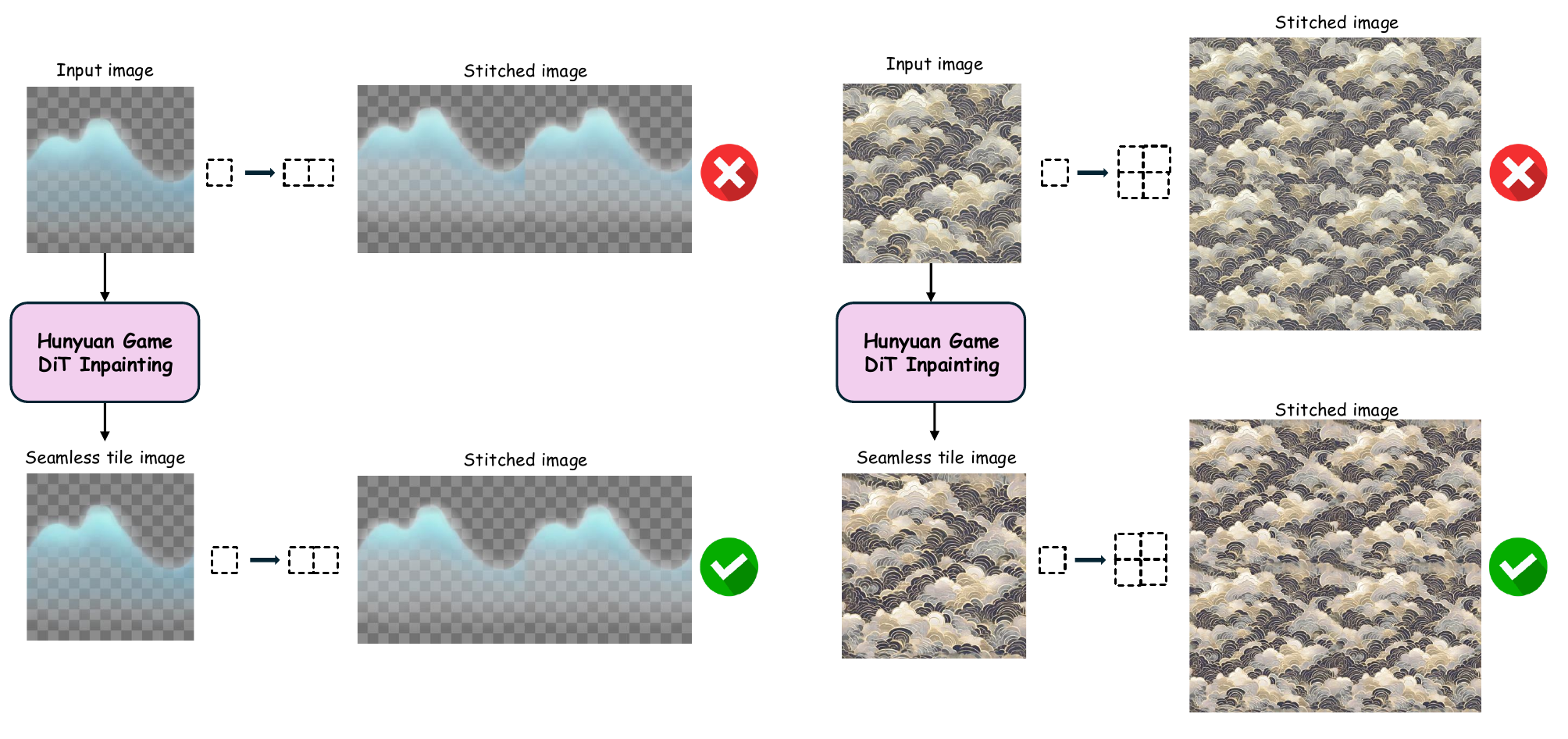}
    % }
    \caption{Seamless tile image generation. The first row represents the generation of seamless tile images in the horizontal direction. The second row represents the generation of seamless tile images in both the horizontal and vertical directions. The last column verifies the effectiveness of the method by stitching together the generated seamless tile images.} 
    \label{fig:seamless_1}
\end{figure}
\subsubsection{Method}

Based on game-based model knowledge, we developed algorithms for transparent image generation and seamless image generation. Inspired by Layer Diffuse \cite{zhang2024transparent}, we trained the VAE encoder and decoder within the DiT model. The encoder encodes transparency by generating offset values over the original latent space, thereby minimizing the impact on the original latent distribution, without the need to retrain the DiT model. The decoder then converts the generated latent into a four-channel image output. Next, we trained a large-parameter LoRA on the DiT model, enabling the model to generate latents with transparency offsets. Additionally, we found that increasing the number of training parameters effectively reduced the impact on the base model's generated content and style. The training process is mainly divided into three stages, each corresponding to the aforementioned three stages of data.

\textbf{Stage 1}: In this stage, we train the upstream DiT model of the game base model using the extensively collected training data, with a focus on enabling the model to learn transparent images from various categories. This allows the model to generate latents with transparency offsets.

\textbf{Stage 2}: We train the game base model using high-quality game domain data, helping the model better understand game-specific features, such as the setup of foreground and background, which directly influences which parts of the image need to be transparent. Through this stage, the model can generate high-quality game transparent images, especially excelling in handling object edges and foreground-background.

\textbf{Stage 3}: This stage primarily focuses on game special effects scenes. Using the data obtained through rejection sampling, we fine-tune the model for different game scenarios. Additionally, we constructed a special effects tagging system, annotating the special effect data with tags and specific descriptions, thereby improving the model's ability to distinguish between various types of special effects. In practical applications, designers can generate transparent images that meet their expectations by using these tags and descriptions.

% After building the capability for transparent image generation, we extended this to the application of seamless texture mapping. The definition of seamless texture mapping is that multiple images can be seamlessly stitched together in a specified direction. Thus, seamless textures can be classified into horizontal, vertical, and square seamless textures. Taking horizontal seamless texture mapping as an example, our goal is to input a transparent texture and generate a new texture with minimal differences from the original image, while ensuring the texture seamlessly and naturally tiles in the horizontal direction.
After developing the capability for transparent image generation, we extended it to seamless texture mapping. Seamless texture mapping refers to the process of stitching multiple images together seamlessly along a specified direction. Accordingly, seamless textures can be categorized as horizontal, vertical, or square. For example, in horizontal seamless texture mapping, the objective is to input a transparent texture and generate a new texture that closely resembles the original while ensuring it tiles seamlessly and naturally in the horizontal direction.

To achieve this, we made improvements primarily during the inference phase. First, we evenly split the input image into two parts, swapped their positions, and created a new image. Next, we only need to make the horizontal middle region of the image continuous. To accomplish this, we introduced adjustable parameters to control the width of this middle region, and implemented an inpainting algorithm
(e.g., BLD~\cite{avrahami2023blended})
% (e.g., Blended Latent Diffusion~\cite{avrahami2023blended}, BLD) 
on the transparent image generation model. By performing inpainting on the middle region and then restoring the left and right sections, we can obtain a horizontally seamless texture. The same approach can be applied to seamless textures in other directions.

\subsubsection{Evaluation}
Our transparent image generation algorithm performs excellently in the gaming domain, as shown in Fig \ref{fig:transparent_1}. In the character category, the algorithm is capable of generating characters that align with different game styles, particularly excelling in the handling of hair and clothing, producing clear edges and effectively generating blurred foregrounds and backgrounds. Additionally, the algorithm can generate game scene images with varying transparency effects across different regions.

Furthermore, the algorithm also performs outstandingly in the game special effects domain, effectively integrating multiple concepts through label prompts and specific descriptions to generate special effects transparent images.

In Fig.\ref{fig:seamless_1}, we demonstrate the capability of seamless tile image generation. The input images are generated by a transparent image model. It can be observed that the generated transparent seamless tile images make minimal modifications to the original content and achieve excellent generation results for different types of tile images.

%% file: sections/image/5_character.tex
\subsection{Game Character Generation}
\subsubsection{Introduction}
In the game creation process, character design, as a core component, directly influences the game experience and market performance. With the deep integration of text-to-image (T2I) technologies in the game development domain, these AIGC tools become indispensable aids within the character design workflow. Current industry practices indicate that designers, when leveraging AIGC technologies, not only emphasize the fundamental efficiency of game character generation but also impose higher requirements on the accuracy and controllability of the generated outputs. In particular, in critical aspects such as character structural rationality and consistency, the industry urgently requires more precise technical control mechanisms to ensure comprehensive quality assurance throughout the entire process, from concept design to final implementation.

However, the current application of text-to-image technologies in game character generation~\cite{tao2025instantcharacter} faces multiple challenges. First, the model outputs often deviate significantly from designers' expectations, struggling to satisfy professional requirements in detail fidelity and creative alignment. Second, existing T2I models from open-source communities are predominantly general-purpose and lack vertical domain optimization tailored for game character design, resulting in insufficient generation quality at the professional level. Most notably, in character design scenarios, models encounter difficulties in maintaining consistency across character images, styles, and settings, representing a technical bottleneck that severely limits the practical value of AIGC technologies within the game character design workflow.

This study proposes a comprehensive controllable generation solution for game characters based on a DiT model fine-tuned with domain-specific game character data. It adopts a multistage progressive generation strategy, beginning with the conversion of lineart to grayscale image to establish the basic structure, followed by generating complete character images from grayscale sketches, thereby forming a clearly layered creative process that effectively addresses key technical bottlenecks in the industry. After completing the character concept design, the solution introduces a character consistency model that achieves consistent generation of characters in specific poses by analyzing character structure maps, including white model maps or depth maps. This technical approach not only enhances the controllability and professionalism of the generated results but also provides a standardized solution for character design in industrial-scale game production.
\subsubsection{Data}
In game character generation, data construction serves as a core technical foundation. Addressing significant differences in data requirements across the three modules, lineart to grayscale image generation, grayscale image to character image generation, and character consistency model, this study designs a multi-dimensional data engineering system.

\textbf{Data construction for Lineart to Grayscale image \& Grayscale image to Character Image.} For the two core modules, "lineart to grayscale image" and "grayscale image to character image," we construct high-quality datasets with millions of images. In the data pre-processing stage, a specialized feature extraction pipeline is designed. For the lineart to grayscale image model, advanced edge detection algorithms~\cite{zhang2023adding} are employed to extract refined lineart from the original character images as input. For the grayscale image to character image model, an optimized grayscale conversion pipeline is utilized to extract input features. To enhance the generalization capability of the model, multilevel data augmentation techniques, including input channel shuffling and color jitter, are applied prior to grayscale extraction.

\textbf{Data construction for Character Consistency Model.} To enhance model generalization and domain-specific performance, the dataset is systematically divided into two components: the former is employed as a large-scale pre-training dataset, while the latter is used for domain-specific fine-tuning. The pre-training phase utilizes a large-scale general dataset containing 30,000 real-world characters with millions of images, whereas the fine-tuning phase leverages a professional game domain dataset comprising 1,000 characters and hundreds of thousands of high-quality images. Data for each character are sampled from videos of the character. To further improve the model’s generalization in game character generation, character videos collected from leading game companies are incorporated during the quality tuning stage. During training, a random sampling strategy is adopted to select different image pairs of the same character as input-output combinations, with depth maps serving as the core structural control condition. To enhance the model’s generalization to depth maps extracted from white models, a specially designed data augmentation scheme is implemented: color space transformations such as channel shuffle and color jitter are applied to the original character images, while the extracted depth maps undergo blurring and grayscale transformation. These augmentations effectively improve the model’s robustness in maintaining character features under complex conditions.

\subsubsection{Method}
This study proposes a hierarchical technical system for game character generation, designing different solutions for three core tasks: lineart to grayscale image, grayscale image to character image, and game character consistency.

\begin{figure}[t]
    \centering
    \includegraphics[width=1.0\textwidth]{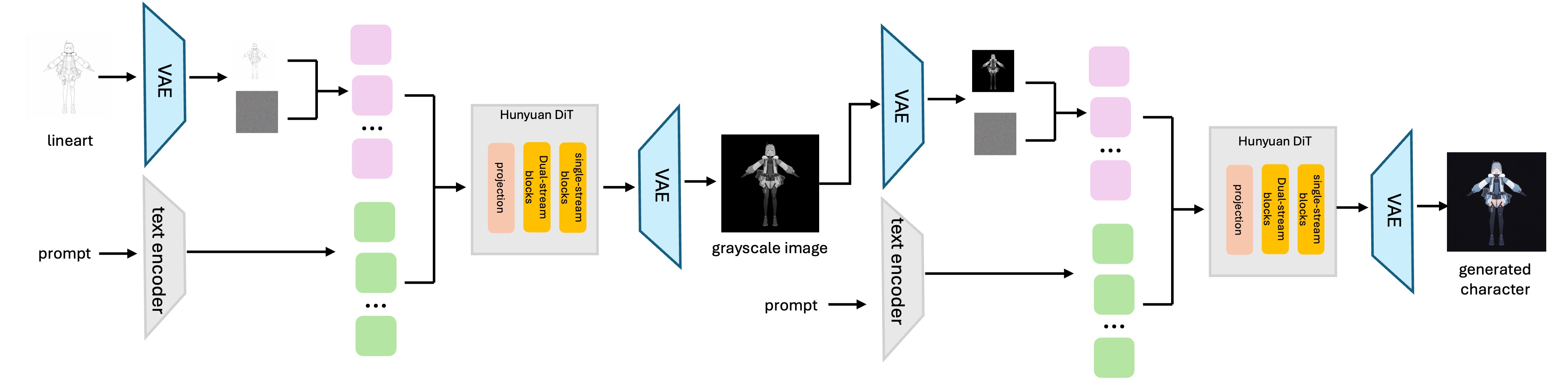}
    \caption{Method of Game Character Generation for lineart to grayscale image and grayscale image to character image.}
    \label{fig:character_control}
\end{figure}

{\bf Lineart $\to$ Grayscale $\to$ Character image.} For the tasks of lineart to grayscale image and grayscale image to character image, the control models are developed based on our game domain DiT ~\cite{peebles2023scalable} model, as shown in ~\Cref{fig:character_control}. The control conditions such as lineart or grayscale image is encoded into latents using the VAE, which are concatenated with the noisy latents to compose the input of the control models. Leveraging millions of high-quality data samples and specific data augmentation strategies, these models achieve excellent generalization capability. 

\begin{figure}[htbp]
    \centering
    \includegraphics[width=1.0\textwidth]{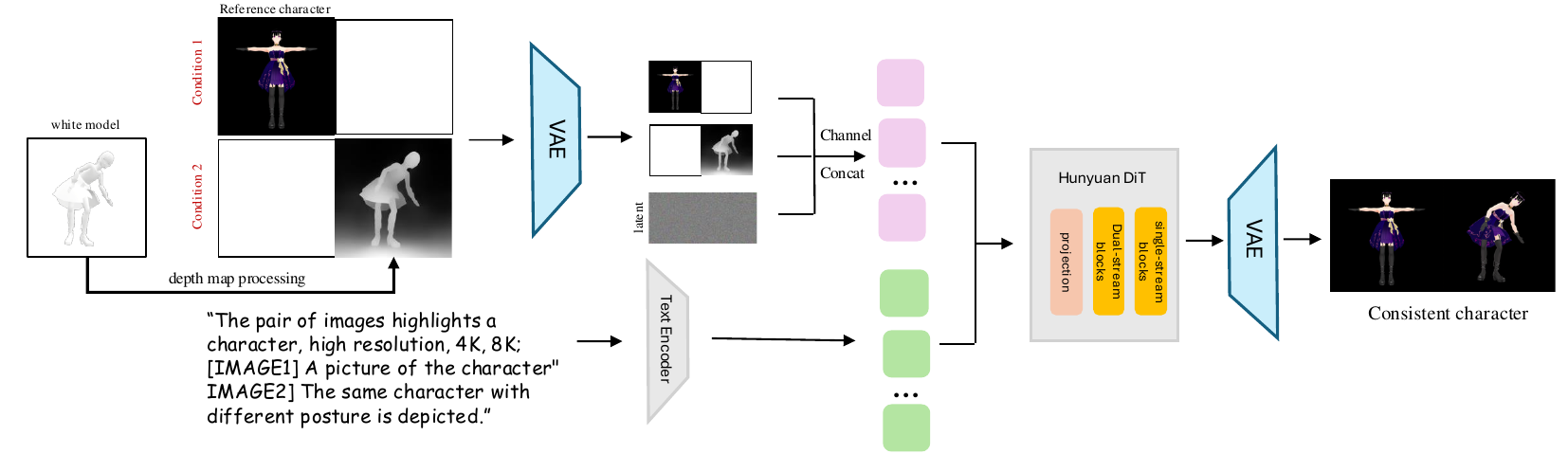}
    \caption{Method of Game Character Generation for character consistency.}
    \label{fig:character_consistency}
\end{figure}

{\bf Game Character Consistency.} Inspired by in-context LoRA~\cite{huang2024context}, the game character consistency model receives concatenated features composed of the VAE-encoded character reference image and depth map features, along with noisy latents, enabling end-to-end generation that preserves character traits and specific structural details, as shown in~\Cref{fig:character_consistency}. Notably, we adopt a two-stage training process consisting of general data pretraining followed by quality tuning with domain-specific data, which preserves the foundational model’s generalization ability while enhancing its specialized performance in the game domain, balancing the model’s universality and professional requirements.

\subsubsection{Evaluation}

\begin{figure}[htbp]
    \centering
    \includegraphics[width=0.85\textwidth]{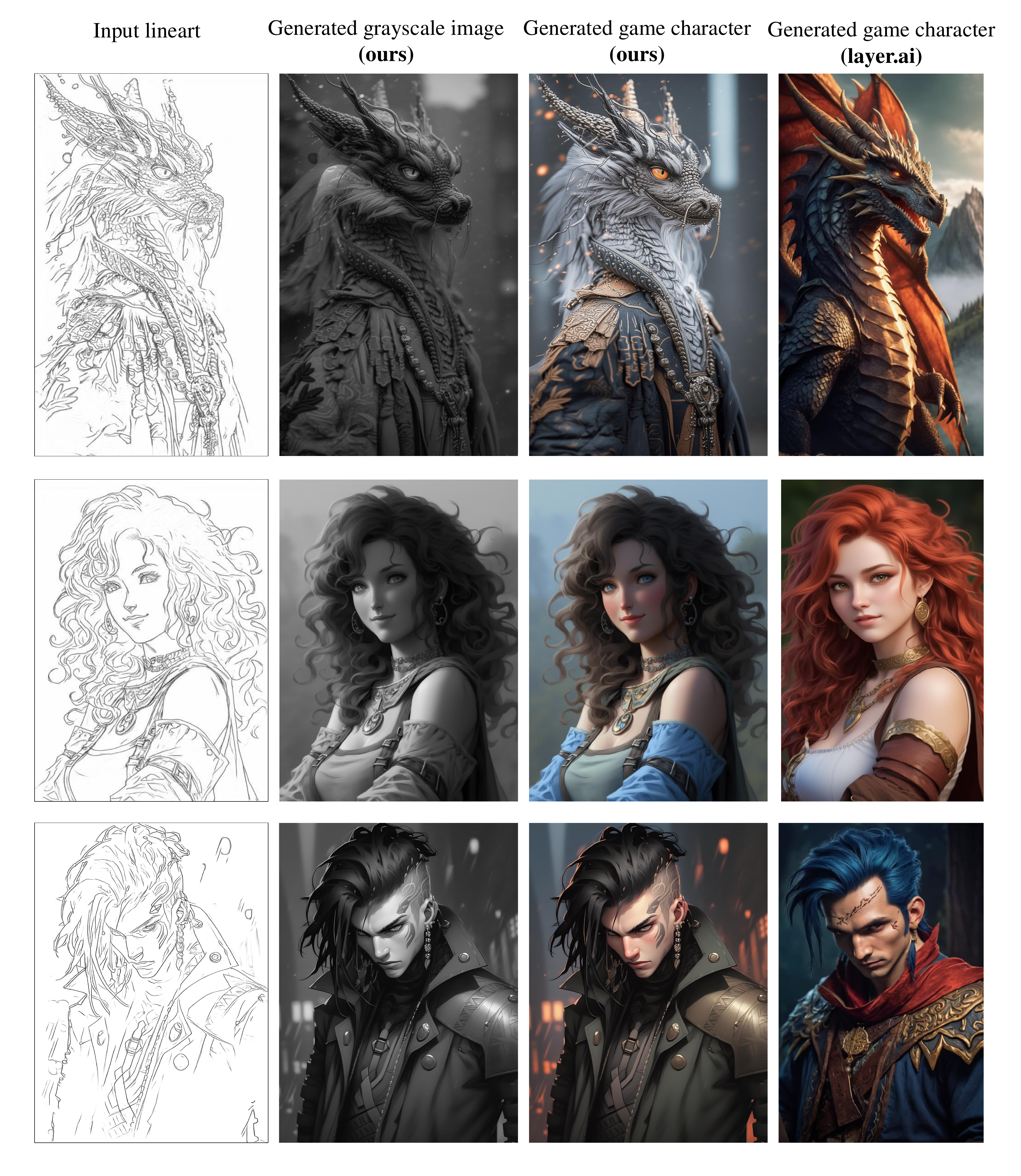}
    \caption{Visualizations of lineart$\to$Grayscale$\to$Character image generation.}
    \label{fig:character_visual}
\end{figure}

\textbf{Qualitative comparison for Lineart → Grayscale → Character image generation.} For the control model of game character generation, this study compares our method with layer.ai~\cite{layerai}, which is a leading game AIGC platform. As illustrated in ~\Cref{fig:character_visual}, our method retains more consistency with the input image. 

\textbf{Evaluation For Game Character Consistency Model.} For our Game Character Consistency Model, this study conducts a quantitative comparison using our constructed benchmark, which contains 110 pairs of images from vroid-dataset~\cite{chen2023panic}. In this section, we compare our method with the current state-of-the-art method MangaNinja~\cite{liu2025manganinja}. Additionally,
this study also conducts comparisons with IP-Adapter~\cite{ye2023ip}. 

\begin{table}[h]
\centering
\caption{Results of Quantitative comparison for Hunyuan-Game Character Consistency Model}
\label{hunyuan-game-character-consistency eval results}
\scalebox{0.95}{
\begin{tabular}{lccccc}
\toprule
Methods& DINO ↑ & CLIP ↑ & PSNR ↑ & MS-SSIM ↑ & LPIPS ↓ \\
\midrule
MangaNinja& 0.8324 & 0.9578 & 10.9660 & 0.7731 & 0.2412 \\
IP-adapter& 0.8562 & 0.9545 & 7.9965 & 0.5235 & 0.1783 \\
Ours & \textbf{0.9545} & \textbf{0.9887} & \textbf{25.1246} & \textbf{0.9577} & \textbf{0.0352} \\
\bottomrule
\end{tabular}
}
\end{table}

\begin{figure}[htbp]
    \centering
    \includegraphics[width=0.95\textwidth]{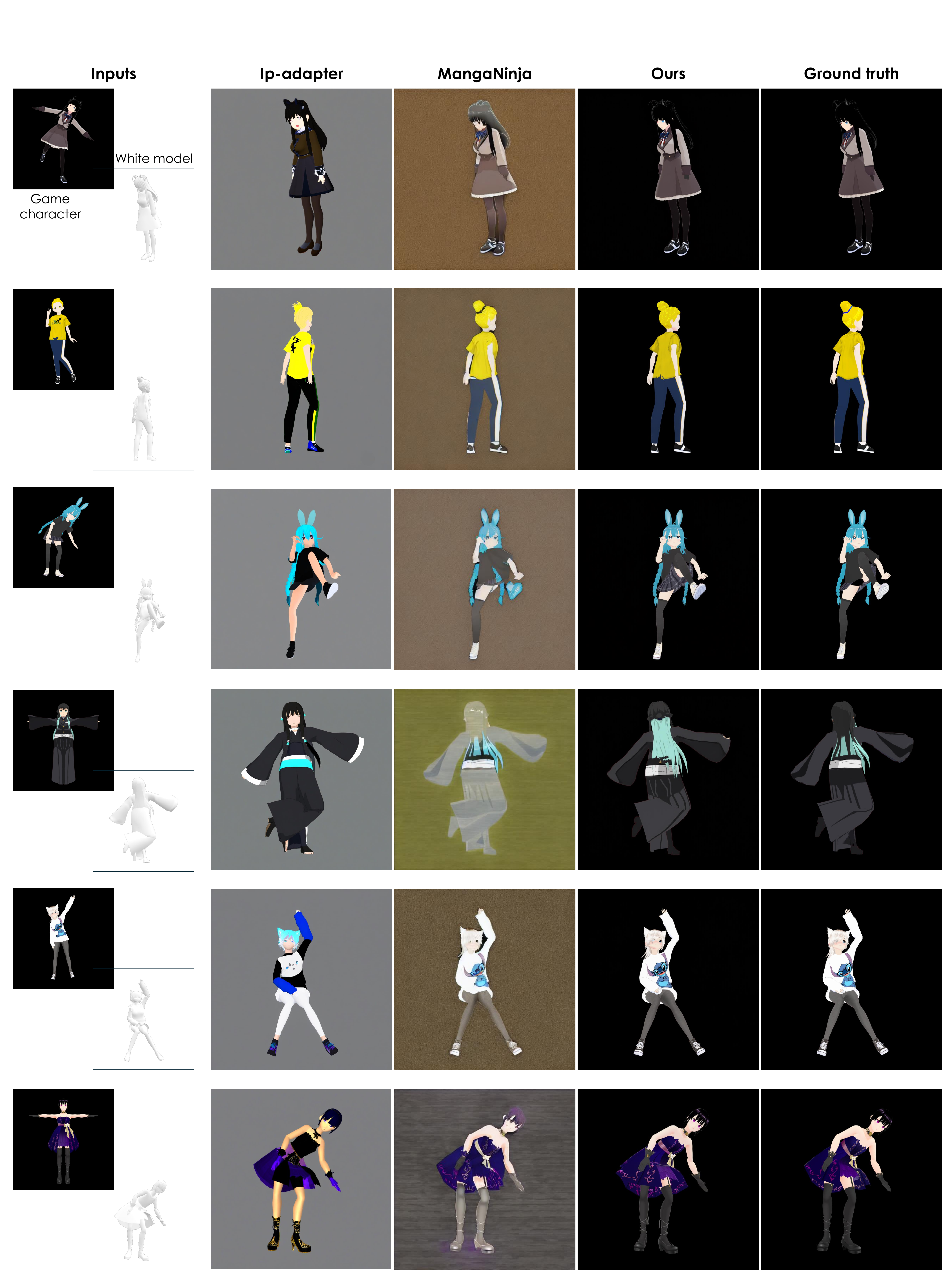}
    \caption{Visualizations of consistent game character generation results.}
    \label{fig:character_consistency_visual}
\end{figure}

\textbf{Qualitative comparison for Game Character Consistency Model.} The comparison results is visualized in ~\Cref{fig:character_consistency_visual}. Our method utilizes white models as the target structure, while MangaNinja and IP-adapter process the ground truth into lineart or depth map. Benefiting from the design of our method, our method achieves more reasonable generation results than other compared methods.

\textbf{Quantitative comparison for Game Character Consistency Model.} This study calculates the CLIP~\cite{radford2021learning} and DINO~\cite{oquab2023dinov2} semantic image
similarities between the generated images and the ground
truth to measure the performance of the compared methods. In addition, Peak Signal-to-Noise Ratio (PSNR) and the Multi-Scale Structural Similarity Index (MS-SSIM)~\cite{wang2004image} are also used to assess the quality of the generated results. As illustrated in Table~\ref{hunyuan-game-character-consistency eval results}, our methods outperform the compared methods in all metrics. 

%% file: sections/video/1_i2v.tex
\clearpage
\section{Hunyuan-Game-Video Generation}
\subsection{Image-to-Video Generation}
\label{i2v}

\subsubsection{Introduction}

With the rapid advancement of artificial intelligence (AI), its applications in the game industry have become increasingly widespread.
Among these, diffusion-based generative models have enabled high-quality and controllable content creation across various modalities such as prompt conditioned image generation~\cite{saharia2022photorealistic,li2024hunyuan,gao2025seedream,rombach2022high}, video synthesis~\cite{blattmann2023align,blattmann2023svd,kong2024hunyuanvideo,seawead2025seaweed,hu2025hunyuancustom,zhou2024allegro,sora,kling,hailuo,veo2}, 3D asset creation~\cite{yang2024hunyuan3d,zhao2025hunyuan3d,yang2025pandora3d,tang2023make,zhang2024clay}, and animation
~\cite{chen2024motionclr, dai2024motionlcm, liu2023plan, lu2024scamo, pan2025tokenhsi, ji2025sport,fan2024freemotion,tang2024make,xiao2025motionstreamer,Tao_2022_CVPR,tao2022motion,tao2023learning,xu2024learning,xu2021move}. In this context, \textbf{image-to-video (I2V)} proves highly valuable for game artists. Its strong controllability and consistency allow the game artists to efficiently create high-quality animated videos directly from \textit{concept arts}, significantly reducing production time and costs. Additionally, I2V enables rapid iteration and experimentation, empowering artists to explore dynamic storytelling and visual effects without extensive manual animation. However, existing I2V models~\cite{wang2025wan, kong2024hunyuanvideo, kling, huang2025step, yang2024cogvideox, hailuo} exhibit notable limitations when applied to game scenarios: (1) These models struggle to accurately capture game-specific concepts and mechanics, often generating corrupted outputs characterized by distorted physics, incoherent interactions, and other artifacts. (2) Existing models suffer from fundamental deficiencies in visual aesthetics, unable to represent the sophisticated artistry and dynamic beauty inherent in game videos. As a result, the generated videos appear visually simplistic and fall short of the high standards expected by professional game artists. Thus, developing an effective I2V model capable of handling diverse game scenarios carries significant practical value for enhancing the quality of game videos and accelerating the creative pipeline for computer-generated imagery (CGI).

To address these challenges, we present \textbf{Hunyuan-Game I2V}, the most professional image-to-video model in the field of game videos. Before delving into details, we sum up contributions as follows.

\vspace{-0.75em}
\begin{itemize}
\item We curate a dataset comprising millions of diverse game and animation videos and establish a multi-stage data filtering pipeline (\Cref{subsection_wx: Data Filtering}) to ensure the selection of high-quality training samples. Additionally, we design a specialized captioning system (\Cref{subsection_wx: Data Annotation}) tailored to game scenarios, which enables effective model domain adaptation.
\item We collaborate with professional game artists to establish comprehensive aesthetic standards for video content. By integrating principles of static composition and dynamic motion aesthetics, we systematically refine our collected dataset to enhance model performance in the game scenarios.
\item Leveraging the constructed dataset, we design an adaptive training strategy and propose our prompt rewriting model in \Cref{subsection_wx: Training}. Extensive experimental results show that our Hunyuan-Game I2V achieves the state-of-the-art performance (\Cref{subsection_wx: Evaluation}).
\end{itemize} 

\subsubsection{Data Filtering}
\label{sec: data_filter}
\label{subsection_wx: Data Filtering}

Our data pre-processing pipeline builds upon the framework of HunyuanVideo~\cite{kong2024hunyuanvideo}. We incorporate specialized adaptations specifically designed for animation-centric video generation. Given the inherent scarcity of high-quality gaming CGI, we implement a comprehensive data collection strategy, significantly expanding our dataset by systematically including diverse animation content to ensure robust model training capabilities.

The primary composition of our raw dataset encompasses both 2D and 3D animation content. We develop a multi-stage cleaning methodology that integrates critical components from the HunyuanVideo pipeline. This includes shot boundary detection, scene transition analysis, OCR-based text detection and filtering, border and resolution standardization, logo detection and removal, and minimum frame count verification. To mitigate the inherent variability in animation data quality, we implement more rigorous quality control thresholds than those used in the original data pipeline, thereby ensuring greater consistency and reliability across the entire dataset.

To establish a robust and reliable labeling system, we engage professional annotators to conduct manual labeling of 100,000 video clips across multiple attributes. These meticulously annotated samples subsequently serve as training data for our specialized captioning model, which facilitates the automation of the classification process for the entire dataset. The model implements a comprehensive categorization system, distinguishing between animation types (2D versus 3D), subject characteristics, cinematographic elements (including camera angles and motion dynamics), aesthetic quality metrics, and thematic content analysis. This sophisticated classification framework enables precise control over dataset composition while maintaining essential diversity in training samples.

\begin{figure}[htbp]
    \centering
    \includegraphics[width=0.8\textwidth]{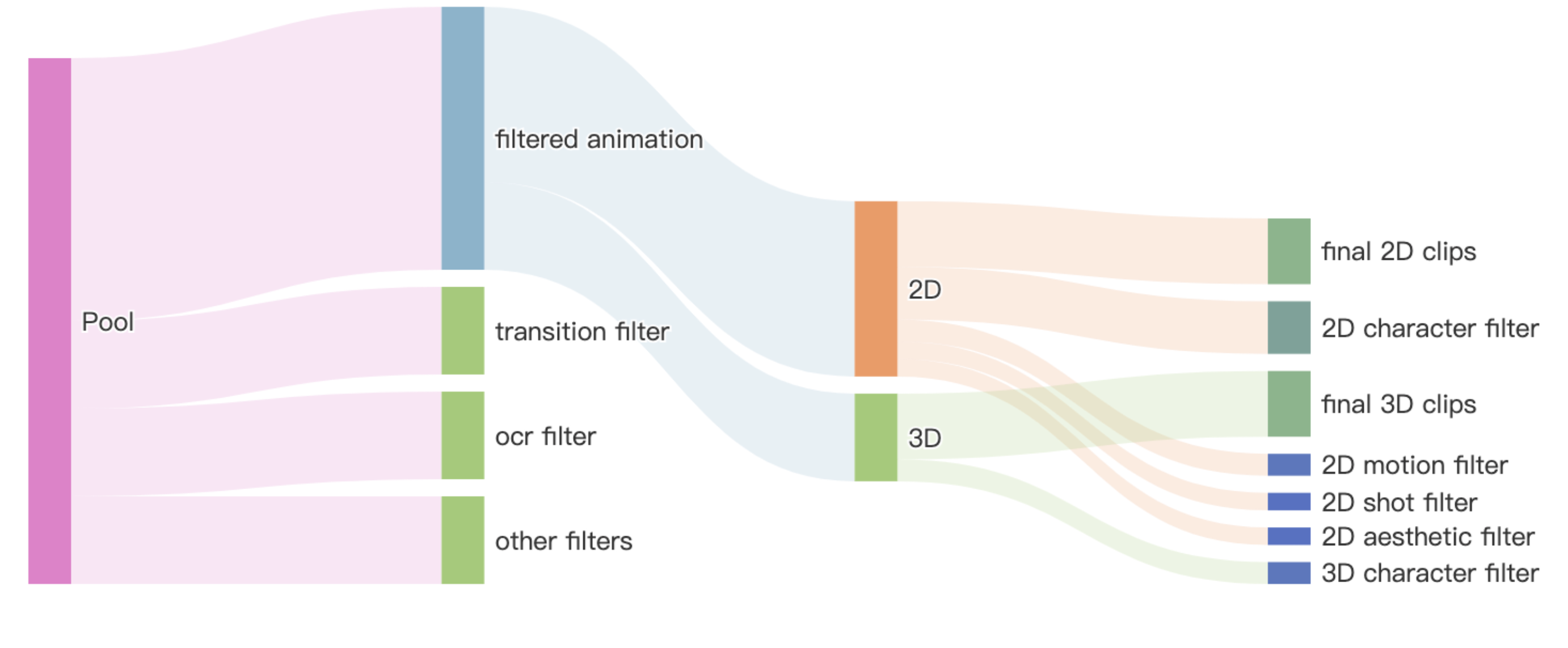}
    \caption{\textbf{The data filtering pipeline}.}
\end{figure}

The filtering methodology was systematically implemented based on the model-generated labels. For 2D animation content, we enforced rigorous motion-based filtering criteria to exclude static or minimally dynamic sequences, retaining exclusively those exhibiting significant character movement and superior aesthetic quality. This stringent approach was necessitated by the predominance of 2D animation in our raw data, which frequently contains substantial static or low-motion content. Conversely, our approach to 3D animation implemented a more inclusive preservation strategy, maintaining a broader spectrum of motion sequences while maintaining stringent aesthetic quality thresholds. This differential treatment methodology was informed by empirical observations indicating that 3D animation inherently exhibits superior motion characteristics, despite constituting a smaller proportion of the raw dataset.

In conjunction with the aesthetic scoring system in Hunyuan-Game-Image, we have developed two aesthetic dimensions specifically for video evaluation: Motion Rationality and Motion Richness. Motion Rationality evaluates whether the movement of primary elements within the video adheres to physical laws, detecting any anomalies or deformations. Motion Richness quantifies the diversity of motion types per unit time in the video sequence.

These two video aesthetic operators serve as crucial components in our data filtering process and model iteration pipeline. They provide quantitative metrics for assessing both the naturalness of movements and the dynamic complexity of video content, thereby enhancing our overall video quality assessment framework.

Through this methodically designed filtering approach, we successfully achieved optimal balance between 2D and 3D content proportions in our final dataset, establishing a near 1:1 ratio. This balanced distribution proved crucial for ensuring unbiased model learning across both animation styles, preventing potential overfit towards the more prevalent 2D content. Across both categories, we maintained strict prioritization of human-centric content, ensuring the final dataset comprised exclusively high-aesthetic-value sequences conducive to effective model training.

\subsubsection{Data Annotation}
\label{sec: data_anno}
\label{subsection_wx: Data Annotation}

Our annotation framework enhances video understanding through structured captioning while leveraging insights derived from our data filtering pipeline. Moving beyond conventional brief or dense captioning approaches, we introduce an innovative dynamic-static hierarchy that comprehensively captures both spatial and temporal aspects of video content.

The comprehensive caption structure encompasses five distinct components: long visual caption, long motion caption, short visual caption, short motion caption, and a structured tag system. Acknowledging the critical role of motion-centric descriptions in enhancing generated video dynamics, we engineered our captioning system to explicitly differentiate between static and dynamic elements. Static captions focus on frame-level visual content analysis, while dynamic captions incorporate temporal evolution patterns, scene transitions, and implied camera movements.

To optimize captioning efficiency while maintaining high quality standards, we employ a 7B vision-language model fine-tuned through knowledge distillation from a larger teacher model. This approach effectively balances computational feasibility with caption generation quality. Following the methodology of Seaweed-7B~\cite{seawead2025seaweed}, our caption model employs a hierarchical generation approach that produces long captions before generating their shorter variants. The approach is inspired by chain-of-thought (CoT) reasoning, where sequential, step-by-step processing enhances coherence and reduces hallucination. Analogous to CoT, our model generates captions in three structured phases: first a detailed caption, then a short caption, and finally a tag system. This systematic progression ensures that each step builds upon the previous one, improving both accuracy and logical flow. 

\begin{figure}[htbp]
    \centering
    \includegraphics[width=0.8\textwidth]{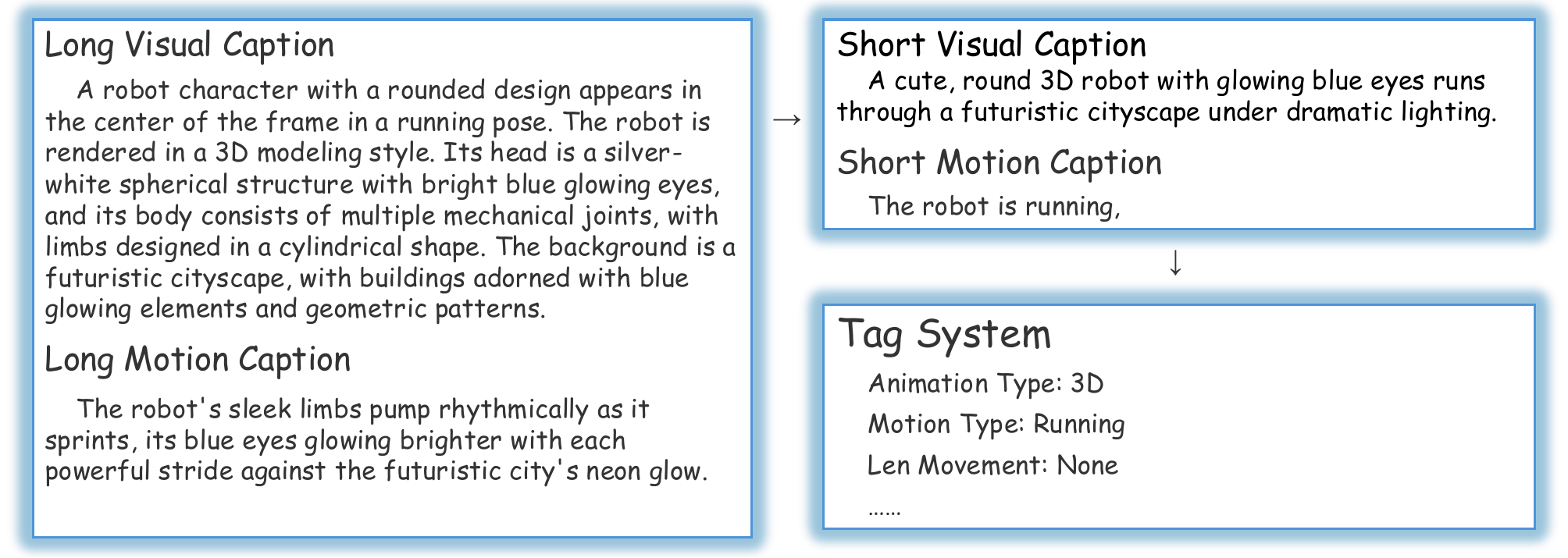}
    \caption{\textbf{An example of the structured caption for video clips}. The VLM generated the caption sequentially from the most detailed caption to a summarised caption to several labels. It mocks a chain-of-thought process that can reduce the probability of model hallucination.}
    \label{fig:video_caption}
\end{figure}

\begin{figure}[htbp]
    \centering
    \includegraphics[width=\linewidth ]{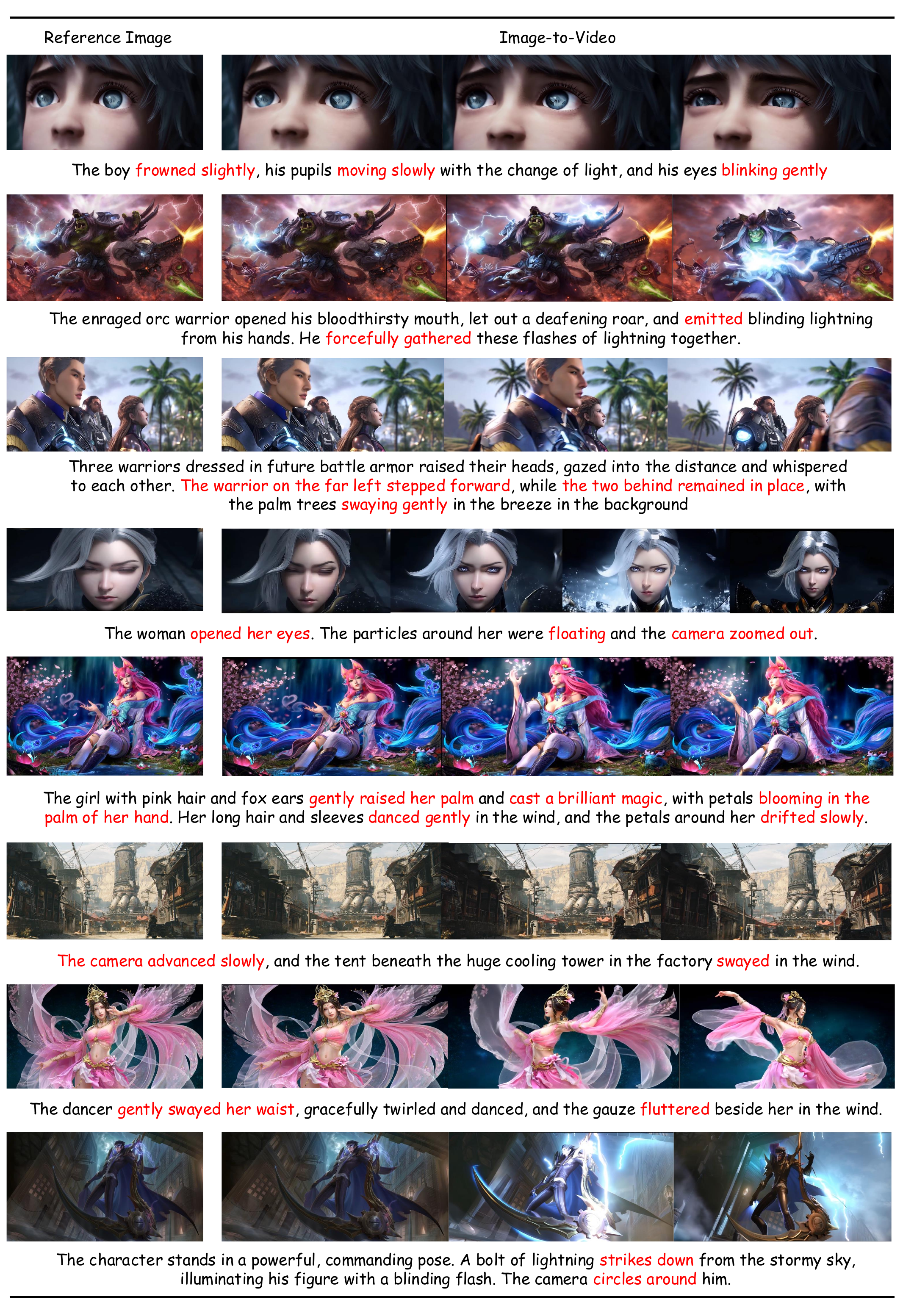}
    \caption{Qualitative results of \name. Our method exhibits robust ID preservation and prompt-following capability, achieving superior visual fidelity and motion naturalness.}
    \label{fig:i2v_results}
\end{figure}

The structured captioning system incorporates an integrated label system, as detailed in Fig.\ref{fig:video_caption}. Section~\ref{sec: data_filter} contains the labeling system specifications. Unified training of captioning and labeling tasks within a single model framework enhances semantic integration of image information, thereby improving output accuracy. 

Beyond semantic descriptions, precise camera movement annotation represents a crucial component for controllable video synthesis. We utilize an internal camera motion classifier developed by Hunyuan Video. Rather than treating camera metadata as isolated labels, we implement direct embedding into the captioning process. Camera motion predictions are explicitly referenced within dynamic captions (e.g., "panning left as the character moves"), ensuring that camera dynamics not only inform but actively integrate with language descriptions.

% \begin{figure}[htbp]
%     \centering
%     \includegraphics[width=0.7\linewidth]{figures/I2V_overall_token_replace.png}
%     \caption{Framework of \name.}
%     \label{fig:i2v_backbone}
% \end{figure}

\subsubsection{Training}
\label{subsection_wx: Training}

\paragraph{Main Model}

We adopt HunyuanCustom~\cite{hu2025hunyuancustom} as the base model for game-specific fine-tuning. The training process consists of two stages, each utilizing different datasets to progressively enhance the model's performance.

\vspace{-0.75em}
\begin{itemize}
    \item \textbf{Full-Scale Data Fine-Tuning (SFT Stage).} In the initial stage, we employ a multi-stage data filtering pipeline to select 700K+ game and anime videos from the full dataset for supervised fine-tuning (SFT). This stage enable the HunyuanCustom model to transition into I2V (image-to-video) mode while achieving domain transfer for game and anime content.
    \item \textbf{High-Quality Data Selection Fine-Tuning (Quality Tuning, QT Stage).} In this stage, we leverage game video aesthetic operators to filter the 700,000 data entries, selecting 80,000 high-quality samples based on static aesthetics and dynamic aesthetics to ensure superior motion fluidity and visual appeal for model training. Additionally, we increase the sampling weight of game-specific data to enhance the model’s performance in game animation.
\end{itemize}
\vspace{-0.75em}

Furthermore, to enhance the model's dynamic responsiveness to text and prevent generated videos from appearing static, we introduce a proprietary video captioning operator during training. This operator simultaneously generates static and dynamic captions: static captions primarily focus on depicting the overall scene, while dynamic captions emphasize motion characteristics within the visual content. By combining both caption types—each further categorized into long and short descriptions—we significantly improve the model's text-to-motion alignment. Notably, sampling short descriptions at a higher ratio further boosts dynamic responsiveness. Additionally, we probabilistically concatenate labels such as mood, lighting, camera movement, background, and style with the captions to strengthen the model’s understanding and response to these attributes.

% During the training process, we employ both static and dynamic captions. Static captions primarily focus on holistic scene descriptions, while dynamic captions emphasize motion characteristics within the visual content. Both caption types include long and short variants, with short captions being sampled at a higher rate. Additionally, meta-information captions describing atmosphere, lighting, camera movement, background, and style are incorporated into training with a predefined probability.

% 我们采用 HunyuanCustom 作为游戏图生视频模型进一步微调的基础模型。模型的训练过程分为两个阶段，每个阶段使用不同的数据，以逐步提升模型的性能：
% 第一阶段：全量数据微调（SFT 阶段）
% 在此阶段，我们通过多层级数据筛选管道，从全量数据中筛选出70万条游戏动漫视频数据进行训练。这一过程使 HunyuanCustom 模型转变为 I2V 模式，同时实现了模型在游戏动漫领域的适应性迁移，为后续的优化训练奠定了坚实基础。
% 第二阶段：高质量数据微调（QT 阶段）
% 在这一阶段，我们利用游戏视频美学算子对70万条数据进行筛选，从静态美学和动态美学两个维度挑选出8万条动态性良好且美感突出的样本进行模型训练。同时，我们还增强了游戏领域数据的采样，以提升模型在游戏动画和 CG 方面的表现。
% 此外，为了提升模型对文本的动态响应，避免生成的视频趋于静态，我们在训练过程中引入了自研的视频描述算子。该算子能够同时输出静态和动态描述，其中静态描述主要聚焦于整体场景的描绘，而动态描述则强调视觉内容中的运动特征。我们将静态描述与动态描述结合使用，以增强模型的表现。这两种描述类型均包含长描述和短描述，通过更高比例采样短描述，显著提升了模型对文本的动态响应能力。此外，我们还将氛围、灯光、相机运动、背景、风格等标签按一定概率与描述进行拼接，以进一步增强模型对相关标签的理解与响应。

\paragraph{Rewriting model}

Similar to image generation tasks, high-quality input prompts are crucial for the quality of video generation. Prompts containing both static and dynamic descriptions can effectively guide the model to produce higher-quality video content that aligns with user requirements. To lower the user entry barrier and enhance interaction convenience, this study designs and implements a specialized prompt optimization framework.

Our rewriting model is trained in two distinct phases. In the first phase, we synthesize training data for the rewriting model using the caption texts from the training set. After training for one epoch, it is observed that the rewritten outputs sometimes failed to maintain consistency with the original input. To address this, reinforcement learning (RL)\cite{shao2024deepseekmathpushinglimitsmathematical} was introduced in the second phase of training, where consistency between the rewritten output and the original input, along with a repetition penalty, was incorporated into the reward function. After the second phase of training, the model achieves a consistency rate 98\% between diverse inputs. Prompt rewriting can reduce the rate of malformation of the generated videos and increase the probability that the generated videos follow instructions.

\begin{figure}[htbp]
    \centering
    \includegraphics[width=1.0\textwidth]{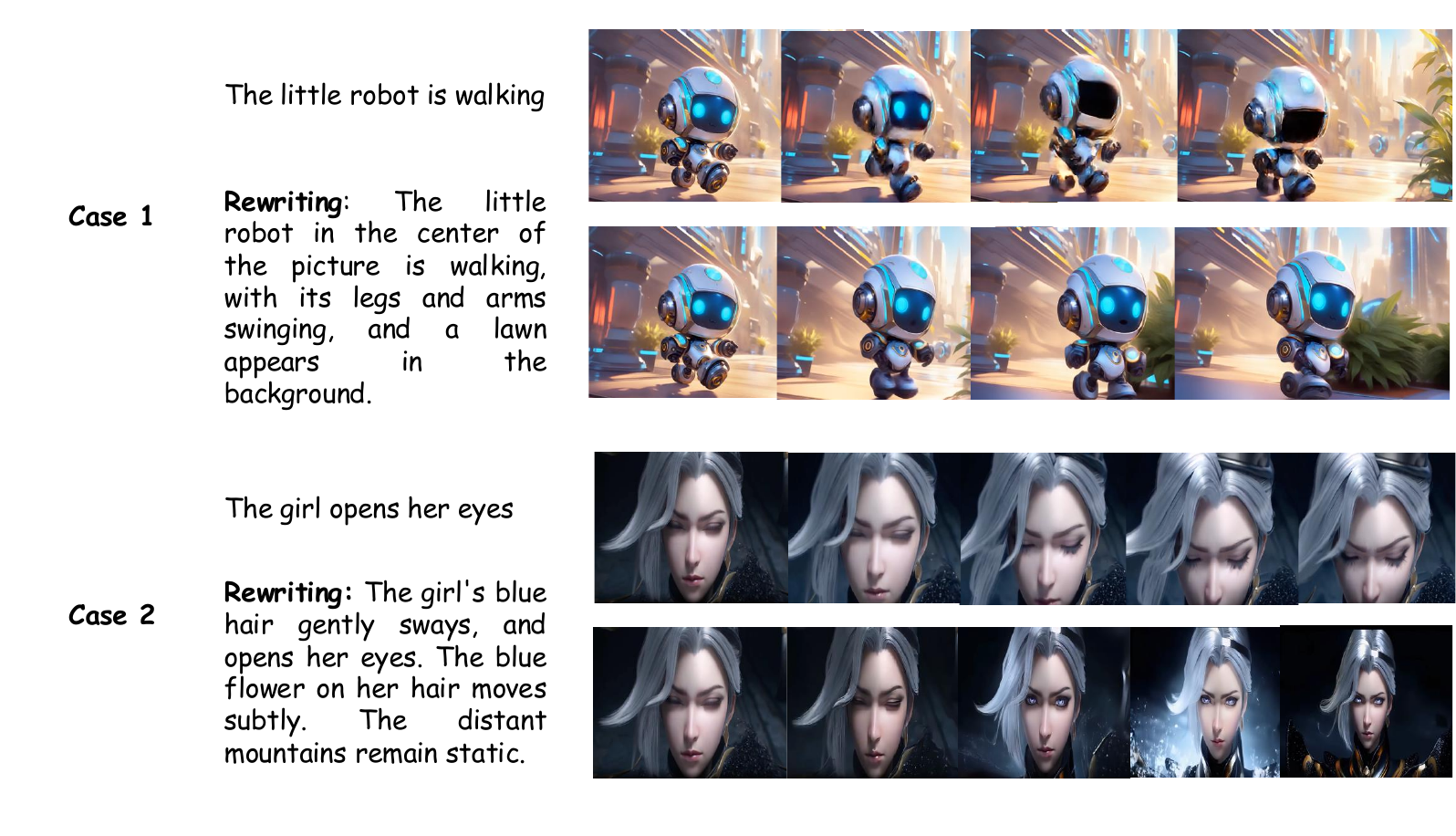}
    \caption{\textbf{Examples of the prompt rewriting}. Rewriting prompts can reduce the probability of subject distortion and enhance the model's ability to follow instructions.}
    \label{fig:i2v_rewriting}
\end{figure}

\subsubsection{Evaluation}
\label{subsection_wx: Evaluation}

We developed an evaluation dataset comprising approximately 200 images, which encompasses a diverse range of visual content, including illustrations, animation screenshots, game promotional materials, and other related imagery. For each image in the dataset, we generated prompts using a Vision-Language Model (VLM), followed by manual refinement and validation to ensure accuracy. 

For each image-generated video, we evaluate five key dimensions: text-video alignment, image-video alignment, visual quality, motion quality, and an overall score. Three annotators independently rate each video on a five-point scale for these criteria. During evaluation, annotators compare outputs from all models side-by-side to ensure a fair assessment. For the overall score, annotators provide a subjective rating based on their holistic preference, weighing all dimensions.

Our comparative analysis included several competitive open-source and proprietary models, such as Kling 1.6 Pro and Wan 2.1, with the detailed results presented in Table~\ref{hunyuan-game i2v eval results}.

\begin{table}[h]
\centering
\caption{Results of Evaluation for Hunyuan-Game-Video Generation}
\label{hunyuan-game i2v eval results}
\scalebox{0.95}{
\begin{tabular}{lccccc}
\toprule
Model       & I-V Alignment & T-V Alignment & Visual Quality & Motion Quality & Overall \\
\midrule
CogVideoX    & 3.32          & 3.06          & 3.04           & 2.88           & 2.78    \\
Wan 2.1& 3.50& \textbf{3.64}& 3.84& 3.36& 3.24\\
Minimax     & 3.44          & 3.46          & 3.48           & 3.16           & 3.10    \\
Kling
1.6 Pro& \textbf{3.92}& 3.47& \textbf{3.94}& \underline{3.45}& \textbf{3.47}\\
Ours & \underline{3.84}& \underline{3.59}& \underline{3.86}& \textbf{3.53}& \underline{3.31}\\
\bottomrule
\end{tabular}
}
\end{table}

In terms of overall quality, our model performs slightly worse than Kling 1.6 Pro but outperforms Wan 2.1. Notably, our model achieves higher ratings than Kling 1.6 Pro in motion quality, demonstrating strengths in dynamic generation.

%% file: sections/video/2_avatar.tex
\subsection{360° A/T Pose Character Video Generation}
\label{avatar}

% @unicorn
\subsubsection{Introduction}
In the field of game creation, character design is a crucial element. After the designer completes the character concept design, it is often necessary to create multi-view drawings of standard poses to further refine the details of various character parts. The multi-view drawings helps to eliminate visual blind spots from a single perspective, allowing the team to evaluate the character design from different dimensions. These drawings can also be used for downstream applications, such as model creation and other processes. Related works \cite{peng2024charactergen} have demonstrated the capability to parse reference images into standard poses and generate multi-view images; however, most multi-view image generation methods \cite{shi2023zero123++, liu2023syncdreamer} do not support standard pose transformation. These approaches are unable to produce rotational videos, thereby limiting the provision of richer viewpoint information and exhibiting bottlenecks in maintaining pose consistency. On the other hand, although existing video-based methods can generate rotational videos \cite{wang2025wan, kling}, they fail to transform characters into standard poses and face significant limitations in gaming applications. Specifically, (1) maintaining character consistency throughout the rotation process is challenging, often resulting in limb distortions; and (2) character renderings from different viewpoints fall short of the high-quality standards required by professional game designers. To address these issues, Hunyuan-Game has explored a novel approach for generating 360-degree rotating videos of characters.

The main objective of this task is to input concept design images of a game character in any pose, calibrate the input to generate a standard pose (A/T-pose), and create a 360-degree rotating video while maintaining the consistency of the character. Our method also enables the generation of additional details, such as the character's clothing and texture, from angles not visible in the input image, such as the back and sides, providing valuable reference for the designer.

Based on the Hunyuan-Game I2V foundation model, we have incorporated visual features into the model structure to ensure consistency during the character's rotation. Additionally, we have carefully designed a data construction pipeline to achieve stable rotation. This section will delve into the challenges faced in this direction, the specific methods employed, and the data construction process.

\begin{figure}[htbp]
    \centering
    \includegraphics[width=0.7\textwidth]{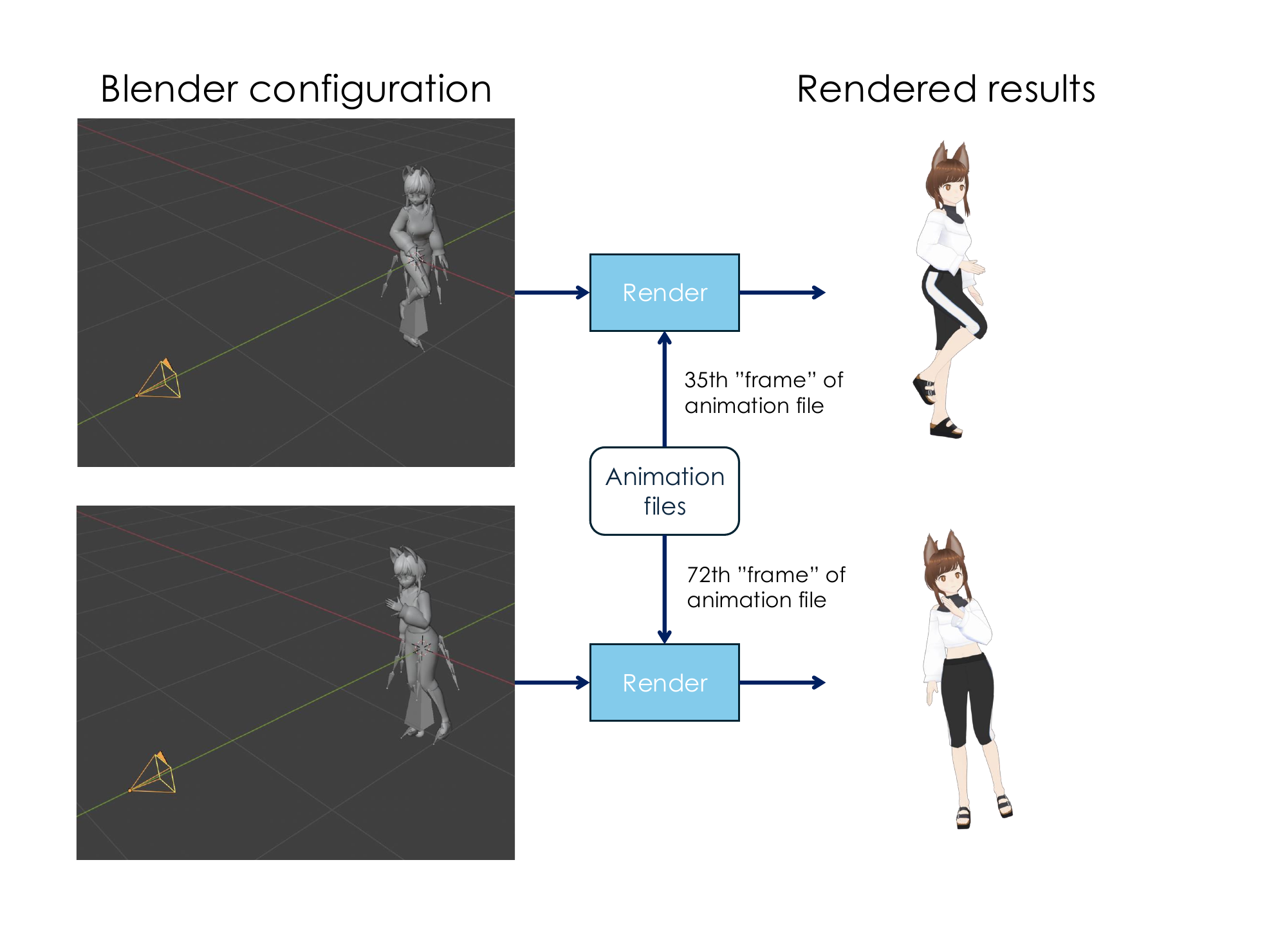}
    \caption{Mesh data processing pipeline used to generate dataset used in training. 
    We filter character meshes by various methods, then render a group of images used in training process.}
    \label{sec4:render_process}
\end{figure}
\subsubsection{Data}

The task of A/T pose video generation requires much more 3D consistency in generated videos compared with other tasks.
Instead of directly using videos related to games, we generate a group of images from a group of 3D character models. The image generation process is based on blender\footnote{\url{https://www.blender.org}} and is similar to methods demonstrated in Pandora3D~\cite{yang2025pandora3d}. 
We prepare around $50k$ character models with correct rigging and skinning created by artists. The whole pipeline of processing these meshes can be found in Fig. ~\ref{sec4:render_process}. 
We discard models without correct rigging and skinning. We also render $9$ images for each model and use Hunyuan vision model~\cite{sun2024hunyuan} to filter meshes without complete human body. 
Correct meshes are used in the following rendering process.
The reference images used during training consist of rendered images of character models in various poses taken from different animations, while the ground truth images are a set of rendered images of these character models rotating 360 degrees in a standard pose.
To generate a large amount of character models with versatile poses, we collect a group of animation files created by artists, and adopt these animation files onto character models.

We randomly choose a group of ``frames'' in one animation file, and render character models at the pose in these ``frames''. We render $432$ posed orthographic images for each character model, and $120$ orthographic images at resting pose for each character model. The images at standard poses are compressed into a video file using lossless video encoding formats FFV1\footnote{\url{https://www.ffmpeg.org/~michael/ffv1-draft/ffv1.pdf}}, which is particularly suitable for archiving and preserving image data~\cite{kromer2017matroska}. All images are rendered at the resolution $2048 \times 2048$. To achieve a more realistic and gentler lighting effect in generated videos, we choose a random hdri image from a group of $60$ hdri images when rendering ground truth videos; whereas prompt images are all rendered using \textbf{only one} carefully chosen hdri map: this hdri map is very close to theoretically uniform light map, making most parts of rendered images have uniform brightness; on important areas such as noses, faint shadows still exists, which will help preserve details (\eg~shape of nose) in generated images.

\begin{figure}[t]
    \centering
    \includegraphics[width=1.0\textwidth]{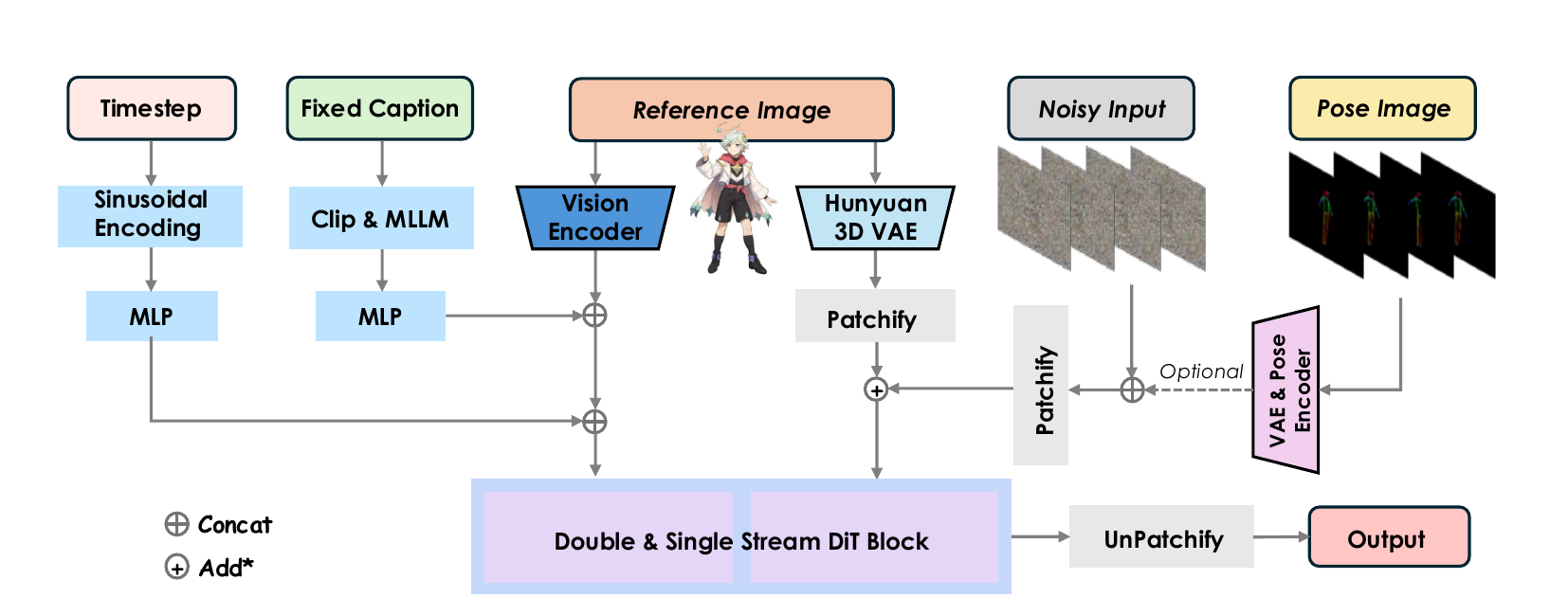}
    \caption{Framework fo Hunyuan-Game 360° Character Video Generation}
    \label{fig:character_360_pipeline}
\end{figure}

\begin{figure}[htbp]
    \centering
    \includegraphics[width=1.0\textwidth]{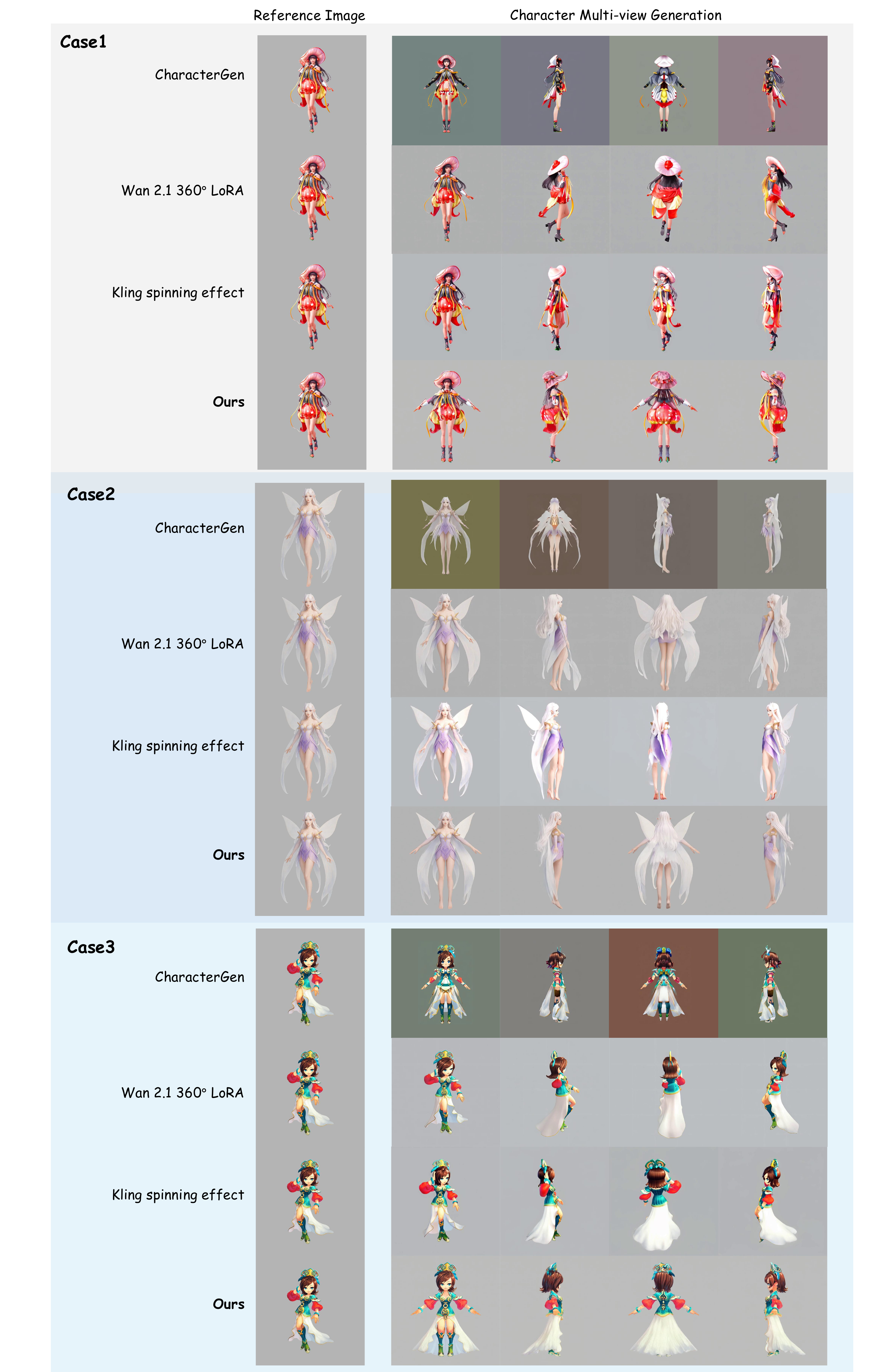}
    \caption{Qualitative comparisons with State-of-The-Art methods, 
to align the results, for the Kling spinning effect, Wan2.1 360° LoRA, and our method, 4 frames are extracted from the generated videos for comparison. In contrast, it can be seen that our method has a better effect.}
    \label{fig:character_360_compare}
\end{figure}

\begin{figure}[htbp]
    \centering
    \includegraphics[width=1.0\textwidth]{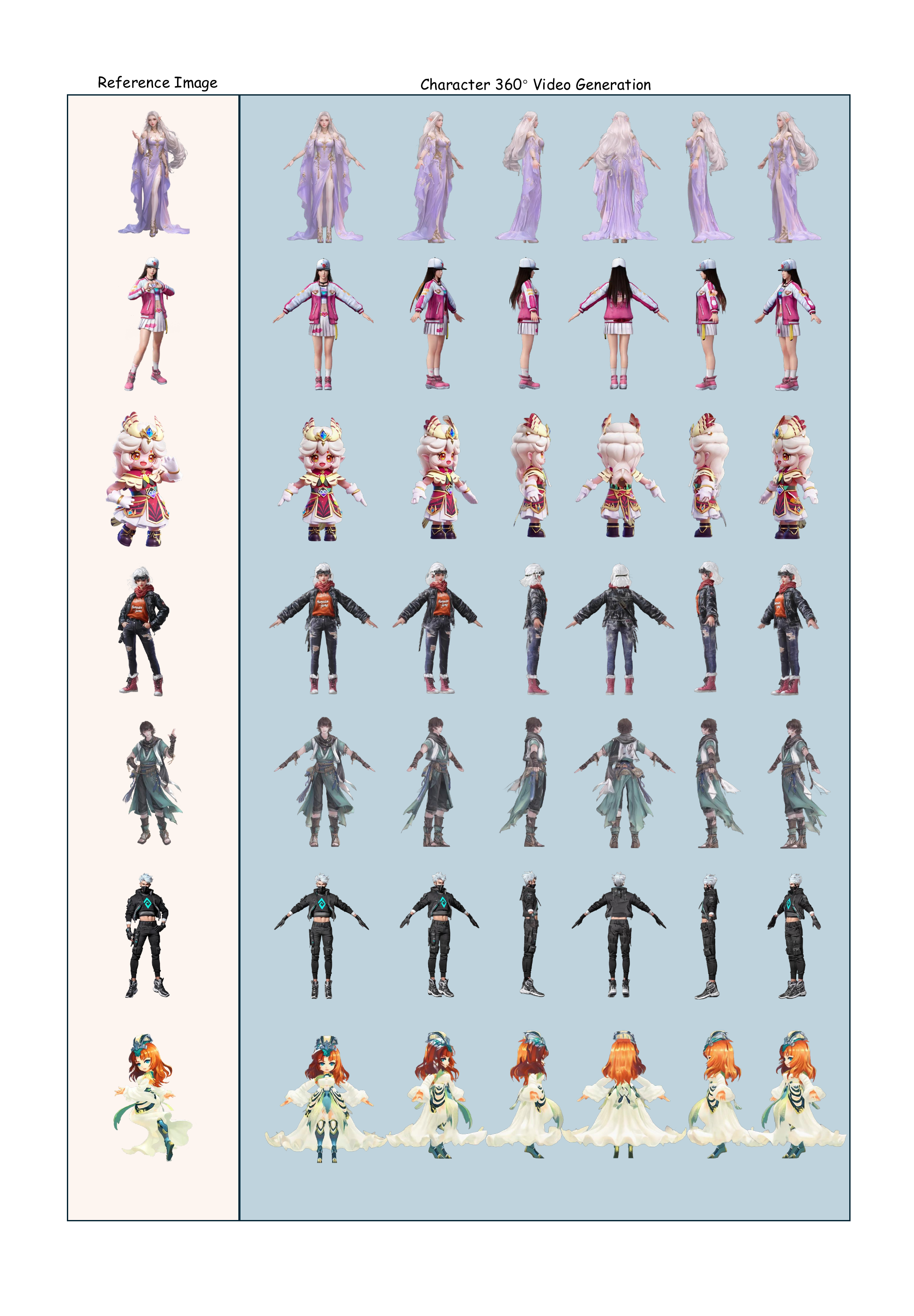}
    \vspace{-3em}
    \caption{Qualitative results of 360° Character Video Generation. Our method demonstrates excellent character consistency and rotation robustness, and it is capable of generating reasonable clothing and texture details from different viewpoints.}
    \label{fig:character_360_1}
\end{figure}

\subsubsection{Method}

Our approach is based on the HunyuanVideo-I2V\cite{kong2024hunyuanvideo, hu2025hunyuancustom} foundation model, with adjustments made to the model structure specifically for this task, followed by downstream training. This task presents two key challenges: first, ensuring the consistency of the character and transforming its pose to a standard pose; and second, maintaining the stability of the rotation while generating reasonable outputs for angles not present in the reference image, such as the back or side views.

To address the aforementioned challenges, we introduced an additional visual encoder, SigLIP~\cite{zhai2023sigmoid, zhang2025packing}, to the foundation model to extract features of the input character. These image features are then concatenated with the LLaVA features~\cite{liu2023visual,liu2024improved} and injected into the model. Optionally, we incorporated skeletal control conditions by training a skeletal feature extraction network, which adds skeletal information to the noise latent as initialization. The  input character image token is concatenated with the noise tokens, thereby enabling the deep integration of character semantic information into the generated frames. The framework of the method is shown in Fig. \ref{fig:character_360_pipeline}.

The overall training process is divided into two stages:

(1) In the first stage, during the initial training, we use the full dataset to learn pose variations and rotation models. For each character, we randomly select a reference image from the pose data rendered from animation  and pair it with 120 frames of rendered rotation images to create a training dataset. To better showcase the character and facilitate later processing such as image matting by designers, we fill the background of the images with gray using the alpha channel.

(2) In the second stage, we filter out character models of higher quality, which have superior details in clothing textures and design. We use approximately 7k high-quality data samples for quality tuning and apply data augmentation techniques, such as random rotation and scaling of the reference images, to enhance the model's generalization ability.

\subsubsection{Evaluation}

As depicted in the Fig. \ref{fig:character_360_1}, we present qualitative results to verify the efficacy of Hunyuan - 360° character video generation across various styles of game characters. To showcase the details of the generated characters from diverse angles, we have decomposed the generated 360° videos into individual frames.

In terms of consistency, our approach demonstrates a remarkable ability to maintain the character's identity, attire, and texture with high fidelity. Even in cases where the input reference images contain occlusions, the model is capable of rationally inferring and completing the information from different viewing angles.

Regarding stability, this method can transform reference images with various postures and actions into an A-pose while accurately preserving the appropriate body proportions of the character. Building on this foundation, it is able to generate stable 360° rotation videos of the character. We compared the proposed method with CharacterGen \cite{peng2024charactergen} A-Pose multi-view generation, Wan2.1 \cite{wang2025wan} 360 degree rotation video generation, and Kling spinning effect \cite{kling}, with corresponding visualizations shown in Fig.\ref{fig:character_360_compare}. Qualitative analysis indicates that our approach demonstrates superior stability and character consistency in multi-view video generation. In contrast, although CharacterGen can generate multi-view images in canonical poses, it exhibits poor character consistency and is limited to producing a finite number of static images. Meanwhile, Kling and Wan2.1 support rotational video generation; however, their character consistency remains suboptimal, and the rendered character backs lack aesthetic quality.
In summary, our method showcases outstanding performance in the gaming domain.

%% file: sections/video/3_lihui.tex
\subsection{Dynamic Illustration Generation}
\label{lihui}

% @ruihuang

\subsubsection{Introduction}
In recent years, the demand for dynamic character illustrations in games has significantly increased. These dynamic illustrations, often referred to as "live portraits," bring static character designs to life by adding subtle animations to elements such as characters, effects, and weapons. We explore a novel approach to generating dynamic character illustrations by inputting a static reference image and producing a looping video sequence that adheres to specific animation constraints.

The primary objective of this task is to create animations that are both seamless and subtle. The animations must loop continuously, repeating the same set of movements to ensure a smooth and uninterrupted visual experience. Additionally, the movement amplitude should be minimal, avoiding any drastic changes or camera shifts that could disrupt the viewer's focus on the character.

Dynamic character illustrations have a wide range of applications in the gaming industry. One of the most prominent uses is in character entry and idle animations, where these animations enhance the player's engagement by providing a more immersive experience. Furthermore, dynamic illustrations can be utilized in promotional videos, offering a captivating way to showcase characters and their unique attributes to potential players.

By transforming static character illustrations into dynamic animations, game developers can enrich the visual storytelling of their games, creating a more vibrant and interactive world for players to explore. This section will delve into the methodologies employed to achieve these dynamic animations, the challenges faced, and the potential impact on the gaming industry.

\subsubsection{Data}
To support the development of our dynamic character illustration generation model, we amassed a comprehensive dataset consisting of numerous game and animation videos. These videos were segmented into millions of short clips using transition operators. However, the quality of these clips varied widely, necessitating a detailed annotation process to ensure the effectiveness of the dataset. 

As shown in Fig.~\ref{sec4:data}, we conducted manual annotation to classify these clips based on several key criteria, including aesthetics, resolution, and motion quality. The clips were categorized into 3 levels:

\begin{figure}[htbp]
    \centering
    \includegraphics[width=1.0\textwidth]{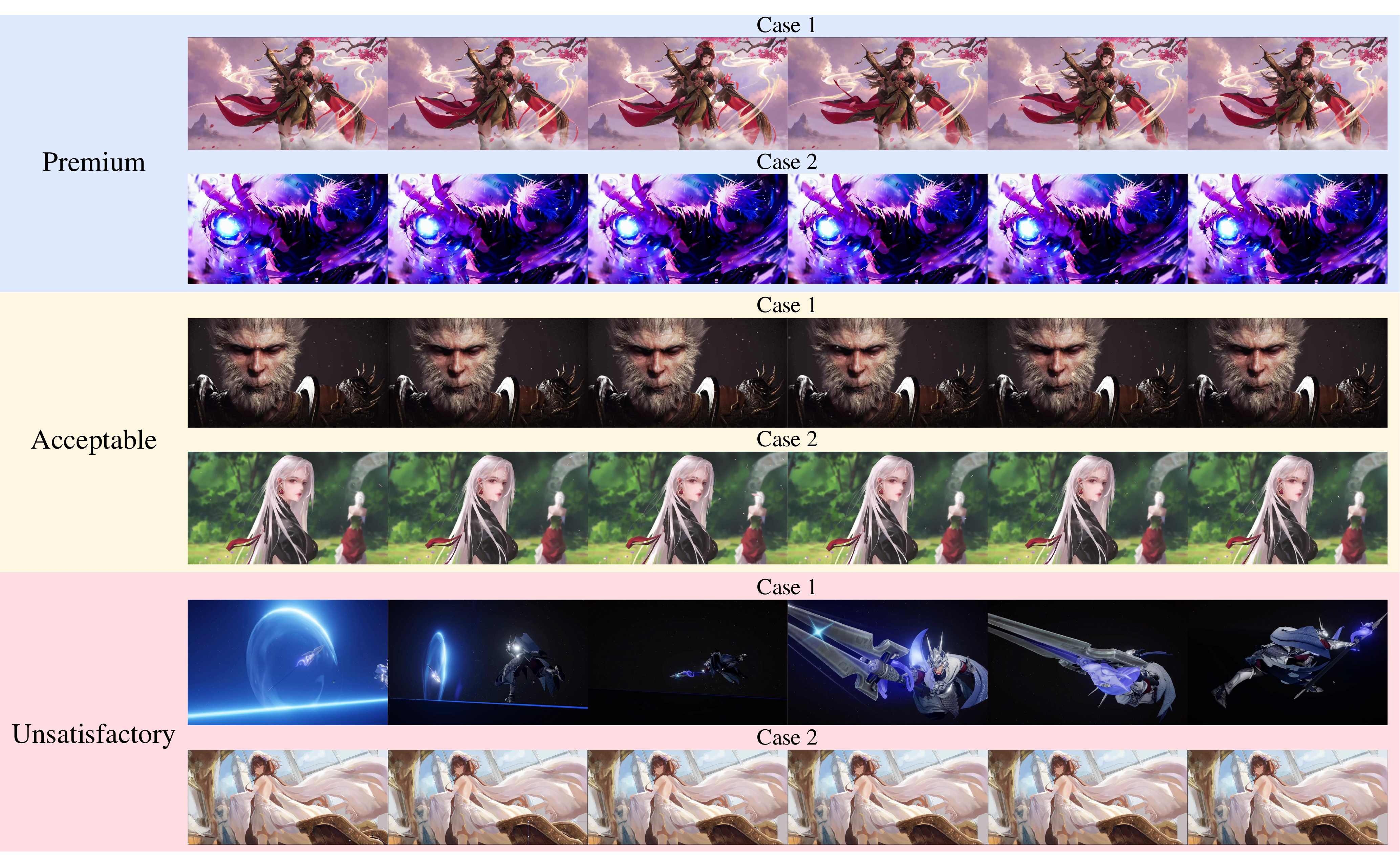}
    \caption{The dynamic illustration training data we collected is divided into three levels: \textbf{Level 1} consists of high-quality videos with obvious looping motions; \textbf{Level 2} includes medium-quality videos (Case 2) or those with minimal, non-looping motions (Case 1); and \textbf{Level 3} comprises static videos (Case 2) or those with transitions (Case 1).}
    \label{sec4:data}
\end{figure}

\begin{figure}[htbp]
    \centering
    \includegraphics[width=1.0\textwidth]{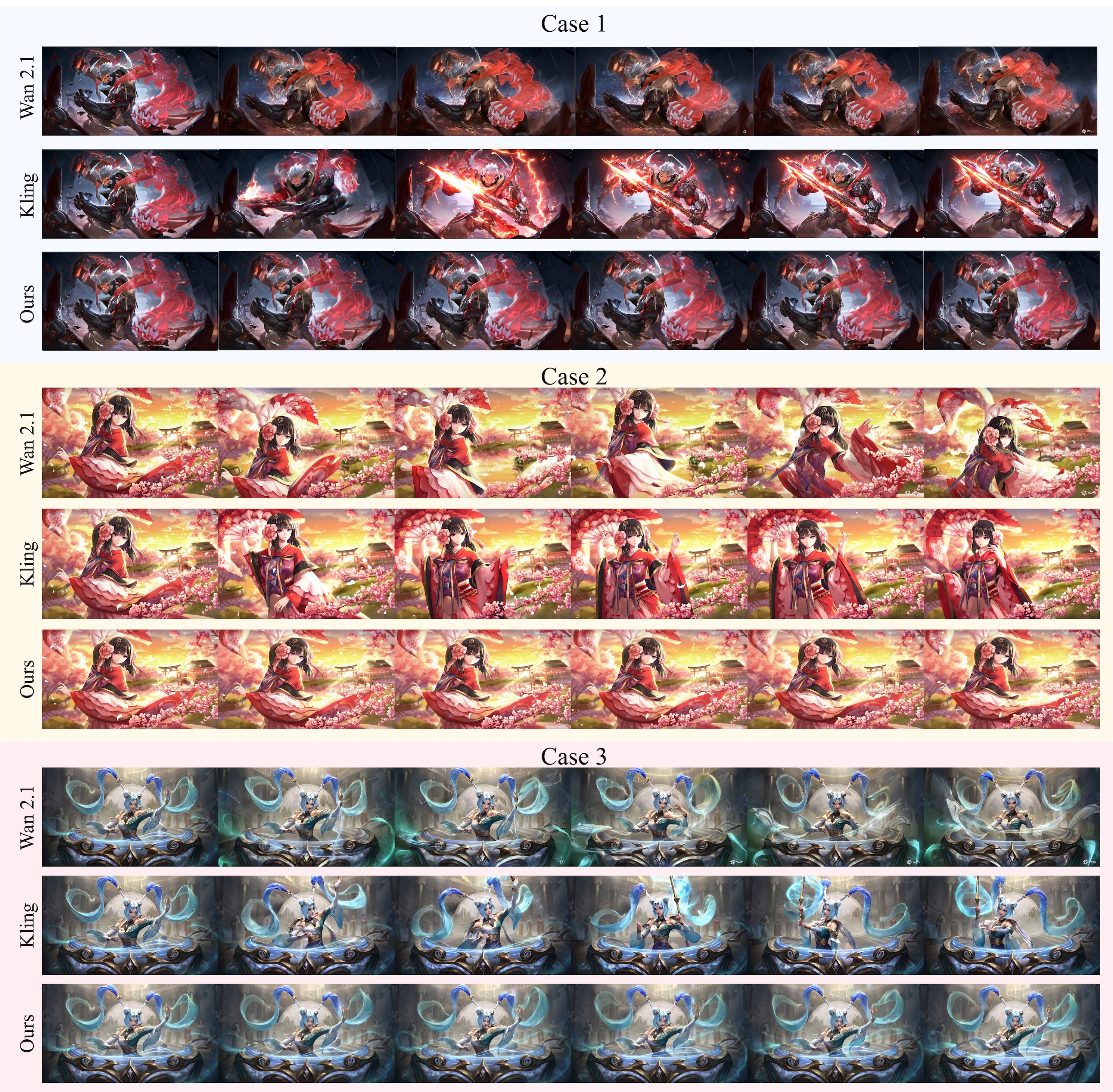}
    \caption{Qualitative comparisons with State-of-The-Art methods, \textit{i.e.}, Wan2.1 and Kling. Our method successfully achieves a looping subtle motion effect. In contrast, the results of Wan2.1 and Kling show significant changes in the characters' postures and movements.}
    \label{sec4:compare}
\end{figure}

\textbf{Level 1 - Premium}: These clips are characterized by high visual quality and resolution (above 1K). The motion within these clips is smooth and meets the requirement for subtle, looping animations. They represent the ideal standard for dynamic illustration generation.

\textbf{Level 2 - Acceptable}: Clips in this category have moderate visual quality with a resolution above 512 pixels. Although the motion is subtle, it is minimal and less pronounced than in Level 1 clips. These clips are adequate for use but do not exhibit the same level of excellence as Level 1.

\textbf{Level 3 - Unsatisfactory}: These clips are marked by low visual quality and a resolution below 512 pixels. They may contain transitions or camera movements, and the imagery is often static. Such clips are unsuitable for dynamic illustration generation and are excluded from further processing.

By meticulously annotating and categorizing our dataset, we ensure that our dynamic character illustration generation system is trained on high-quality data, leading to more accurate and appealing results. This rigorous data preparation process is crucial for achieving the desired level of animation fidelity and enhancing the overall user experience in video games and promotional materials. 

\subsubsection{Method}
Our approach leverages the foundational model of HunyuanVideo-I2V to perform Supervised Fine-Tuning (SFT) for dynamic character illustration generation. A key challenge in this task is ensuring that the generated video is both seamless and capable of looping, which requires a high degree of consistency throughout the video sequence.

To address this, we implemented a strategy that generates videos with identical first and last frames. By selecting the same frame for both the start and end of the video, we ensure high consistency and create a seamless loop where the end of the video transitions smoothly back to the beginning.

Our training process is divided into two distinct phases:

\textbf{Phase 1:} In this initial phase, we utilize Level-2 data as the training set. The focus here is on enabling the model to generate videos with subtle, looping motions. This phase establishes the foundational ability of the model to produce consistent and repetitive animations.

\textbf{Phase 2:} In the second phase, we introduce high-quality, high-resolution video data to the training process. This phase aims to enhance the visual quality and improve the smoothness of the motion. By incorporating premium data, we refine the model's output, ensuring that the generated videos not only loop seamlessly but also exhibit superior aesthetic and motion quality.

Through this two-phase training approach, our method effectively balances the need for looping consistency with the desire for high-quality visual output. This ensures that the dynamic character illustrations generated by our system are both visually appealing and functionally robust, meeting the demands of modern video games and animation applications.

\subsubsection{Evaluation}
We compare our method with several mainstream video generation techniques in Fig.~\ref{sec4:compare}, such as Wan2.1 and Kling, using the same prompt: "Make the characters, effects, weapons, and other elements in the image move slightly and repeat the same actions." Based on qualitative comparisons, our method successfully achieved a looping subtle motion effect by introducing identical first and last frames as conditional frames. In contrast, the results generated by Wan2.1 and Kling showed significant changes in the characters' postures and movements, which were not subtle nor looping, thus failing to meet the requirements for dynamic illustrations.

%% file: sections/video/4_vsr.tex
\subsection{Generative Video Super-Resolution}
\label{vsr}

% @wenqing
\subsubsection{Introductrion}

Due to constraints in computational resources and processing time, directly generating videos with higher resolutions has remained challenging, limiting its broader application in the gaming domain. Additionally, traditional video super-resolution methods \cite{he2024venhancer, wang2024apisr, zhou2024upscale, wang2025seedvr, wang2021real, yue2023resshift, zhang2024realviformer, yang2024motion, wang2024exploiting, pan2021deep, yang2021real} often suffer from issues related to temporal stability and detail preservation, resulting in flickering and abrupt changes in details during consecutive frame processing. To address these challenges, we introduce a novel approach of generative video super-resolution within the Hunyuan-Game framework. The primary objective of this task is to convert low-resolution input game videos into high-resolution outputs while preserving the intricate details of the video. As illustrated in Figure~\ref{fig:sr_pipe}, we design the latent-channel-concat strategy for latent feature fusion between low and high-resolution videos based on the HunyuanVideo-T2V \cite{kong2024hunyuanvideo} framework. Additionally, we employ a patch-wise training and inference methodology to reduce both the memory requirements and processing time during inference.

\subsubsection{Data}

To support the training of our generative video super-resolution model, we have curated a comprehensive dataset comprising both general videos and game videos.

\textbf{General Videos.}
All general videos included in the dataset meet the high-resolution standard of 2K or above. The dataset encompasses millions of general videos, ensuring a sufficient volume of training data. Videos are evenly distributed across diverse categories, such as animals, plants, landscapes, and human activities, providing a comprehensive and balanced representation.

\textbf{Game Videos.}
As detailed in Section~\ref{sec: data_filter}, we implement rigorous filtering criteria based on animation type, aesthetics, resolution, and motion scores, resulting in a collection of tens of thousands of high-quality game videos. We re-utilize the structure captions mentioned in Section~\ref{sec: data_anno} for annotating the game videos. These captions are sampled with different weights across four distinct types (e.g., long dynamic caption, long static caption, short dynamic caption, and short static caption) to enhance the model's ability to learn both static and dynamic video information during the training process. Furthermore, the camera motion annotations within the captions assist the model in better understanding and replicating the video's motion characteristics.

\subsubsection{Method}
\begin{figure}[htbp]
    \centering
    \includegraphics[width=1.0\linewidth]{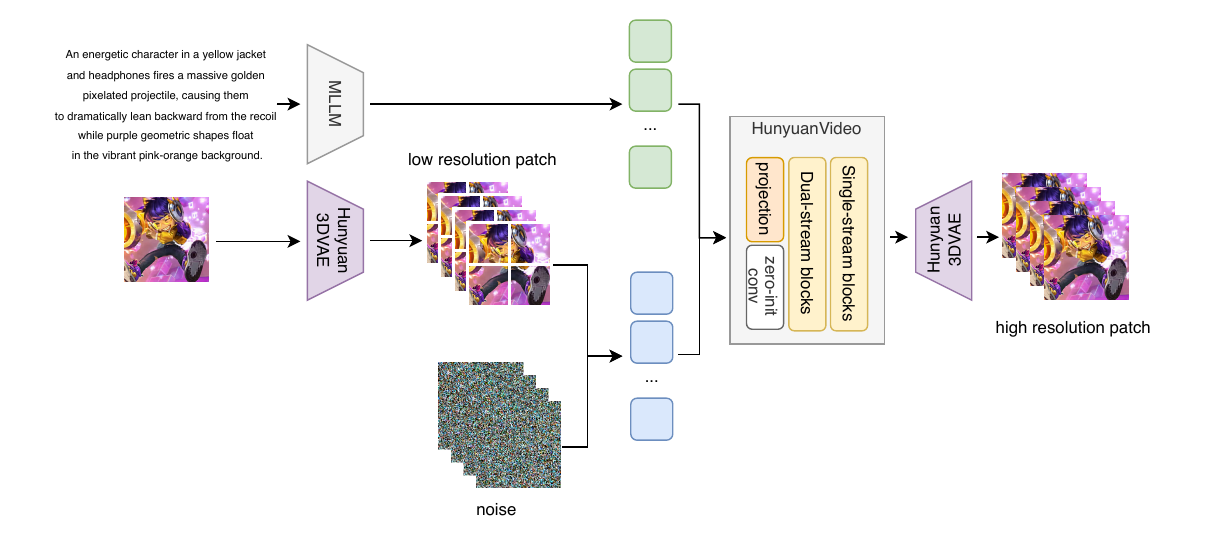}
    \caption{Framework of \name-Generative Video Super-Resolution.}
    \label{fig:sr_pipe}
\end{figure}

The task of video super-resolution encounters three significant challenges.
1) \textbf{Object attribute distortion.} As the weighting coefficient decreases during the denoising process, object attributes can be lost. Simply reducing the rate of coefficient decline can preserve attributes, but at the cost of decreased clarity.
2) \textbf{Temporal stability and detail preservation.} Existing algorithms often suffer from flickering and abrupt changes in details during consecutive frame processing.
3) \textbf{Computational efficiency.} Direct generative methods are computationally intensive and require substantial resources, leading to long processing times.

To address these challenges, we utilize the 13B HunyuanVideo-T2V \cite{kong2024hunyuanvideo} as our backbone and introduce several strategies.
1) \textbf{Latent-channel-concat strategy.} We replace traditional weighted-sum fusion with latent-channel-concat, which concatenates the latent features of high-resolution videos (after adding noise corresponding to different time steps) with those of low-resolution videos along the channel dimension. Moreover, the convolutional layers in HunyuanVideo controlling the number of channels are modified accordingly: half inherit the weights from HunyuanVideo-T2V (i.e., the orange block projection of HunyuanVideo in Figure~\ref{fig:sr_pipe}), and the other half are initialized to zero (i.e., the white block zero-init conv of HunyuanVideo in Figure~\ref{fig:sr_pipe}).
2) \textbf{Two-stage training strategy.} In the first stage, we use millions of general videos for pretraining to enable the model with universal video super-resolution capabilities. In the second stage, we employ carefully curated high-quality game videos for fine-tuning to enhance the model's performance in gaming scenarios, allowing it to learn the unique visual characteristics and motion patterns of game videos.
3) \textbf{Patch-wise training and inference.} We use a patch-wise training strategy with a spatial scale of 768×768 pixels and a temporal window of 129 frames to construct 3D training sample units. Each sample randomly selects 2 training units, optimizing memory usage and improving computational efficiency during spatiotemporal joint modeling. We introduce a tiled patch-wise inference strategy, dividing the input video into multiple overlapping sub-regions. Each sub-block is processed independently during the diffusion process, and the overlapping regions between adjacent blocks are averaged to eliminate boundary artifacts. Additionally, we reduce the number of inference steps through techniques like distillation, further enhancing computational efficiency alongside the block-wise approach.

Through these architectural designs and training/inference strategies, our model effectively balances video content and temporal consistency with significant improvements in clarity, meeting the demands of game video applications. 

\subsubsection{Evaluation}

\begin{figure}[htbp]
    \centering
    \includegraphics[width=1.0\textwidth]{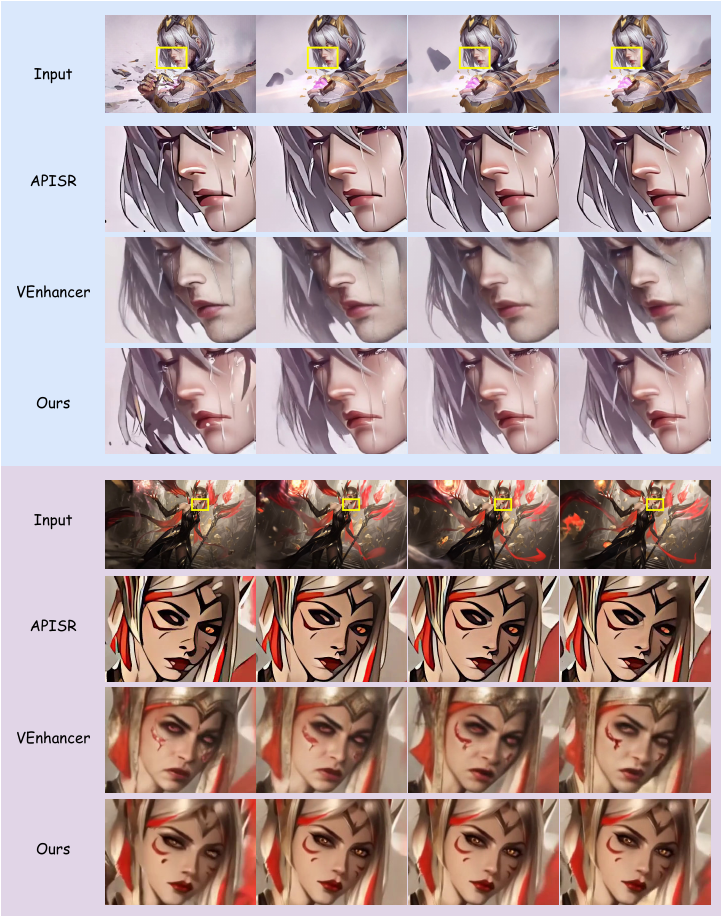}
    \caption{Qualitative results of Hunyuan-Game-Video Super-Resolution. We compare our method with other SOTA methods, including APISR \cite{wang2024apisr} and VEnhancer \cite{he2024venhancer}. Ours present clearer and better restoration with more natural details, preserving original color saturation or contrast, without sharper edge lines.}
    \label{fig:sr_comp}
\end{figure}

\begin{table}[htbp]
\centering
\caption{Quantitative results of Hunyuan-Game-Video Super-Resolution. }
\label{tab:sr_results}
\scalebox{0.95}{
\begin{tabular}{lccc}
\toprule
Model & Success score & Overall score \\
\midrule
VEnhancer \cite{he2024venhancer} & 0.0256 & 0.0385 \\
APISR \cite{wang2024apisr} & 0.3038 & 0.4430 \\
Ours & \textbf{0.5696} & \textbf{0.6076} \\
\bottomrule
\end{tabular}
}
\end{table}

\textbf{Qualitative results.}
 The qualitative comparisons between our method and other SOTA methods are visualized in Figure~\ref{fig:sr_comp}. To align the results, we consistently extracted the same four frames and enlarged the same specific region from each restored video. As depicted in Figure~\ref{fig:sr_comp}, our method achieves the clearest and most natural video restoration performance. In contrast, APISR \cite{wang2024apisr} tends to produce an outlined appearance, introducing sharp edge lines and enhancing color contrast in its video results. Meanwhile, VEnhancer \cite{he2024venhancer} exhibits significant alterations in video content and a noticeable degree of blurriness in its outputs.

\textbf{Quantitative results.}
Our evaluation dataset is composed of two equally balanced parts, real videos and generated videos. We selected 40 real video samples, with a similar distribution of the game videos in training data, and 40 game videos generated by the Hunyuan-Game I2V (illustrated in Section~\ref{i2v}), forming a comprehensive test set of 80 videos. For the real videos, prompts were generated using the caption model illustrated in Section~\ref{sec: data_anno}. Moreover, for the generated videos, we re-utilized the prompts originally employed during their generation.

Each video after super-resolution was independently evaluated by three annotators. To ensure scoring fairness, each annotator viewed the results from different models simultaneously. The scoring result was divided into three categories. 0 points for "unqualified", indicating noticeable super-resolution artifacts or disharmony. 1 point for "qualified", signifying a minor improvement after super-resolution. And 2 points for "excellent", denoting a significant enhancement post-super-resolution. In Table~\ref{tab:sr_results}, the success score represents the average score achieved by the model in the qualified category, while the overall score denotes the average score across all scoring categories. We compare our Hunyuan-Game-Video Super-Resolution with several state-of-the-art methods, including VEnhancer \cite{he2024venhancer} and APISR \cite{wang2024apisr}. As illustrated in Table~\ref{tab:sr_results}, our model achieves the best performance in both the success score and the overall score metrics.

%% file: sections/video/5_interactive.tex
\subsection{Interactive game video generation}
\label{interactive}
\subsubsection{Introduction}

Recent advances in visual generation have explored developing World Models~\cite{agarwal2025cosmos,feng2024matrix,guo2025mineworld,ha2018world,sora,worldlabs2024,parker2024genie,yang2023learning,yang2024position}, the creation of diverse worlds in various scenes. These models focus on interactivity and exploration, enabling dynamic 3D/4D virtual environments with temporal-spatial coherence.  
% These models has the potential to change the next generation of game interactive experience. 
Unlike traditional static scene generation or video generation, world models integrate physics simulation and behavioral interaction, allowing players to manipulate terrain through natural inputs like keyboard/mouse operations or even single-image prompts. For instance, WorldLabs~\cite{worldlabs2024} demonstrates real-time 3D scene reconstruction from static images, while Genie 2~\cite{parker2024genie} enables physics-compliant interactions through its latent action modeling. However, previous methods still face fundamental constraints, including computational efficiency, the fidelity of dynamic scenes, and long-sequence consistency.

Building upon Hunyuan-Game, we propose Hunyuan-GameCraft, an interactive game scene video generation model with high dynamics and fidelity. To achieve fine-grained controllable game video synthesis with temporal coherence, we unify diverse common keyboard/mouse options in games (W, A, S, D, ↑, ←, ↓, →, Space, etc.) into a shared camera representation space. This enables seamless interpolation and combination between different cinematographic operations while maintaining physical plausibility. Besides, to accelerate the inference speed and improve the interaction experience, we implement the model distillation, based on Phased Consistency Model~\cite{wang2024phased}. This distillation achieves a 10–20× acceleration in inference speed, reducing latency to less than 10s per action.

\subsubsection{Data Preprocessing}
To enable the generation of high-fidelity gameplay videos, we curated a diverse dataset comprising gameplay recordings from over 100 AAA first-person perspective games from critically acclaimed titles, such as \textit{Assassin's Creed}, \textit{Red Dead Redemption}, \textit{Hogwarts Legacy}, \textit{Cyberpunk 2077}, and so on. These recordings were meticulously sourced to cover a wide spectrum of environments, actions, and visual styles, ensuring rich representation of high-fidelity graphics, diverse environments, dynamic lighting, and complex in-game interactions.

Our end-to-end data processing framework comprises four stages to ensure cinematic quality and accurate and controllable outputs for gameplay video generation:
We first split gameplay sequences into semantically coherent clips using PySceneDetect~\cite{Pyscenedetect}. Then we remove low-fidelity segments by scoring quality and luminance based on pixel histograms.
% Flag abrupt scene transitions using optical flow variance thresholds for later stabilization.
% \textbf{Action Segmentation and Analysis:} 
To temporally partition long gameplay sequences into distinct action units (e.g., turn left, move forward), we compute dense optical flow from RAFT~\cite{teed2020raft} and leverage gradients as motion saliency indicators. According to the flow vectors, sudden movements in camera or player motion (e.g., rapid aiming, scene transitions) are detected as candidate split points.
% After partition, we reconstruct 6-DoF camera trajectories via Monst3r~\cite{} to model viewpoint dynamics.
% As for game video data annotation, we follow the design introduced in Section~\ref{}, focusing on the description of camera motion and various environments.  
After partition, we reconstruct six-degree-of-freedom (6-DoF) camera trajectories using MonST3R~\cite{zhang2024monst3r}, enabling precise modeling of viewpoint dynamics. For game video annotation, we follow the annotation pipeline described in Section~\ref{sec: data_anno}, with particular emphasis on environment transitions and lighting variations. 

\subsubsection{Method}

\begin{figure}[t]
    \centering
    \includegraphics[width=\linewidth]{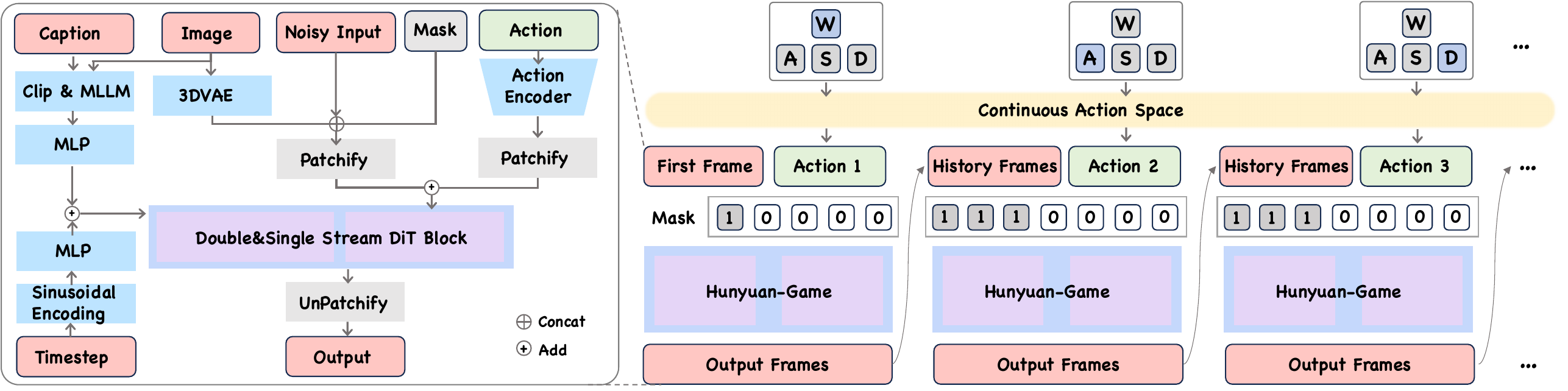}
    \caption{Overall framework of Hunyuan-GameCraft.}
    \label{fig:gamecraft}
\end{figure}

Based on the image-to-video backbone, the core architecture of Hunyuan-GameCraft combines a light-weight action encoder and a variable mask indicator, which enables flexible inference and training for long videos. The overall framework is shown in Fig.~\ref{fig:gamecraft}. The mask indicates the input is history or noise to be denoised, represented by 1 or 0, respectively.

Specifically, given discrete keyboard/mouse options (W, A, S, D, ↑, ←, ↓, →, Space, etc), we transform these options to the continuous camera space with pre-defined motion parameters, such as speed, angle, and so on. As for camera representations, following previous camera-controlled arts~\cite{he2024cameractrl,wang2024motionctrl,bahmani2024ac3d}, we leverage Plücker embeddings~\cite{sitzmann2021light} as a more geometric interpretation for each pixel of a video frame, which can provide a more informative description of camera pose information to the base video generators. 
The designed light-weight action encoder includes two spatial-temporal compression convolution modules. The encoder is zero-initialized.

\begin{figure}[t]
    \centering
    \includegraphics[width=\linewidth]{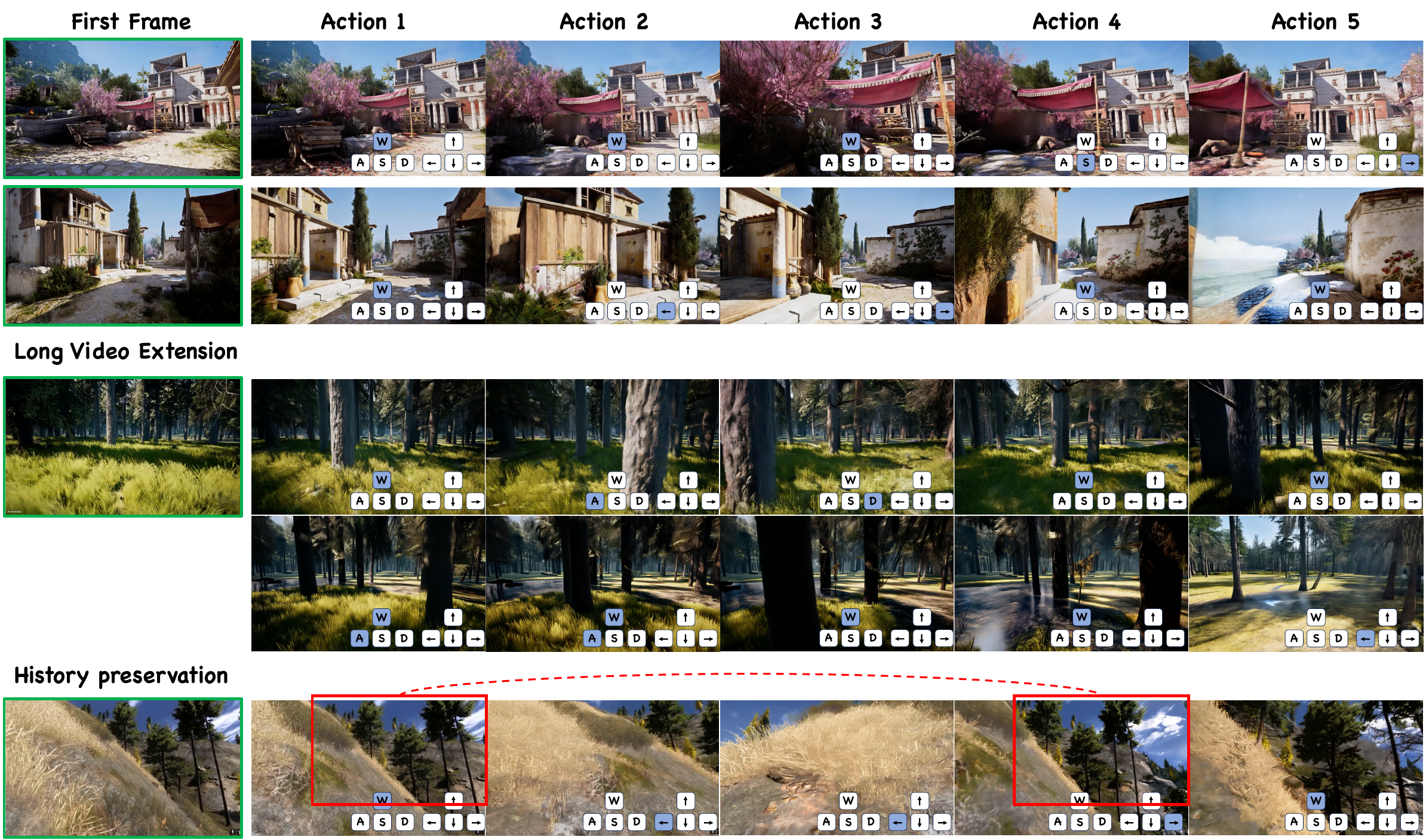}
    \caption{Qualitative results of interactive game video generation. Given discrete keyboard action signal, Hunyuan-GameCraft supports action control across diverse open-domain scenes with spatial coherence. Besides, our introduced hybrid-conditioned auto-repressive video extension framework helps to extent short clips to long sequence with history information preservation. In our case, blue-lit keys indicate key presses. W, A, S, D represent transition movement and ↑, ←, ↓, → denote changes in view angles.  }
    \label{fig:gamecraft-eval}
    \vspace{-0.5em}
\end{figure}
To overcome challenges in generating long, consistent and interactive videos, we introduce a hybrid-conditioned auto-regressive video extension framework. Unlike prior approaches that suffer from temporal inconsistency and quality loss, this method combines multiple guidance signals—such as single frames, previous latents, or full clip segments—during training to balance fidelity, consistency, and interactivity. Each video segment is denoised using a causal VAE, guided by clean "head" conditions through flow-matching. A binary mask enables precise control over the denoising process.

Extensive experiments reveal a trade-off: richer head conditions improve coherence and quality but reduce responsiveness to new inputs. To mitigate this, hybrid-conditioned training blends all conditioning types, achieving a strong balance between interactivity and generation quality. This unified approach also simplifies deployment by supporting both initial frame generation and video extension in a single model architecture.

\subsubsection{Evaluation}

In Fig~\ref{fig:gamecraft-eval}, we present qualitative results demonstrating the effectiveness of Hunyuan-GameCraft in generating interactive game videos across diverse open-domain environments. Given discrete keyboard action signals as input, our model supports fine-grained action control with high spatial and temporal coherence. Specifically, the model responds accurately to common gameplay controls such as movement (W, A, S, D) and camera adjustments (↑, ←, ↓, →), enabling realistic agent navigation and viewpoint changes within dynamically evolving scenes. For visualization clarity, key press events are denoted by blue-lit keys overlaid on the video frames.

Our proposed hybrid-conditioned autoregressive video extension framework further enables seamless extension from short clips to long-form video sequences, while preserving historical context and visual continuity (see {\color{red}red} box). This is achieved by conditioning each generated segment on both past latent representations and action signals, effectively maintaining motion consistency and narrative flow over extended durations.

The results highlight Hunyuan-GameCraft’s ability to synthesize high-fidelity video sequences that align with user inputs while adapting to complex scene dynamics. The generated outputs exhibit consistent object positioning, smooth transitions between actions, and plausible environmental responses, thereby enhancing the overall realism and immersion of the generated video sequences.

%% file: sections/conclusion.tex
\section{Conclusion}
% In this report, we have introduced \textbf{\name}, a groundbreaking foundation model specifically designed for the generation of professional-grade game videos. By leveraging the HunyuanVideo-I2V framework and a customized data pipeline, \name\ addresses the unique challenges of game video synthesis, achieving state-of-the-art performance in visual fidelity and motion naturalness.

% Our extensive experiments demonstrate that \name\ not only surpasses existing models like Kling and Wan but also opens up new possibilities for the gaming industry through its versatile applications. The implementation of four professional downstream tasks—360° Avatar Video Synthesis, Dynamic Illustration Generation, Real-time Game Scene Generation, and Generative Super-Resolution for Game Videos—showcases the model's adaptability and potential to transform game content creation.

% By open-sourcing \name\ and its associated models, we aim to foster a collaborative environment that encourages innovation and exploration within the community. We believe that this initiative will not only drive advancements in game video generation but also pave the way for broader applications across various sectors of the gaming industry.

% As the demand for high-quality, immersive gaming experiences continues to grow, \name\ stands as a testament to the potential of foundation models in shaping the future of game development and storytelling. We look forward to seeing how the community will build upon this work, driving further innovation and creativity in the field.

In this report, we present \textbf{\name}, the groundbreaking foundation model suite dedicated to professional-grade game asset generation. Our framework comprises four image generation models and five video generation models, covering a wide spectrum of tasks including text-to-image generation, game visual effects generation, transparent and seamless image generation, game character generation, image-to-video generation, 360° A/T pose avatar video synthesis, dynamic illustration generation, generative video super-resolution and interactive game video generation. Extensive experiments demonstrate that \name\ achieves state-of-the-art performance in visual fidelity, temporal consistency, and adaptability across diverse game scenarios, outperforming existing baselines.
By 
% open-sourcing 
introducing
\name, we aim to empower both the research community and industry practitioners to accelerate innovation in game asset creation, reduce manual workload, and explore new creative possibilities. We believe \name\ lays a solid foundation for future research and applications in intelligent game development, ultimately contributing to more immersive and dynamic gaming experiences. We look forward to seeing how the community will build upon this work, driving further innovation and creativity in the field.
% Addressing the critical challenges in intelligent game creation—such as the scarcity of large-scale game-specific datasets, alignment with vertical game scenarios, multi-dimensional aesthetic evaluation, and the need for high visual fidelity and interactive content generation—\name\ integrates extensive domain knowledge and advanced generative techniques to significantly enhance the quality and efficiency of game content production.

\newpage
\textbf{\Large Project Contributors}

$\bullet$ \textbf{Project Sponsors:} Jie Jiang, Linus, Yuhong Liu, Di Wang

$\bullet$ \textbf{Project Leaders:} Qinglin Lu, Shuai Shao, Longhuang Wu

$\bullet$ \textbf{Core Contributors:} 

\begin{quote}
$\circ$ \textbf{Data \& Recaptioning:} Chao Zhang, Hongxin Zhang, Qiaoling Zheng, Weiting Guo, Yingfang Zhang, Xinchi Deng, Duojun Huang, Yixuan Li
\vspace{0.5em}

$\circ$ \textbf{Algorithm:} Ruihuang Li, Caijin Zhou, Shoujian Zheng, Jianxiang Lu, Jiabin Huang, Comi Chen, Junshu Tang, Guangzheng Xu,  Hongmei Wang, Jiale Tao, Donghao Li, Wenqing Yu, Senbo Wang, Zhimin Li, Yetshuan Shi, Junkun Yuan
\vspace{0.5em}

$\circ$ \textbf{Art Designer:} Renjia Wei, Yulin Jian
\end{quote}

$\bullet$ \textbf{Contributors:} Yuan Zhou, Joey Wang, Qin Lin, Tianxiang Zheng, Jingmiao Yu, Jihong Zhang, Caesar Zhong, Haoyu Yang, Yukun Wang, Wenxun Dai, Jiaqi Li, Linqing Wang, Qixun Wang, Zhiyong Xu, Jiangfeng Xiong, Weijie Kong, Xuhua Ren, Zhengguang Zhou, Jiaxiang Cheng, Bing Ma, Shirui Huang, Jiawang Bai, Chao Li, Sihuan Lin, Yifu Sun

\newpage